\documentclass[10pt,letterpaper]{article}
\usepackage[top=0.85in,left=2.75in,footskip=0.75in]{geometry}

\def\highlight{0}

\usepackage{nameref,hyperref}
\usepackage[table]{xcolor}

\usepackage{times}
\usepackage{epsfig}
\usepackage{graphicx}
\usepackage{amssymb}
\usepackage{amsthm}
\usepackage{amsmath}
\usepackage{amsfonts}
\usepackage{url}
\usepackage{array}
\usepackage[]{algorithm2e}
\usepackage{subcaption}
\usepackage{multirow}
\usepackage{verbatim}
\usepackage{hyperref}
\usepackage[font=normalsize]{caption}
\usepackage{rotating}

\theoremstyle{definition}
\newtheorem{definition}{Definition}[section]

\newcommand{\norm}[1]{\|{#1}\|}

 \newcommand{\velf}{\mathbf{v}}

\newcommand{\normvec}{\mathbf{g}}

\DeclareMathOperator*{\argmin}{arg\,min}

\graphicspath{{Figures_inpdf/}}

\newcolumntype{+}{!{\vrule width 2pt}}

\newlength\savedwidth

\raggedright
\usepackage[aboveskip=1pt,labelfont=bf,labelsep=period,justification=raggedright,singlelinecheck=off]{caption}

\makeatletter
\renewcommand{\@biblabel}[1]{\quad#1.}
\makeatother

\usepackage{lastpage,fancyhdr,graphicx}
\usepackage{epstopdf}
\fancyheadoffset[L]{2.25in}
\fancyfootoffset[L]{2.25in}
\begin{document}
\vspace*{0.2in}

\begin{flushleft}
{\Large
\textbf\newline{Discovery and recognition of motion primitives in human activities} 
}
\newline
\\
Marta Sanzari\textsuperscript{1},
Valsamis Ntouskos\textsuperscript{1},
Fiora Pirri\textsuperscript{1},
\\
\bigskip
\textbf{1} Dipartimento di Ingegneria Informatica Automatica e Gestionale, University of Rome 'Sapienza', Alcor LAB
\bigskip

\end{flushleft}
\section*{Abstract}
We present a novel framework for the automatic discovery and recognition of  motion primitives in videos of human activities. Given the 3D pose of a human in a video,  human motion primitives are discovered by optimizing the `motion flux', a quantity which captures the motion variation of a group of skeletal joints.  A normalization of the primitives is proposed in order to make them  invariant with respect to a subject anatomical variations and data sampling rate. The discovered primitives are unknown and unlabeled and  are unsupervisedly  collected into classes via a hierarchical non-parametric Bayes mixture model. Once classes are determined and labeled they are further analyzed for establishing models for recognizing discovered primitives. Each primitive model is defined by a set of learned parameters. 
 Given new video data and given the estimated pose of the subject appearing on the video, the motion is segmented into primitives, which are recognized with a probability given according to the parameters of the learned models.
 Using our framework we build a publicly available dataset of human motion primitives, using sequences taken from well-known motion capture datasets. We expect that our framework, by providing an objective way for discovering and categorizing human motion, will be a useful tool in numerous research fields  including video analysis, human inspired motion generation, learning by demonstration,  intuitive human-robot interaction, and human behavior analysis.

\section{Introduction}
\newcommand{\figwdthA}{1.8cm}
\newcommand{\figwdthB}{2.5cm}

\noindent
\begin{figure}[tph!]
\centering
\includegraphics[width=1\textwidth]{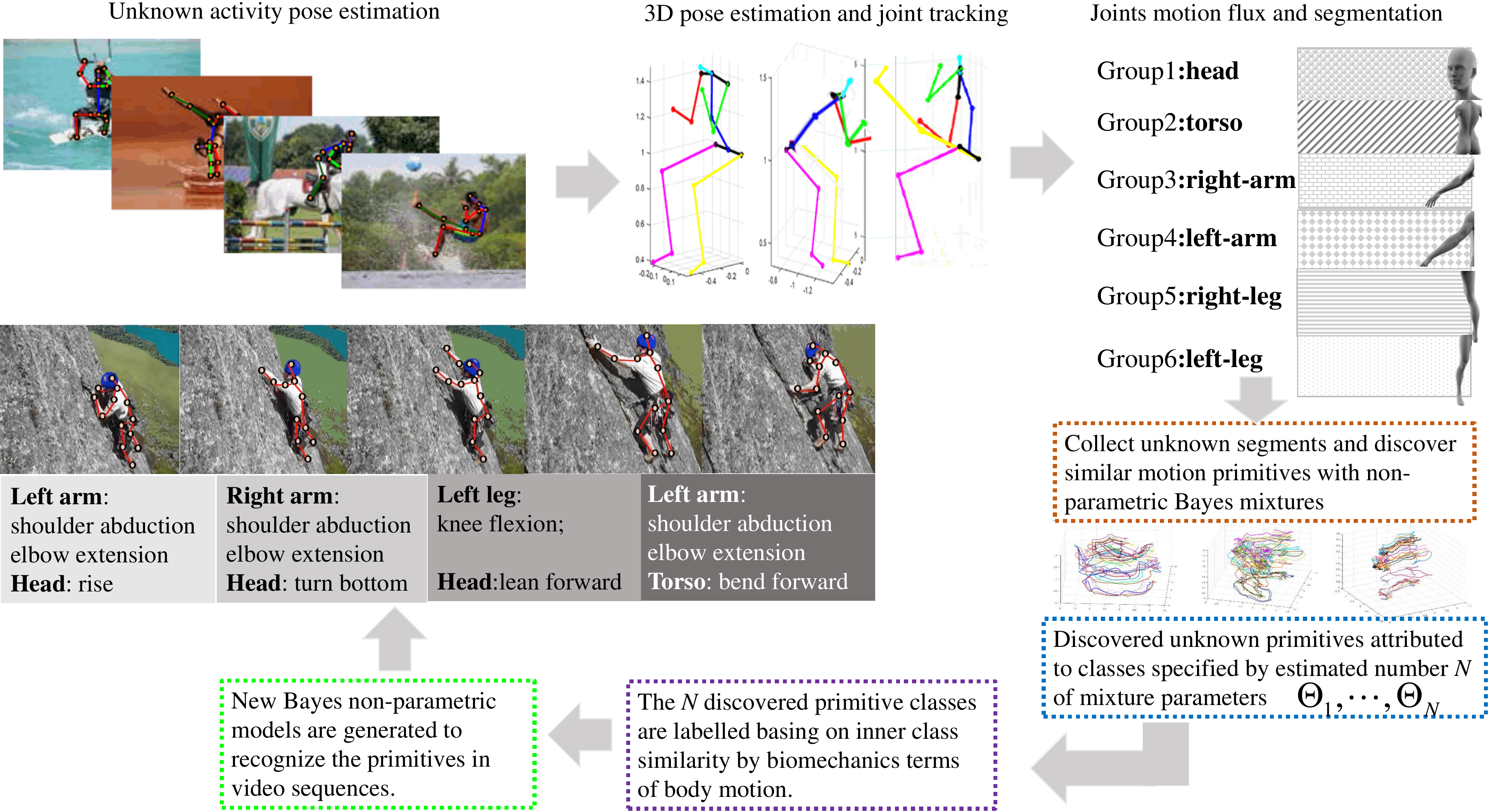}
\caption{The above schema presents the proposed framework and the process to obtain from video sequences the discovered motion primitives.}\label{fig:framew-schema}
\end{figure}

Activity recognition is widely acknowledged as a core topic in computer vision, witness the huge amount of research done in recent years spanning a wide number of applications  from sport to cinema, from human robot interaction to security and rehabilitation.   

Activity recognition has evolved from earlier  focus   on action recognition and gesture recognition. 
The main difference being that activity recognition is completely general as it concerns any kind of human activity, which can last few seconds or minutes or hours, from daily activities such as cooking, self-care, talking at the phone, cleaning a room, up to sports or recreation such as playing basketball or fishing. Nowadays there are a number of publicly available datasets dedicated to the collection of any kind of human activity, likewise a number of challenges (see for example the ActivityNet challenge \cite{ghanem2017}).

 On the other hand, the interest in motion primitives is due to the fact that they  are essential for deploying an activity.   Think about sport activities, or cooking, or performing arts, which require to  purposefully select a specific sequences of  movements. Likewise daily activities such as cleaning, or cooking, or washing the dishes or preparing the table require precise motion sequences to accomplish  the task. 
Indeed, the compositional nature of human activities, under body and kinematics constraints, has attracted the interest of many research areas such as in computer vision  \cite{yang2013,holte2010}, in  neurophysiology \cite{flash2005,polyakov2017}, in  sports and rehabilitation \cite{ting2015}, and in  biomechanics \cite{hogan2012} and in robotics   \cite{amor2014,moro2012,azad2007}.  

The goal of this work is to automatically discover the  start and end points  where primitives  of 6 identified  body parts occur throughout the course  of an activity, and recognize each of the occurred primitives. The idea is that these primitives sort out a non-complete set of human movements, which combined together can form a wide range of human activities, in so providing a compositional approach to the analysis of human activities.

The steps of the proposed method are as follows. Given a video of a human activity  both the 2D pose and 3D pose of the human are estimated  (see \cite{sanzari2016}, and also \cite{tome2017}). 
Once the 3D poses of the joints of interest are determined,  we compute the {\em motion flux}. The motion flux method provides a model from first principles for  human motion primitives, and  it effectively  discovers  where primitives begin and end on human activity motion trajectories.

Motion primitives discovered by the motion flux are  unknown: they are segments of motion about which only the  involved specific body part is known. 
These primitives  are collected into classes by a non-parametric Bayes model, namely the Dirichlet process mixture model (DPM), which gives the freedom to not choose the number of mixture components. By suitably eliminating very small clusters it turns out that discovered primitives can be collected into 69 classes (see Fig. \ref{fig:primitives}). For each of them the mixture model returns a parameter set identifying the precise primitive class. We label the computed parameters with terms taken from the biomechanics of human motion, by inspecting only a representative primitive for each discovered class. Out of these generated classes we form a new layer of the hierarchical model, to generate the parameters for each class, further used for primitives recognition.
Under this last models each primitive category is approximated by a DPM with a number of components mirroring the inner idiosyncratic behavior of each primitive class.

Motion primitives classification is finalized by providing a label for each primitive. Namely,  given an activity (possibly unknown) and an unknown primitive  discovered by motion flux, we find the model the primitive belongs to, hence  the primitive is labeled by that model.  

Experiments show that the motion flux is a good model for segmenting the motion of  body parts.    Likewise, the unsupervised non-parametric model provides both a good classification of similar motion primitives and a good estimation of primitive labels, as shown in the results (see Section \ref{sec:prim_results}).  
The approach therefore is quite general and  it turns out to be very useful to any researcher who would like to explore the compositional nature of any activity, using both the proposed method and the motion primitives dataset provided. 

To the best of our knowledge just  few works, among which we recall \cite{yang2013,holte2010}, have  faced the problem of discovering motion primitives in video activities or motion capture (MoCap) sequences,  quantitatively evaluating the ability to recognize them.

{
\if\highlight1
\color{blue}
\fi
Despite the lack of works on motion primitives we show that they are  quite an expressive {\em language} for ascertaining specific human behaviors.
To prove that, in a final application for video surveillance, described in Section \ref{sec:application},  we show that motion primitives can play a compelling role in detecting distinct classes of dangerous activities. In particular, we show that dangerous activities can be detected with off-the- shelf classifiers, once motion primitives have been extracted in the videos. Comparisons with state of the art results  prove the relevance of motion primitives in discovering  specific  behaviors, since motion primitives  embed significant time-space features easily usable for classification.
}

The contributions of the work, schematically shown in Fig. \ref{fig:framew-schema} are the followings:\\
\noindent
1. We introduce the motion flux method to discover motion primitives, relying on the variation of the velocity of a group of joints.\\ 
\noindent
2. We introduce a hierarchical model for the classification and  recognition of the unlabeled  primitives, discovered by the motion flux.\\
{
\if\highlight1
\color{blue}
\fi
\noindent
3. We show a relevant application of human motion primitives for video surveillance.
}
\\
\noindent
4. We created a new dataset of human motion primitives  from three  public MoCap datasets (\cite{ionescu2014}, \cite{CMU}, \cite{mandery2015}).  

\section{Related work}\label{sec:prim_related}
Human motion primitives are investigated in several research areas, from  neurophysiology to vision to robotics and biomechanics.
Clearly, any methodology has to deal with the vision process, and many of the earliest more relevant approaches to human motion highlighted that understanding human motion requires view independent representations \cite{weinland2006,li2010} and
that a fine grained analysis of the motion field is paramount
to identify primitives of motion.  
In early days this required a massive effort in visual analysis \cite{chellappa2008} to obtain the poses, the low level features, and segmentation.   Nowadays, scientific and technological advances have made it possible to exploit several methods to measure human motion, such as the availability of a number of MoCap databases \cite{ionescu2014,mandery2015,Sigal2009}, see for a review \cite{moeslund2006}. Furthermore recent findings  result in methods that can deliver 3D human poses from videos if not even from single frames \cite{akhter2015,sanzari2016,zhou2016,tome2017}.
Since then 3D MoCap data have been widely used to study and understand human motion, see for example \cite{Ntouskos-2013ICPRAM,Ntouskos-2014,Pirri-2011} in which Gaussian Process Latent Variable Models or Dirichlet processes are used to classify actions, or \cite{natola2015} in which a non-parametric Bayesian approach is used to generate behaviors for body parts and classify actions based on these behaviors. In \cite{natola2015b} temporal segmentation of collaborative activities is examined, or in \cite{fanello-2010} different descriptors are exploited to achieve arm-hand action recognition.

\paragraph{Neurophysiology}
Neurophysiology studies on motion primitives \cite{bizzi1995,flash2005,flash2013,viviani1995,flash2007,biess2007} are based on the idea that  kinetic energy and muscular activity are optimized in order to conserve energy.
In these works it has been observed that curvature and velocity of joint motion are related. Earliest works such as Lacquaniti et al. \cite{Lacquaniti1983} proposed a relation between curvature and angular velocity. In particular, using their notation, letting $C$ be the curvature and $A$  the angular velocity, they called the equation $A=KC^{\frac{2}{3}}$  the Two-Thirds Power law, valid for certain class of two-dimensional movements. Viviani and Schneider \cite{viviani1991} formulated an extension of this law, relating the radius of  curvature $R$ at any point $s$ along the trajectory with the corresponding tangential velocity $V$, in their notation:

\begin{equation} \label{eq:viviani}
V(s)=K(s)\left(\frac{R(s)}{1+{\alpha}R(s)}\right)^{\beta}
\end{equation}
\noindent
where the constants $\alpha\geq 0$, $K(s)\geq 0$ and $\beta$ has a value close to $=\frac{1}{3}$. 
An equivalent  Power law for trajectories in 3D space is introduced  by \cite{Maoz2014} and it is called the curvature-torsion power law and is defined as $\nu = \alpha \kappa^{\beta} |\tau|^{\gamma}$, 
where $\kappa$ is the curvature of the trajectory, $\tau$ the torsion, $\nu$ the spatial movement speed, $\beta$ and $\gamma$ are constants. 

\paragraph{Computer Vision}
The interpretation of motion primitives as simple individual actions or gestures is often purported, in any case they are related to segmentation of videos and 3D motion capture data.
Many approaches explore video sequences segmentation to align similar action behaviors \cite{gong2014} or for spatio-temporal annotation as in \cite{lillo2016}. Lu et al. \cite{corso2015} propose to use a hierarchical Markov Random Field model to automatically segment human action boundaries in videos. Similarly, \cite{Bouchard2007} develop a motion capture segmentation method.
Besides these works,  only \cite{delvecchio2003,yang2013,holte2010,endres2013} have targeted motion primitives, to the best of our knowledge. \cite{delvecchio2003} focuses on 2D primitives for drawing, on the other hand \cite{yang2013} does not consider 3D data and generate the motion field considering Lukas-Kanade optical flow for which Gaussian mixture models are learned. None of these approaches provide quantitative results for motion primitives, but only for action primitives, which makes their method not directly comparable with ours. \cite{holte2010,endres2013} use 3D data and explicitly mention motion primitives, providing quantitative results.  The authors   account for the velocity field via optical flow basing the recognition of motion primitives  on harmonic motion context descriptors.  Since  \cite{holte2010} deal only with upper torso gestures we compare with them only the primitives they mention. In \cite{endres2013} the authors achieve motion primitives segmentation from wrist trajectories of sign language gestures, obtaining unsupervised segmentation  with Bayesian Binning. Again here no comparison for motion primitives discovery or recognition is possible as original data are not available.

\paragraph{Robotics}
 In robotics  the paradigm of transferring human motion primitives to robot movements is paramount for imitation learning and, more recently to implement human-robot collaboration \cite{ijspeert2013}.  
A good amount  of research in robotics has approached primitives in terms of Dynamic Movement Primitives (DMP) \cite{ijspeert2013} to model  elementary motor behaviors as attractor systems, representing them with differential equations. Typical applications are learning by imitation or learning from demonstration \cite{gams2016,Pastor2009,kober2009,park2008}, learning task specifications \cite{Ureche2015}, modeling interaction primitives \cite{amor2014}. 
Motion primitives are   represented  either via Hidden Markov models or Gaussian Mixture Models (GMM).  \cite{asfour2008} present an approach based on HMM for imitation learning of arm movements, and   \cite{luo2015} model arm motion primitives  via GMM.

It is apparent that in most of the  approaches motion primitives are only observed and modeled, instead we are able to learn and model them using respectively the {\em motion flux} quantity and a hierarchical model. The main contribution of our work is indeed the introduction of a new ability for a robot to automatically discover motion primitives observing 3D joints raw pose data. The outcome of our approach is also a motion primitives dataset not requiring human manual operation.

Our view of motion primitive shares the hypothesis of energy minimality during motion,  fostered by neurophysiology, likewise the idea to characterize movements using the proper geometric properties of the skeleton joints space motion. However, for primitive discovery, we go beyond  these approaches capturing the variation of the velocity of a group of joints using this as the baseline for computing the change in motion by maximizing the motion flux.

\section{Preliminaries}\label{sec:prim_prelim}

\noindent
\begin{figure}[tp!]
\centering
\includegraphics[height=7cm, width =12.cm]{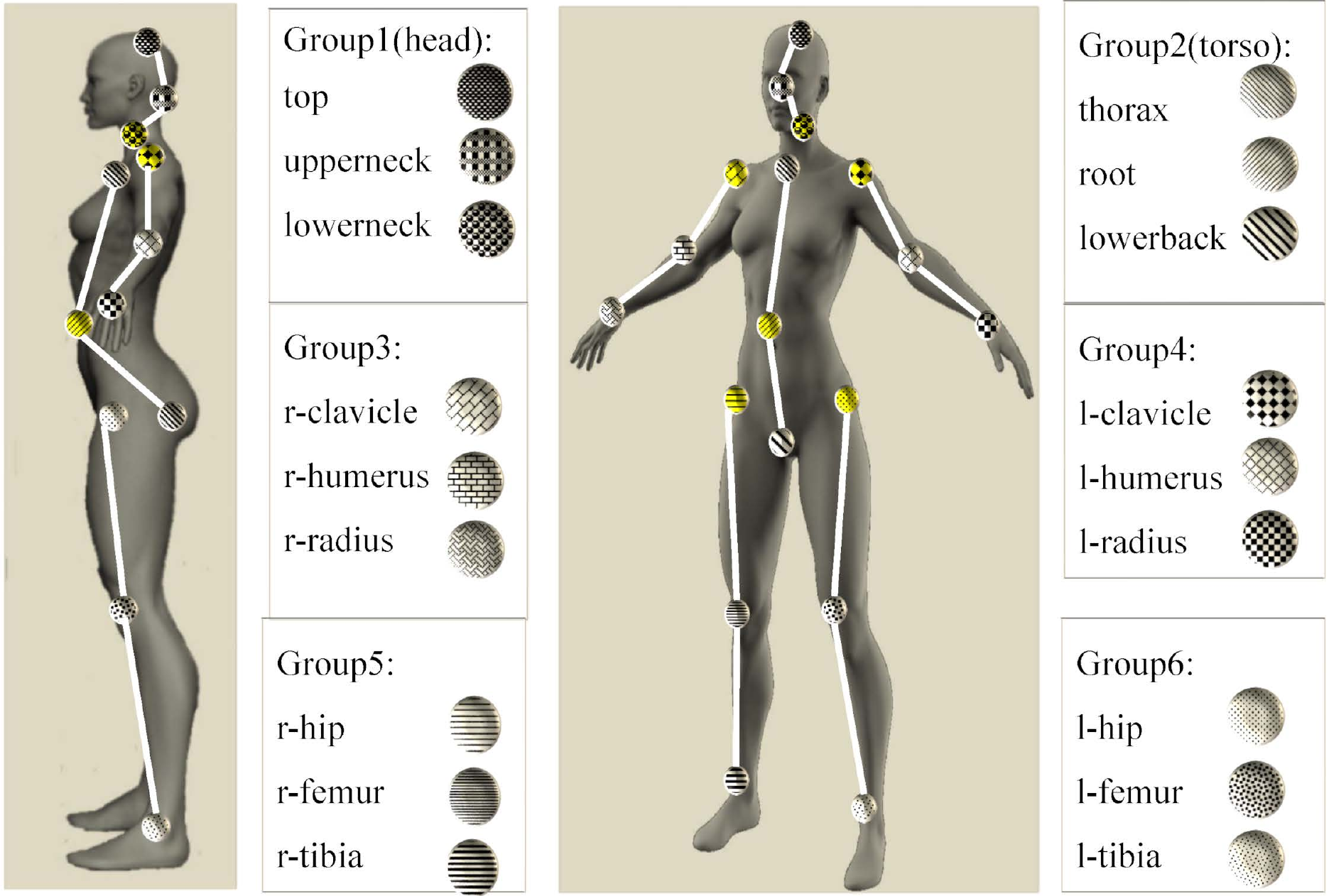}
\caption{The six groups partitioning the human body with respect to motion primitives are shown, together with the joints specifying each group and the skeleton hierarchy inside each group: joints in yellow are the {\em parent joints} in the skeleton hierarchy. 
}\label{fig:groups}
\end{figure}

The 3D pose of a subject, as she appears in each frame of a video presenting a human activity, is inferred according to the method introduced in \cite{sanzari2016}.    Other methods for inferring the 3D pose of a subject are available, we refer in  particular, to the method introduced by \cite{tome2017}, which improves   \cite{sanzari2016} in accuracy.

 3D pose data for a single subject  are  given by the joints configuration.  Joints are associated with the subject skeleton as shown in Fig. \ref{fig:groups} and are expressed via  transformation matrices ${\mathcal T}$ in  $SE(3)$:
\begin{equation} \label{eq:SE3}
  {\mathcal T} = \begin{bmatrix}
    R & {\bf d} \\[0.3em]
    {\bf 0}^{1 \times 3} & 1
  \end{bmatrix}
\end{equation}
Here $R\in SO(3)$ is the rotation matrix, and ${\bf d}\in\mathbb{R}^3$ is the translation vector.  
${\mathcal T}\in SE(3)$ has $6$ DOF and it is used to describe the pose of the moving body with respect to the world inertial frame.  $SO(3)$ and $SE(3)$ are Lie groups and their  identity elements  are  the $3\times3$ and $4\times4$ identity matrices, respectively. 
 We consider an ordered list ${\mathcal J}=\{j_{1}^1,j_{2}^1,\ldots,j_{K-1}^m,j_{K}^m\}$ of $K=18$ joints forming the skeleton hierarchy, as shown in Fig. \ref{fig:groups}, with $m = 1,\ldots, 6$ being the groups each joint belongs to. The 6 groups $G_1,\ldots, G_6$ we consider here correspond to head,  torso, right and left arm, right and left leg. 

 Each  joint $j_{i}^m$, $i=1,\ldots,18$, belonging to a group $G_m$, $m=1,\ldots, 6$, has one parent joint $j_{i}^{m,\star}$, which is the  joint of the group closest to the root joint $root = j_{4}^2\in\mathcal{J}$, according to the skeleton hierarchy, namely it is the fourth joint in the ordered list ${\mathcal J}$ and it belongs to the group $G_2$, the torso. Parent joints for each group are illustrated in yellow on the woman body in the left of Fig. \ref{fig:groups}, they are in the order $(j_3^1,j_4^2,j_7^3,j_{10}^4, j_{13}^5,j_{16}^6)$. 

\begin{figure}[thp!]
\captionsetup[subfigure]{justification=centering}
  \centering
   \begin{subfigure}{0.5\textwidth}
   \includegraphics[width=0.9\textwidth]{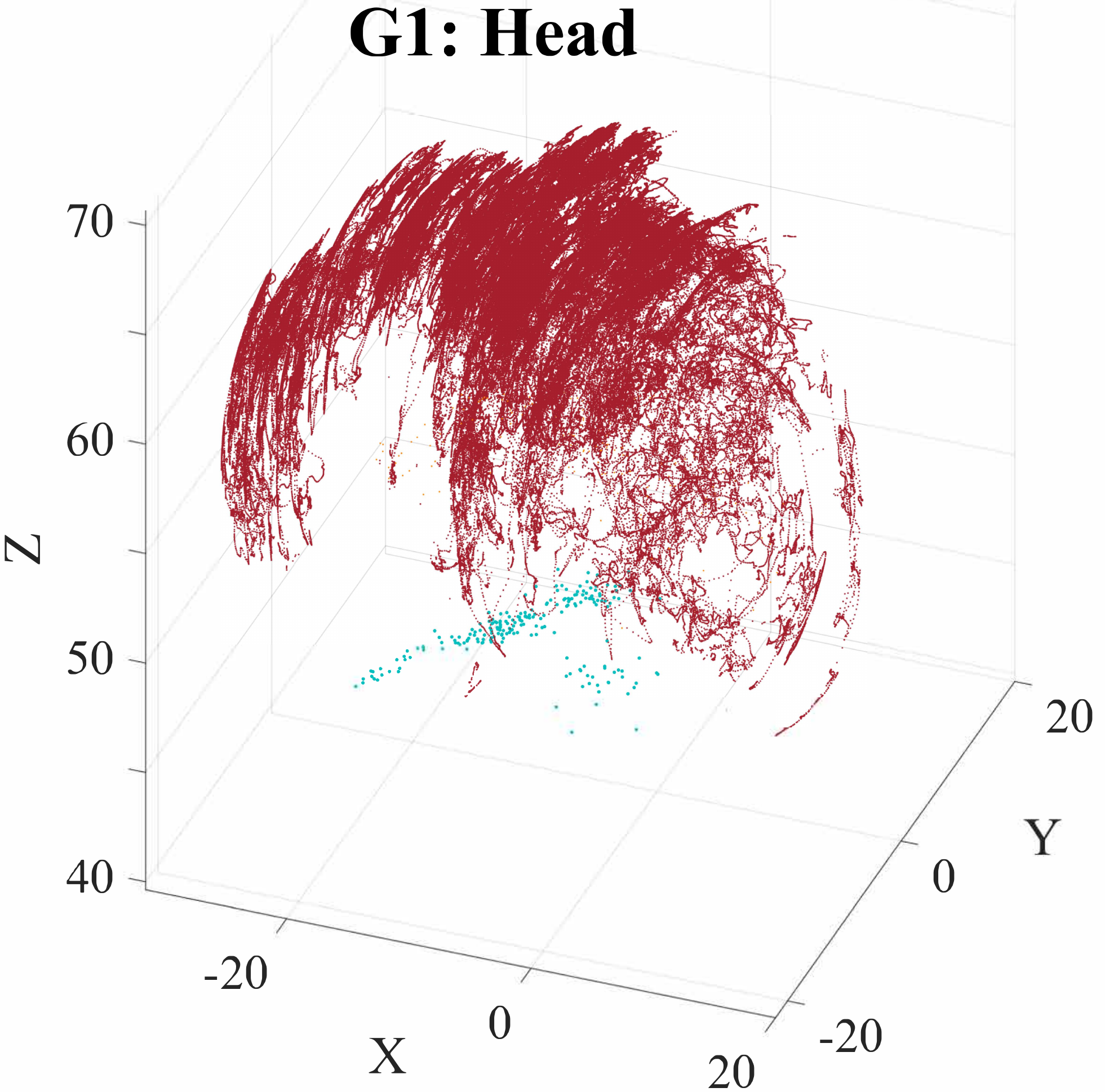}
   \caption{}
   \end{subfigure}%
      \begin{subfigure}{0.5\textwidth}
   \includegraphics[width=0.9\textwidth]{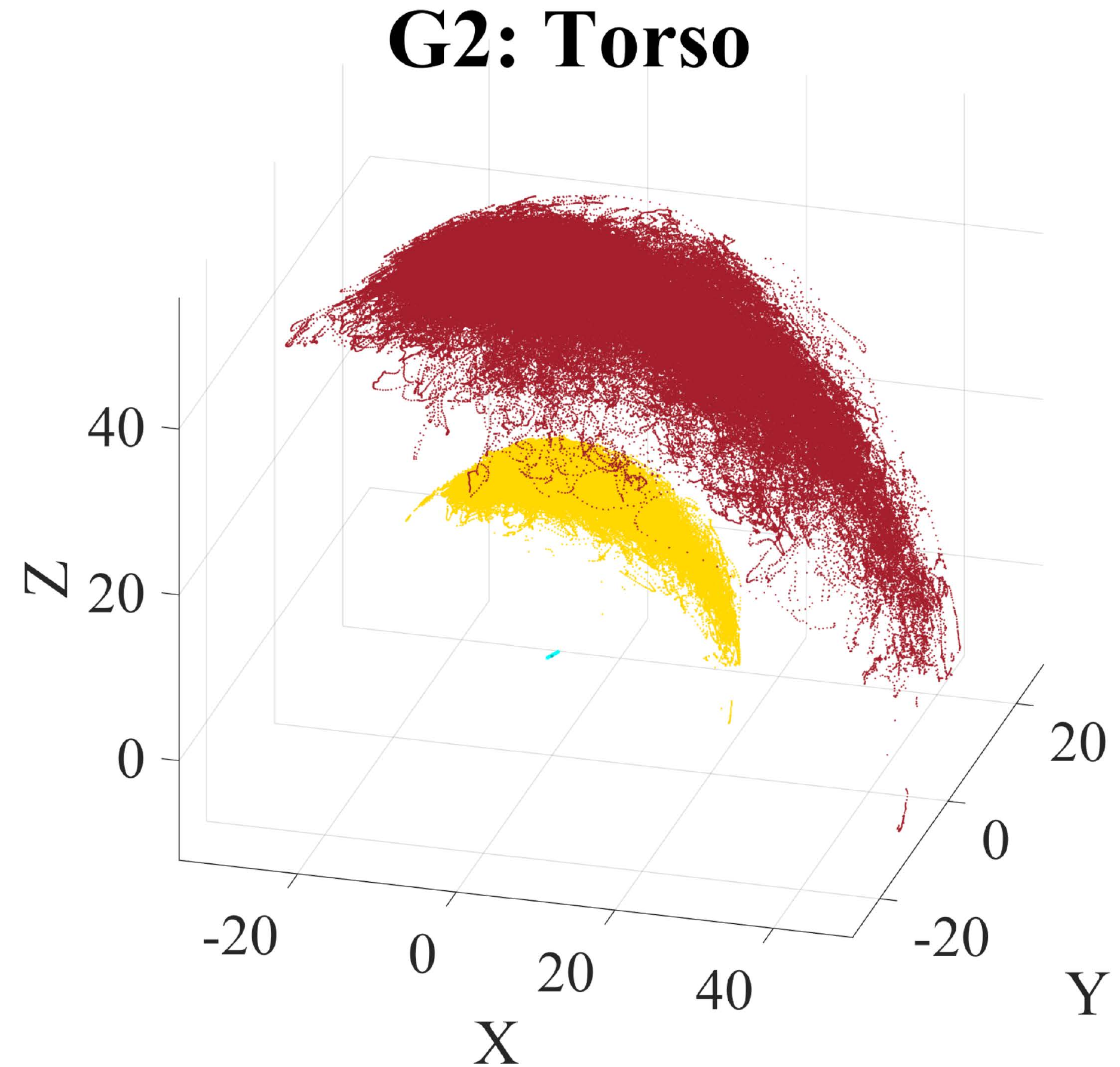}
   \caption{}
   \end{subfigure}
      \begin{subfigure}{0.5\textwidth}
   \includegraphics[width=0.9\textwidth]{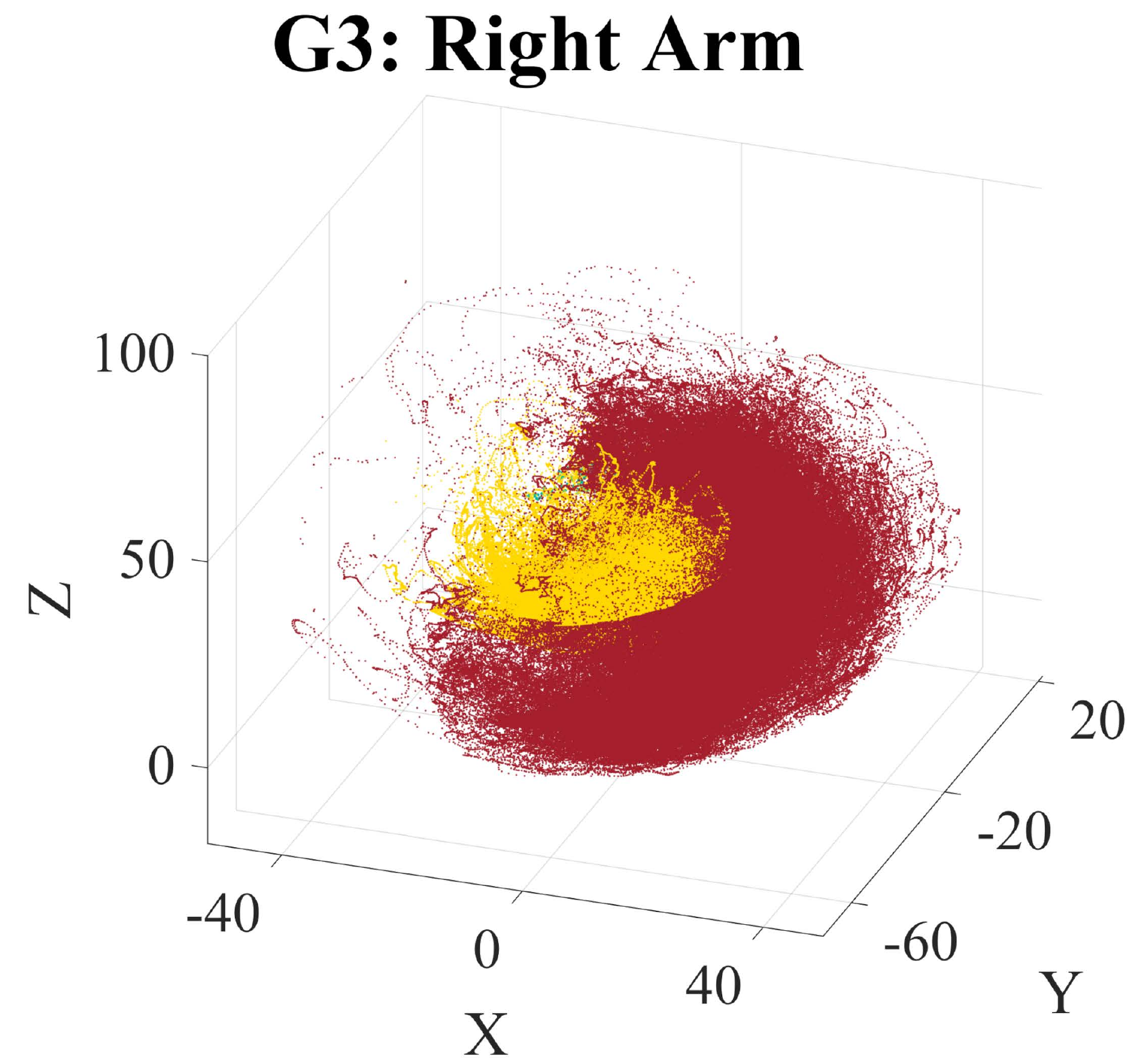}
   \caption{}
   \end{subfigure}%
      \begin{subfigure}{0.5\textwidth}
   \includegraphics[width=0.9\textwidth]{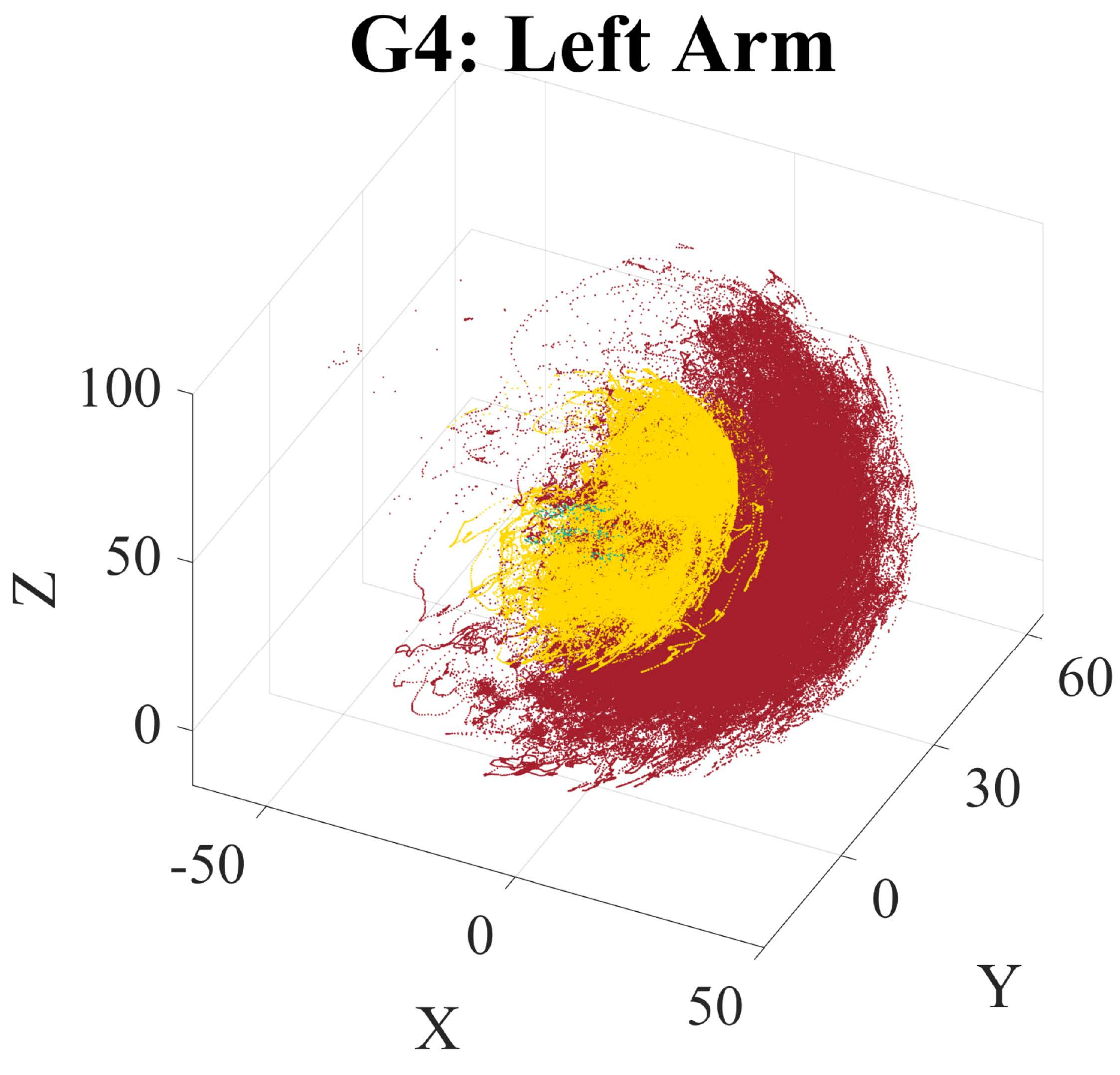}
   \caption{}
   \end{subfigure}
        \begin{subfigure}{0.5\textwidth}
   \includegraphics[width=0.9\textwidth]{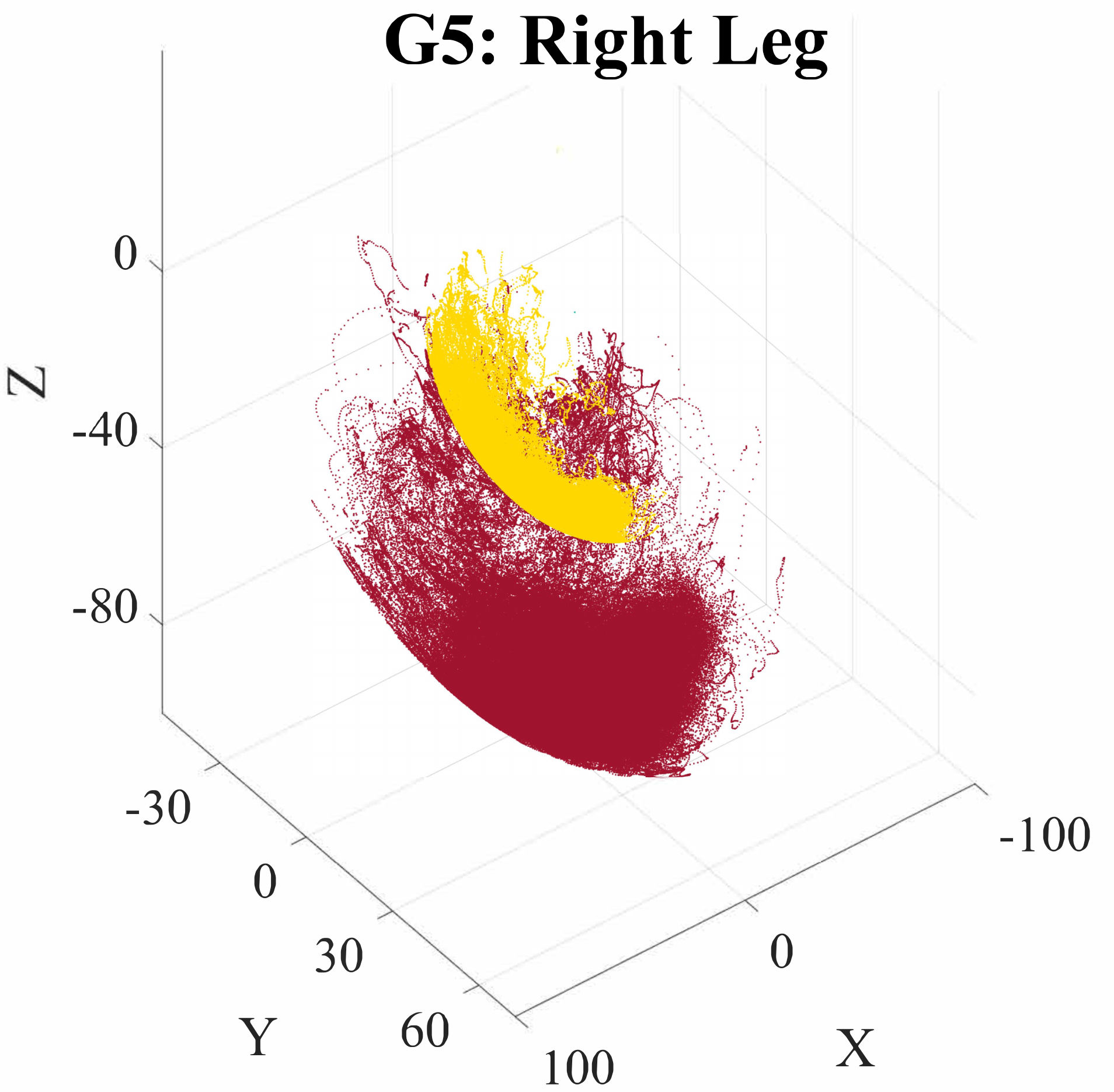}
   \caption{}
   \end{subfigure}%
      \begin{subfigure}{0.5\textwidth}
   \includegraphics[width=0.9\textwidth]{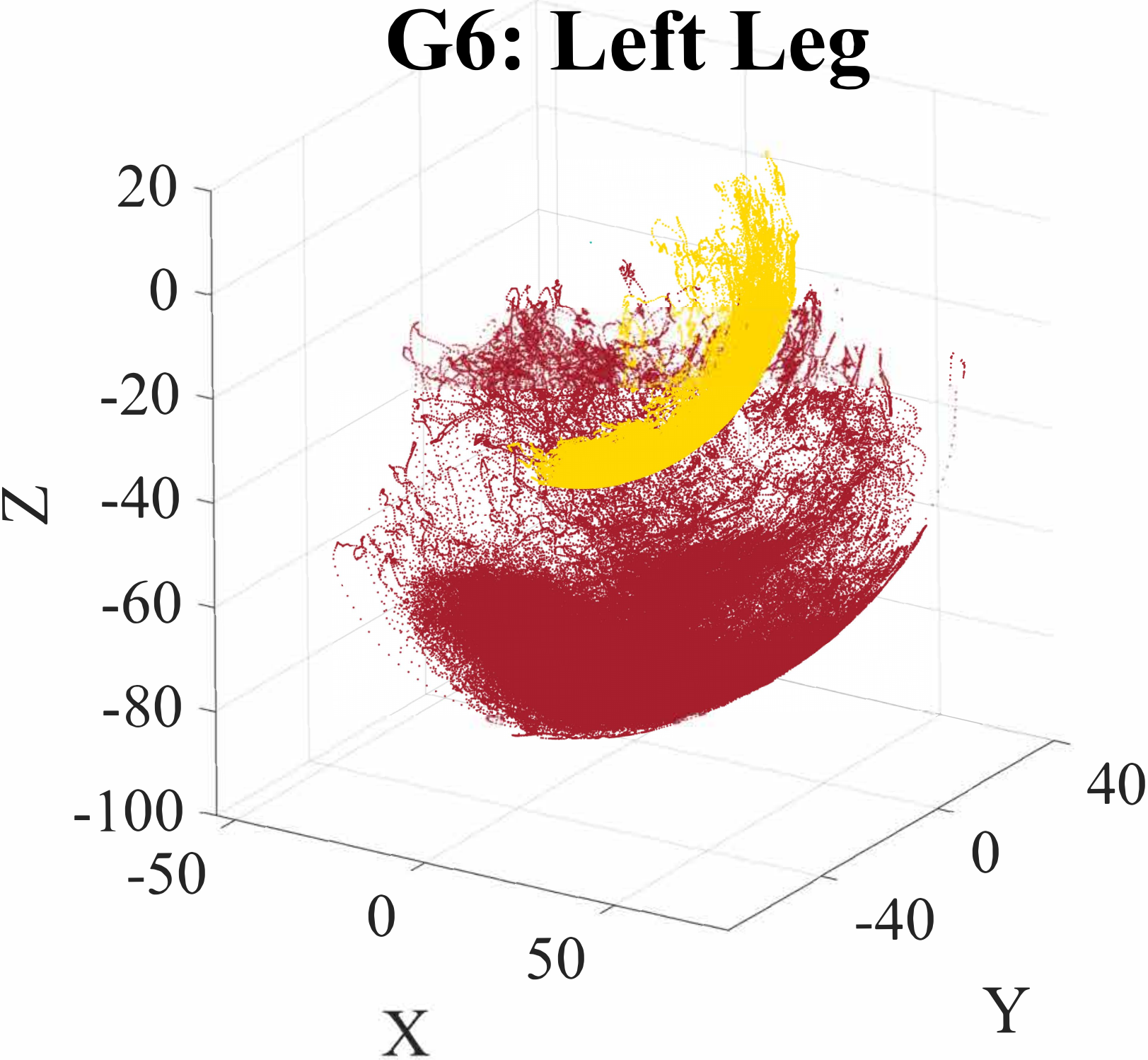}
   \caption{}
   \end{subfigure}
\caption{Sequences of joint positions, for each skeleton group, after the {\em root-sequence} normalization described in Section \ref{sec:prim_prelim}. Position data are in cm. The green points show the most internal group joint data (e.g. the hip for the leg); the yellow points show the intermediate group joint data (e.g. the knee for the leg); the red points show the most external group joint data (e.g. the ankle for the leg). The joints data are collected from the datasets described in Section \ref{sec:prim_results}.}\label{fig:jointlimits}
\end{figure}

A MoCap sequence  of length $N$ is formed by a sequence of  frames of poses.  Each frame of poses is defined by a set of transformations $\{{\mathcal T}^k_{i,m}{\in}SE(3): k = 1,\ldots,N, m=1,\ldots,6\}$ involving all joints $j_i^m\in {\mathcal J}$, $i =1,\ldots,18$, according to the skeleton hierarchy. Given a MoCap sequence  of length $N$, for each frame $k$ the pose of each joint is {\em root-sequence} normalized,  to ensure pose invariance with respect to a common  reference system of the whole skeleton.  Let ${\mathcal T}_{i,m}^k$ be the  pose of the joint $j_{i}^m$, according to the skeleton hierarchy, at frame $k$ in the sequence, and let $j_i^{m,\star}$ be the parent node of $j_i^m$, then  the {\em root-sequence} normalization is defined as follows:
\begin{equation}\label{eq:1}
\mathcal{\hat{T}}_{i,m}^{k} = \left((\mathcal{T}^1_{root,2})^{-1}\mathcal{T}^1_{j_i^{m,\star},m}\right)\left((\mathcal{T}^k_{j_i^{m,\star},m})^{-1}\mathcal{T}^{k}_{i,m}\right).
\end{equation}
Here $(\mathcal{T}_{root,2})$ is the transformation of the root node, which is the joint $j_{4}^2$ belonging to the group $G_2$, the torso.
Equation (\ref{eq:1}) says that the pose $\mathcal{T}_{i,m}^{k}$ of joint $j_{i}^m \in G_m$ at frame $k$ is {\em root-sequence} normalized if obtained by a sequence of transformations seeing first a transformation with respect to its parent node $(\mathcal{T}^k_{j_i^{m,\star},m})^{-1}$, at frame $k$, and then with respect to the transformation of the parent node with respect to the root node, taken at the initial frame of the sequence. 
In Fig. \ref{fig:jointlimits} are shown joints position data for each skeleton group after {\em sequence-root} normalization for all sequences in the dataset. More details on the skeleton structure and its transformations can be found in \cite{natola2015,sanzari2016}.

\section{Motion Primitive Discovery}\label{sec:flux}
 \begin{figure*}
 \centering
\includegraphics[height=20cm]{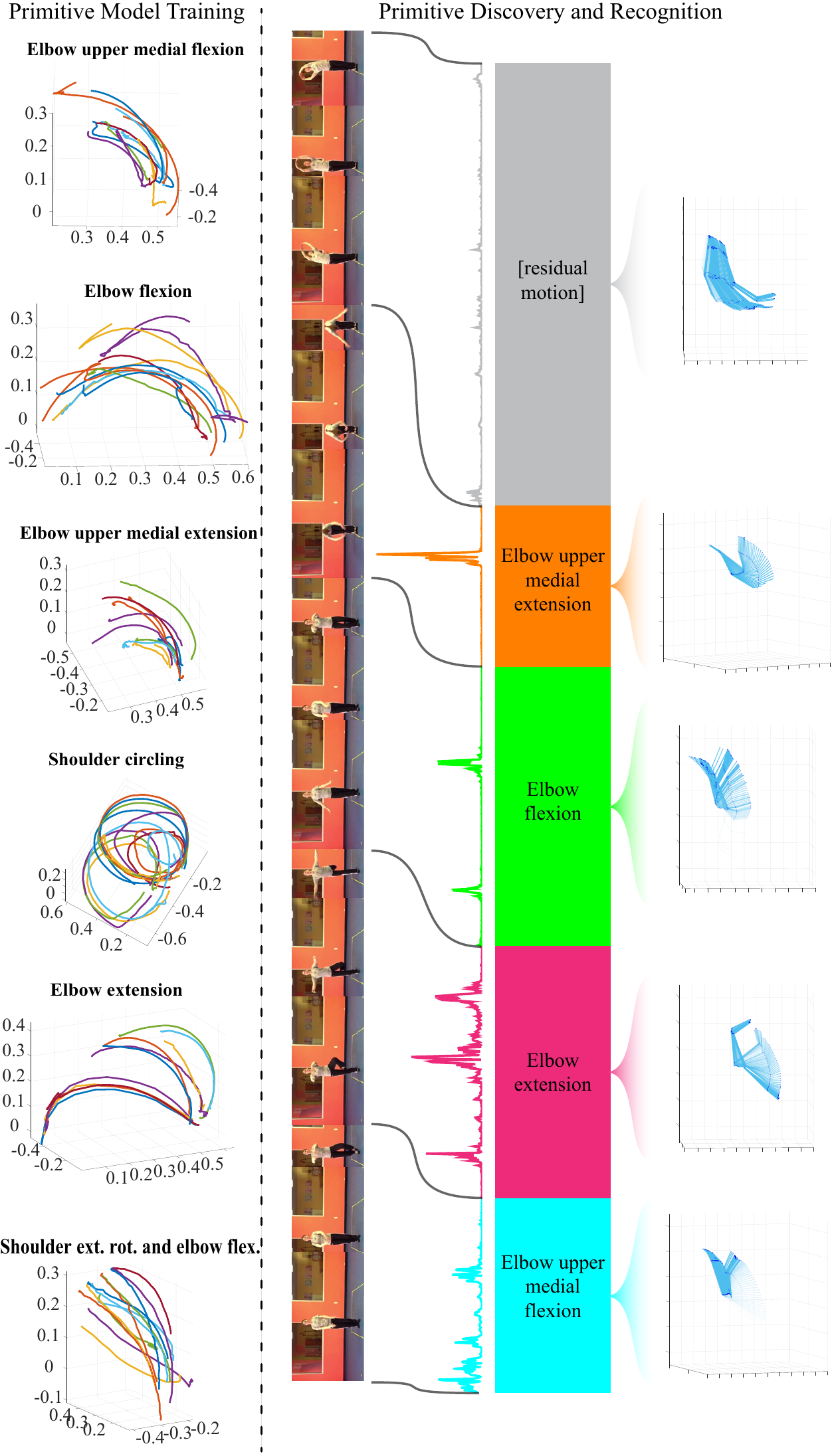}
\caption{Overview of motion primitive discovery and recognition framework. The top section shows primitives of the group `Arm' from six different categories. Primitives are discovered by maximizing the \emph{motion flux} energy function, presented here above the colored bar, though deprived of velocity and length components. These sets of primitives are used to train the hierarchical models for each category. Primitives are then recognized according to the learned models. The recognized motion primitive categories are depicted with different colors. At the bottom, the group motion in the corresponding interval is shown.}\label{fig:segmentation}
\end{figure*}

 We are considering now the problem of discovering motion primitives within a motion sequence displaying an activity in a video. 
 We begin by providing the definition of a joint trajectory on which the temporal analysis is performed.

 \begin{definition}[Joint Trajectory]
 The trajectory of a joint $j$ is given by the path followed by the skeletal joint $j$ in a given interval of time $I = [t_1,t_2]$. Formally:
\begin{equation}\label{eq:traj}
\xi_j : I\subset\mathbb{R} \mapsto \mathbb{R}^3,
\end{equation} 
\end{definition}

Based on the definition above, motion primitives correspond to segments of the joint trajectories of a group $G$. We identify motion primitives as trajectory segments where the variation of the velocity of the joints is maximal and where the endpoints of the segment correspond to stationary poses of the subject \cite{Marr501}. 
 
\noindent
{\bf Preprocessing} 
 To overcome problems related to the finite sampling frequency of the poses in the data, we compute smooth versions of the joint trajectories by cubic spline interpolation.  
 This interpolation provides a continuous-time trajectory for all the joints of the group with smooth velocity and continuous acceleration, satisfying natural constraints of human motion.

\noindent
{\bf Motion Flux} The motion flux captures the variation of the velocity of a group with respect to its rest pose. The total variation of the joint group velocity is evaluated along a direction $\normvec$ that corresponds to stationary poses of the group. For groups 1, 3 and 4 this direction is defined by the segment connecting the `lowerneck' and `upperneck' joints while for groups 2, 5 and 6 by the segment connecting the `root' with the `lowerback' joints.

\begin{definition}[Motion Flux]
Let $G=\{j_1,\ldots,j_K\}$ be a group consisting of $K$ joints and $\velf_{j}$ the velocity of joint $j\in G$. The \emph{motion flux} with respect to the time interval $I=[t_1,t_2]$ is defined as
\begin{equation}\label{eq:flux}
\Phi(t_2,t_1) \doteq \sum_{j\in G}{\int_{t_1}^{t_2} \left\lvert \dot\velf_{j}(t)\cdot\normvec\right\rvert \ dt}.
\end{equation} 
\end{definition}

\noindent
{\bf Discovery}  In order to discover a motion primitive, we identify a time interval between two time instances (endpoints) where the group velocity is minimal while the motion flux within the interval is maximal. This is done by performing an optimization based on the motion flux of a group $G$, as defined in eq. (\ref{eq:flux}).
 More specifically, the time interval of a motion primitive is identified by maximizing the following energy-like function:
\begin{equation}\label{eq:prim}
P(\rho;t_0)\doteq\  \Phi(\rho,t_0) - \frac{\beta_v}{2}\sum_{j\in G} \left(\norm{\velf_{j}(\rho)}^{2}{+} \norm{\velf_{j}(t_0)}^{2}\right) +  \beta_s  \sum_{j\in G} (s_j(\rho)-s_j(t_0)),
\end{equation}
where $s_j(t) {=} \int_{0}^{t} \|\dot\xi_{j}(\tau)\|d\tau$ is the arc length function of $\xi_{j}$. 
 The last term of eq. (\ref{eq:prim}) is a regularizer based on the length of the trajectory segment, introduced in order to avoid excessively long primitives.
 The hyper-parameter $\beta_v$ acts as penalizer associated to the soft-constraint on the stationarity of the poses at the start and end of the primitive, while $\beta_s$ controls the strength of the regularization on the primitive length. Both  $\beta_v$ and $\beta_s$ depend on the scaling of the data and the sampling rate of the joint trajectories. 
 
 Given a starting time instant $t_0$, a motion primitive is extracted by identifying the time instant $\rho$, which corresponds to a local maximum of (\ref{eq:prim}). The optimality condition of (\ref{eq:prim}) gives:
\begin{equation} \label{eq:prim_opt}
\sum_{j\in G}\left(\left\lvert \dot\velf_{j}(\rho)\cdot\normvec\right\rvert - \beta_v \frac{\dot\velf_j(\rho)}{ \norm{\velf_{j}(\rho)}} - \beta_s  \|\dot\xi_{j}(\rho)\| \right){=} 0.
\end{equation}
 Given the one-dimensional nature of the problem, finding the zeros of (\ref{eq:prim_opt}) and verifying whether they correspond to local maxima of (\ref{eq:prim}) is trivial. 

 Based on the previous we provide a formal definition of a motion primitive.
\begin{definition}[Motion Primitive]
 A motion primitive of a group of joints $G$ is defined by the trajectory segments of all joints $j\in G$ corresponding to a common temporal interval $I=[t_{start},t_{end}]\subset\mathbb{R}$ such that $P(t_{start};t_{end})> P(\rho;t_{start})\ \forall \rho\in(t_{start},t_{end})$. Namely
\begin{equation}
 \gamma_G^I = \{\xi_{j_1}(t),\ldots,\xi_{j_K}(t)\}\ \mbox{for}\ t{\in} [t_{start},t_{end}].
 \end{equation}
\end{definition}

\noindent
{\bf Primitive discovery in an activity}
 A set of primitives is extracted from an entire sequence of an activity $\varsigma$  by sequentially finding the time instances which maximize (\ref{eq:prim}). 
 
 Let $t_0$ and $t_{seq}$ denote the starting and ending instances of the sequence, respectively. Let also
 \begin{equation}
 t_{n}=\underset{\rho\in[t_{n-1},t_{seq}]}{\arg\max}P(\rho;t_{n-1}),
\end{equation} 
 and $\mathcal{I}_{\varsigma} = \{[t_{n-1},t_{n}]\,|\, n\in\mathbb{N}\ \mbox{and}\ t_n\leq t_{seq}\}$ the set of time intervals defining successive motion primitives in the sequence. 
 The set of motion primitives discovered in the entire sequence $\varsigma$ is given by
\begin{equation}
 {\Gamma}^{\varsigma}_G = \{\gamma_G^I\, |\, I\in\mathcal{I}_{\varsigma} \}.
\end{equation}
  
\begin{figure}[t!]
  \centering
   \includegraphics[width=.95\textwidth]{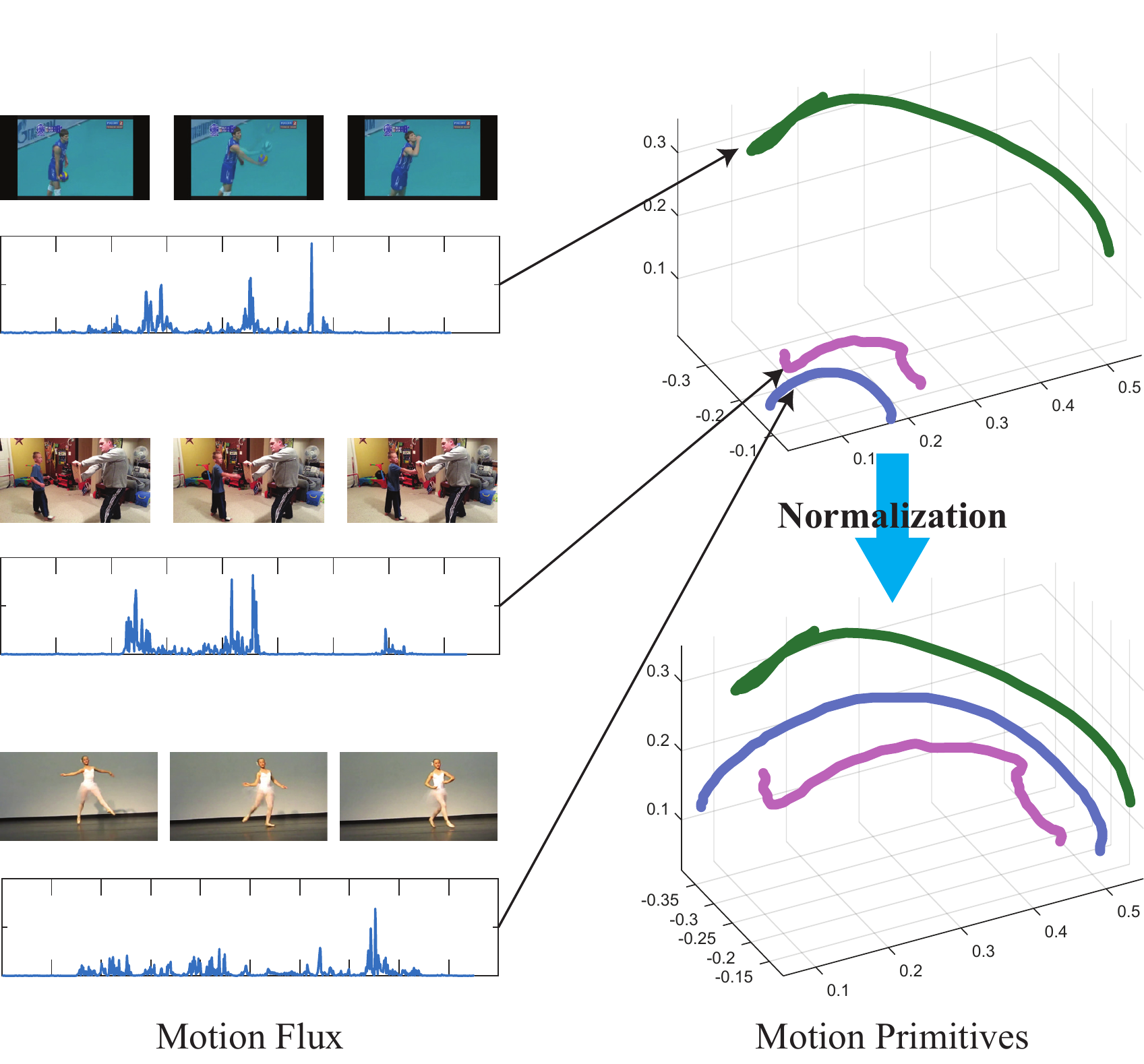}\hfill
\caption{\textbf{Left:} Motion flux of three motion primitives of group $G3$ labeled as `Elbow Flexion', discovered from video sequences taken from the ActivityNet dataset. \textbf{Right:} Motion primitives before and after the normalization, for clarity only the curve of the out most joint is shown.(Best seen on screen, zoomed-in)}\label{fig:fluxcurv}
\end{figure}

 As noted in the introduction, and also shown in Figure \ref{fig:fluxcurv}, there is a significant motion variation  across subjects, activities and  sampling rates. 
  For example, for the upper limbs it is known that the range of motion  varies from person to person and is influenced by gait speed \cite{de2014}. This is in turn influenced by the specific task, and determining ranges of motion is still a research topic \cite{gates2016} (for a review on range of motions for upper limbs, see \cite{de2014}). 
  This makes analysis and recognition of motion primitives taken from different datasets, activities and subjects problematic. To induce invariance with respect to these factors we apply anatomical normalization.

 More specifically, the main source of variation of the primitives is due to the anatomical differences among the subjects. To remove the influence of these differences on the primitives we consider a scaling factor $k_G$ based on the length $\ell_G$ of the limb defined by group $G$, namely $k_G=1/\ell_G$. Hence, given a primitive $\gamma_G^I$ we scale the trajectory of each joint by the constant $k_G$.    
 By applying the anatomical normalization to the entire collection of motion primitives for group $G$ discovered across all sequences of a dataset $\mathcal{D}$ we obtain the set of motion primitive of the group, namely
\begin{equation}\label{eq:allprimitivesG}
  {\Gamma}_G = \{\Gamma^{\varsigma}_G\, |\, \varsigma\in\mathcal{D} \}.
\end{equation}

In Section~\ref{sec:prim_results} we provide a quantitative evaluation of the normalization effectiveness, together with a comparison with additional normalization candidates.

\section{Motion Primitive Recognition}\label{sec:features}
In the previous section we have shown that for each group of joints $G_m, m=1,\ldots,6$, the  motion flux obtains the interval $I=[t_{start},t_{end}]$  matching the joint trajectory of a sequence in so determining  a primitive as a path $\gamma_{G_m}^I: I\subset {\mathbb R}\mapsto {\mathbb R}^9$, given a video sequence of a  human activity. Here  ${\mathbb R}^9$ is due to the path being related to the 3 joints of each group $G_m$,  as indicated in Fig. \ref{fig:groups}.  We have also seen that the path is normalized  by the link length of a limb, to limit variations due to bodies dissimilarities.
For clarity from now on we shall denote each primitive with $\gamma$ unless the context requires to add superscripts and subscripts, and in general subscripts and superscripts are local to this section, also we shall refer to  the group a primitive or trajectory belongs to both with $G_m$ and more in general with $G$.

 We expect that the following facts will be true of the discovered motion primitives:
\begin{enumerate}
\item Each primitive of motion is  independent of the gender, (adult) age, and body structure,  under  normalization.
\item Each primitive of motion can be characterized independently of the specific activity, hence the same primitive can occur in several activities (see Section \ref{sec:prim_results} for a distribution of discovered primitives  in a set of activities).
\item The motion flux  ensures that each unknown segmented primitive belongs to a class such that: the number of classes is finite and the set of classes can be mapped onto a subset of motion primitives defined in biomechanics (see e.g table 1.1 of \cite{hamill2006}).
\end{enumerate}

To show experimentally the above results we shall introduce a hierarchical classification. The hierarchical classification first partitions  the primitives of each group into classes. Once the classes are generated a class representative is chosen and inspected  to assign a label to the class. 
We show that the classes correspond to a significant subset of the motion primitives defined in biomechanics, thus ensuring a proper partition. Each class is then further partitioned into subclasses  to comply with the inner diversification of each class of primitives. This last classification is further used for recognition of unknown discovered primitives. 

Primitive recognition is  used to both test experimentally  the three above results of the introduced motion flux method and for  applications where discovering and recognition of primitives of human motion is relevant (see for example \cite{abernethy2013}).

\begin{figure}[t!]
\captionsetup[subfigure]{justification=centering}
  \centering
   \begin{subfigure}{0.3\textwidth}
   \includegraphics[width=0.9\textwidth]{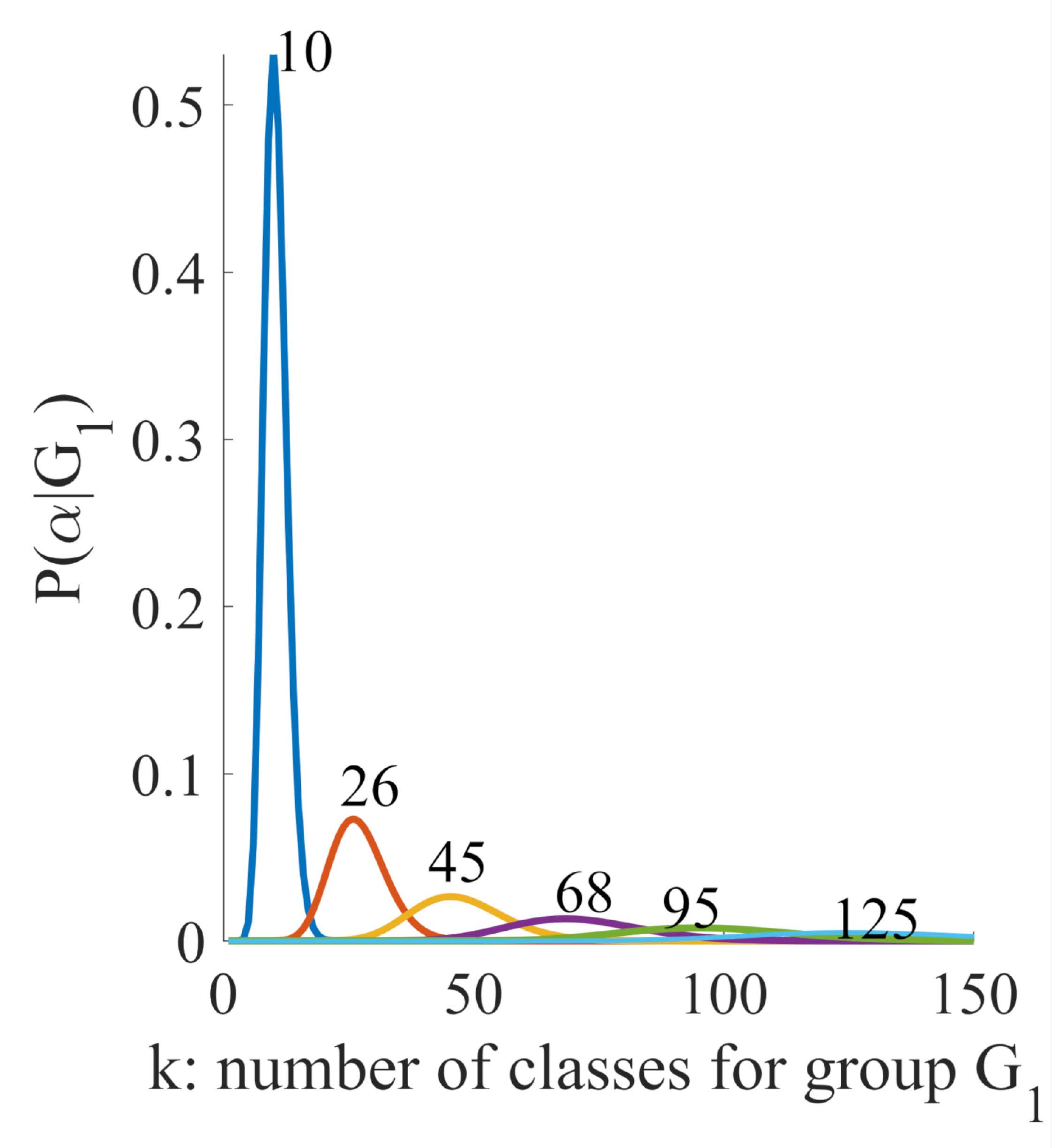}
   \caption{}
   \end{subfigure}%
      \begin{subfigure}{0.32\textwidth}
   \includegraphics[width=0.9\textwidth]{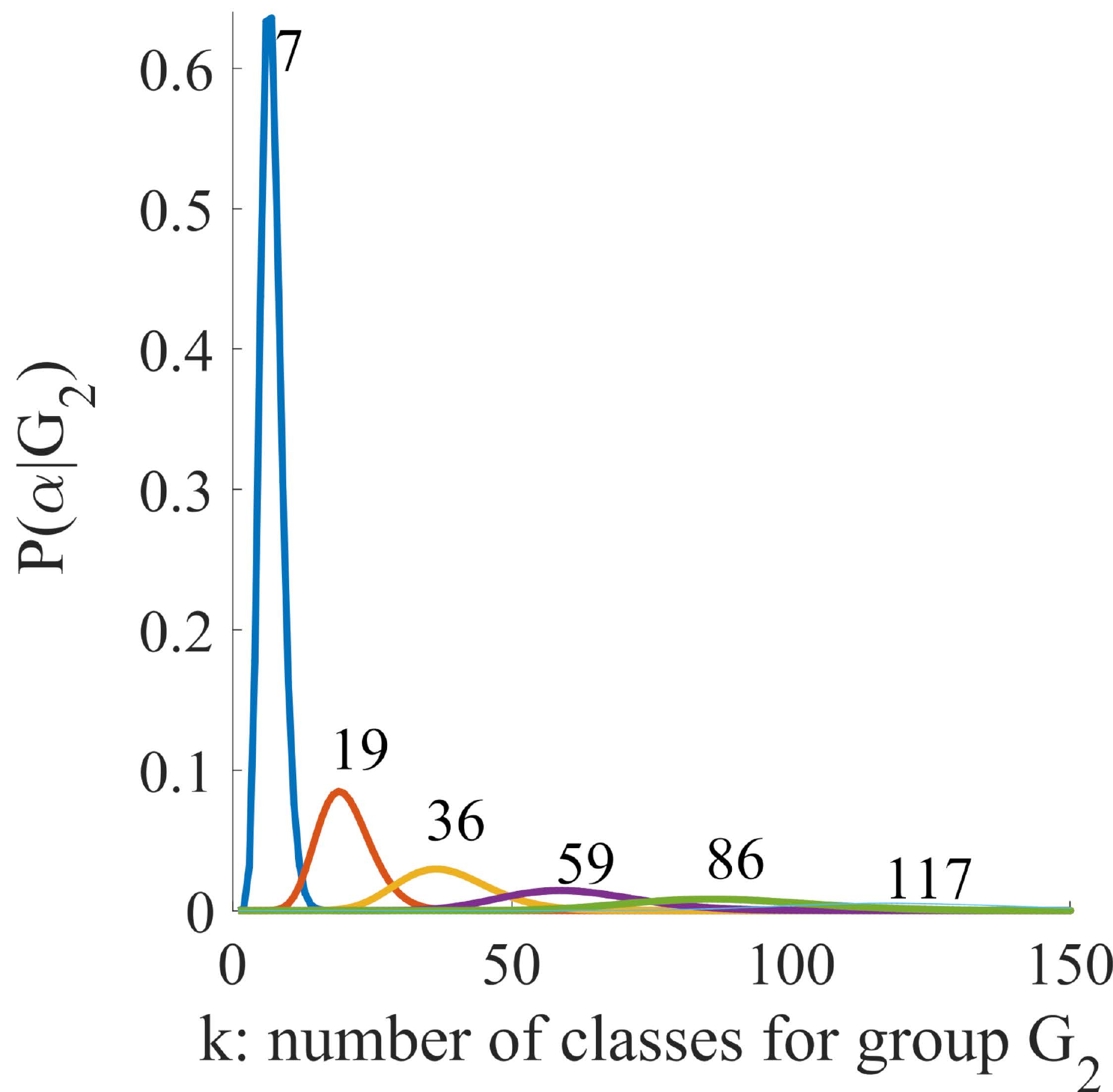}
   \caption{}
   \end{subfigure}
\begin{subfigure}{0.33\textwidth}
   \includegraphics[width=0.9\textwidth]{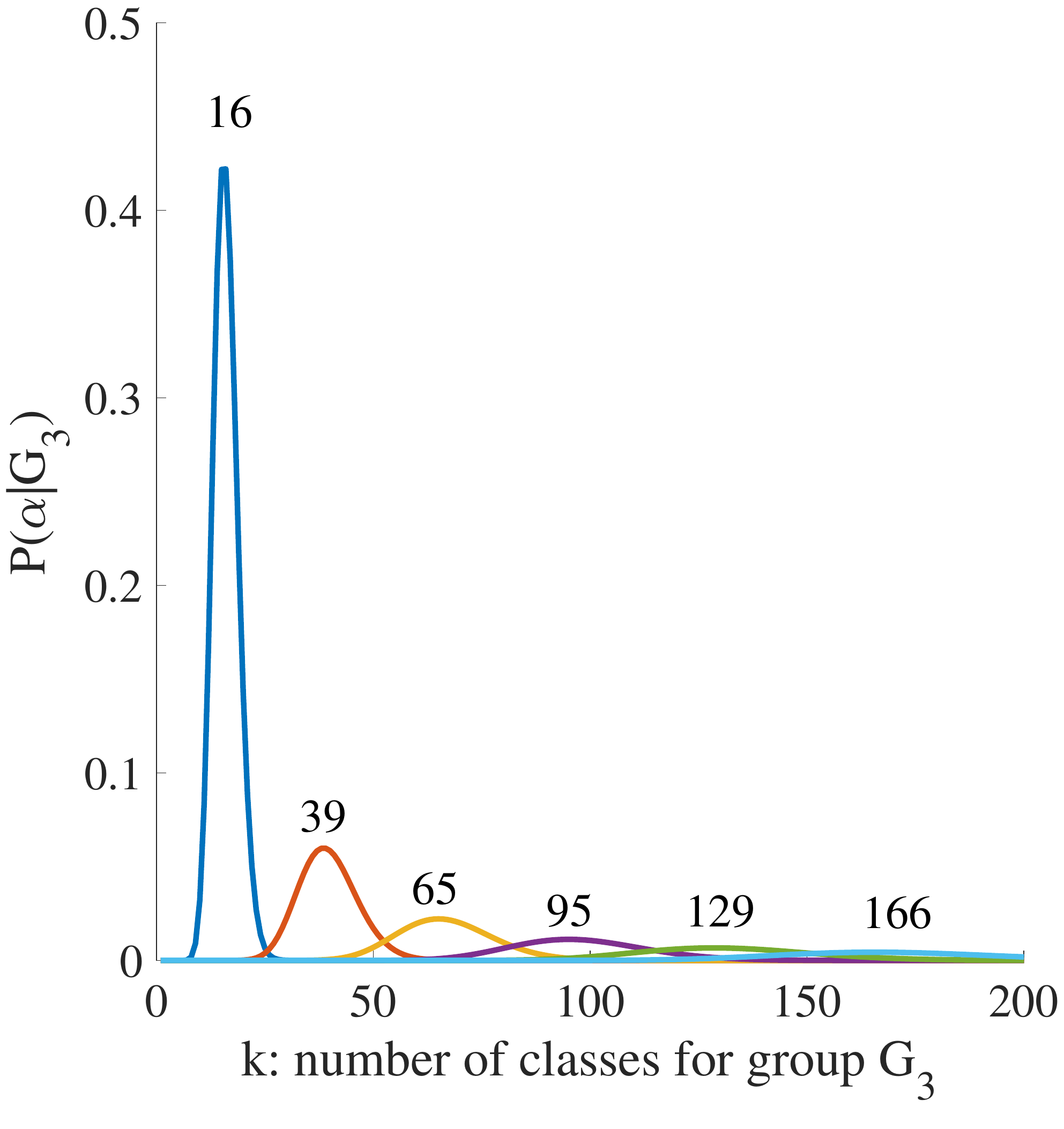}
   \caption{}
   \end{subfigure}
\caption{Number $k$ of components for groups $G_1,G_2$ and $G_3$. Values of $k$ are computed adjusting $\alpha$ so as to maximize the posterior $p(\alpha, G_m)$, given the data, namely the sampled primitives in the groups.}\label{fig:alfaV}
\end{figure}

\subsection{Solving primitive classes}

We describe in the following the  method leading to the generation of all the primitive classes illustrated in Fig \ref{fig:primitives}. 

We consider   three  MoCap datasets \cite{mandery2015,ionescu2014,CMU}  guaranteeing  the ground truth for  the human pose and segment the activities according to the motion flux method, described in the previous section. 
Let $\Gamma_{G}$ be the set of primitives collected for group $G$ according to equation (\ref{eq:allprimitivesG}). Let $\gamma_{\nu}\in \Gamma_{G}$, $\nu =1,\ldots,S$, with $S$ the number of primitives in $\Gamma_{G}$,  $\gamma_{\nu} = (\xi^{\nu}_{j_1},\xi^{\nu}_{j_2},\xi^{\nu}_{j_3})$ is formed by the trajectories of the joints in $G$. Out of these trajectories we choose the one of the most external joint (see Figure \ref{fig:groups}) that we indicate with $\xi_{E}^{\nu}$. We order these trajectories, each designating a primitive in group $G$, with an enumeration $\langle \Gamma_{G}\backslash {\xi_E}\rangle_{\nu=1}^S$, $S$ the number of discovered primitives for group $G$.  Note that we can arbitrarily  enumerate the primitives of a group, restricted to a single joint,  though they are unlabeled and unknown, and this is what the first model should solve.

At this step, model generation amounts to  find the classes of primitives for each group $G$, taking the trajectories  $\xi_{E}^{\nu}$ in the enumeration $\langle \Gamma_{G}\backslash {\xi_E}\rangle_{\nu=1}^S$  as observations.

\noindent
\paragraph{Feature Vectors}
Given a trajectory $\xi_E^{\nu}$, with $\nu$ the index in the enumeration $\langle \Gamma_{G}\backslash {\xi_E}\rangle_{\nu=1}^S$,  a  feature vector is obtained by  first computing curvature $\kappa(s(t))$ and torsion $\tau(s(t))$ on the trajectory $\xi_{E}^{\nu}$, where $s(t)$ indicates the arc length as already defined in Section \ref{sec:flux} for trajectories. Then we  take three contiguous points $(x_{i-1},y_{i-1},z_{i-1}),\ldots, (x_{i+1},y_{i+1},z_{i+1})$  on the    trajectory $\hat{\xi}^{\nu}_{E}$ decimated by a factor of $5$ \cite{alt2000}, keeping the curvature and torsion of the sampled points, after decimation.   We choose curvature and torsion  as they suffice to specify a 3D curve up to a rigid transformation.   The formed feature vector  is indicated by ${\mathcal F}_i$, where the index $i$ is the index of the middle point $(x_{i},y_{i},z_{i})$,   it is of  size $17 \times 1$ and it is defined as follows: 

\begin{figure}[h!]
  \centering
 \includegraphics[width=1\textwidth]{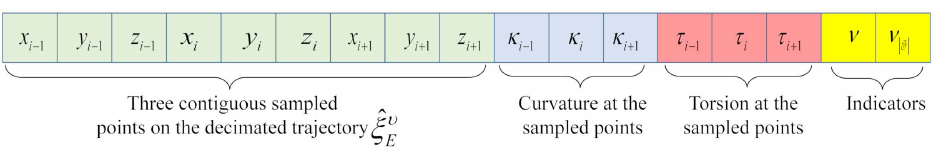}
\caption{Transposed feature vector of 3 contiguous sampled points on the decimated trajectory.}\label{fig:fv}
\end{figure}

The last two elements  $\nu,\nu_{|{\mathcal F}_i|}\in{\mathbb R}$ of ${\mathcal F}_i$ are  indicators.  Namely, the indicator $\nu$  is the index,  in the enumeration $\langle \Gamma_{G}\backslash {\xi_E}\rangle_{\nu=1}^S$, identifying the trajectory
  the 3 points  belong to, the three points are the first 6 element of the feature vector. On the other hand, the indicator $\nu_{|{\mathcal F_i}|}$ specifies the number of features vectors the decimated trajectory $\hat{\xi}^{\nu}_{E}$  is decomposed into, here $|\cdot|$ indicates the cardinality; These two indicators, allow to recover the path a feature vector belongs to, and are normalized and denormalized  as follows. Let ${\mathbb F}^{\xi}_G$ be the set of all feature vectors for the trajectories in $\langle \Gamma_{G}\backslash {\xi_E}\rangle_{\nu=1}^S$, and let their number be $W$. Accordingly,  let $\nu_{|{\mathcal F}|} = (\nu_{|{\mathcal F_1}|},\ldots,\nu_{|{\mathcal F_W}|})$, then the normalization and denormalization for the element $\nu_{|{\mathcal F}_i|}$ (and similarly for $\nu$) is defined as follows, with $g$ indicating the  denormalization:
\begin{equation}
\begin{array}{ll}
\hat{\nu}_{|{\mathcal F}_i|} = \displaystyle{\frac{\nu_{|{\mathcal F}_i|}-min(\nu_{|{\mathcal F}|})}{max(\nu_{|{\mathcal F}|})-min(\nu_{|{\mathcal F}|})}}\\
g(\hat{\nu}_{|{\mathcal F}_i|}) = \hat{\nu}_{|{\mathcal F}_i|}(max(\nu_{|{\mathcal F}|})-min(\nu_{|{\mathcal F}|}))+min(\nu_{|{\mathcal F}|})
\end{array}
\end{equation}

\paragraph{Generation of the primitives classes}
 Given the feature vectors for each trajectory in the enumeration $\langle \Gamma_{G}\backslash {\xi_E}\rangle_{\nu=1}^S$, the goal is to cluster them and return a cluster for each class of primitives. 
Since we do not even know the number of classes the primitives should be partitioned into, a  good generative model to approximate the distribution of the observations is the Dirichlet process mixture (DPM) \cite{ferguson-1973,antoniak-1974}. The Dirichlet process assigns  probability measures to the set of measurable partitions of the data space. This induces in the limit a finite mixture since, by the discreteness of the distributions sampled from the process,  parameters have positive probability to take the same value, in so realizing  components of the mixture.   Here we assume that  feature vectors in the data space are realizations of normal distributions with a conjugate prior. Namely the variables have precision priors following the Wishart distribution and location parameters prior following the normal distribution. The Dirichlet mixture model is based on the definition of a Dirichlet process $\Pi(\cdot,\cdot)$ with $\Pi\sim DP(H,\alpha)$ ($D$ being the Dirichlet distribution), where $H$ is the base distribution and $\alpha$ the precision parameter of the process (see \cite{teh2011}).   In the Dirichlet process mixture the value of the precision $\alpha$ of the underlying Dirichlet process influences the number of classes generated by the model. 

For determining the number of classes for each group $G$ we estimate the posterior $P(\alpha | G)$, of the  precision parameter $\alpha$  according to a mixture of two gamma distributions, as described in \cite{west-1992}, choosing the best value. This is a rather complex simulation process since it requires different initializations of the parameters of the gamma distribution for $\alpha$ within the estimation of the parameters of the DPM, for each group $G$. Here the parameters of the DPM are estimated according to \cite{jain-2004}.  Distributions  of $\alpha$ for the groups $G_1,G_2$ and $G_3$, according to different simulation processes, are given in Fig. \ref{fig:alfaV} where the number of components $k$ for the maximum values of each distribution, are indicated.
Finally the DPM returns the parameters of the components (for each group $G$) given the feature vector ${\mathcal F}_i$, as:
\begin{equation}\label{eq:mixture0}
\begin{array}{ll}
\Theta_{G} =\langle k, \{ \Theta_w \ |\ \Theta_w = (\pi_w,\mu_w,\Sigma_w), w =1,\ldots,k\}\rangle, \ k\geq 1. \\
p({\mathcal F}_i|\Theta_{G}) = \sum_{w=1}^k \pi_w {\mathcal N} ({\mathcal F}_i|\mu_w,\Sigma_w). 
\end{array}
\end{equation} 
Note that the number of components $k$ is unknown and estimated by the DPM, hence it is one of the parameters for each group. The parameters  $\mu_w$ and $\Sigma_w$ are the mean vector and covariance matrix of the $w$-th Gaussian component of the mixture, indicated by ${\mathcal N}$, and $\pi_w$ is the $w$-th weight of the mixture, with $\sum_w\pi_w=1$. Hence,  $p({\mathcal F}_i|\Theta_{G})$ is  the probability of the feature vector ${\mathcal F}_i$, given the parameters $\Theta_G$.

We expect that each $\Theta_w\in \Theta_{G}$ indicates the parameters of a component $C_w^G$, collecting primitives of the same type, in group $G$. In other words, we expect that two  feature vectors, say ${\mathcal F}_p, {\mathcal F}_q$, of group $G$, belong to the same component   if   their likelihood are both maximized by the same parameters $\Theta_w\in \Theta_G$.

\paragraph{Assigning primitives to classes}
The classification returns, for each group $G_m$, the number $k$ of components indicated in Fig. \ref{fig:primitives}, say $k=10$  for $G_1,G_5,G_6$, $k=7$  for $G_2$ and $k=16$  for $G_3,G_4$,  also thanks to the specification of the $\alpha$ parameter, as highlighted above (see Fig.  \ref{fig:alfaV}).
Components are formed by features vectors. To retrieve the trajectories  and generate a corresponding class of primitives, ready to be labeled, we use the normalized indicators placed in position 16th and 17th of the feature vector (Fig \ref{fig:fv}) and the  denormalization function $g$. Let $C_w^{G_m}$ be a component of the mixture  of the group $G_m$, identified by parameters $\Theta_w\in \Theta_{G_m}$. Algorithm \ref{algo:algo0} shows how to compute the class of primitives:

\begin{algorithm}%
\caption{Obtaining  classes of primitives from DPM components. Here $|\cdot|$ indicates cardinality.} \label{algo:algo0}%
\KwIn{Component $C_{w}^{G_m}$ of DPM}%
\KwOut{Class ${\mathcal L}_w^{G_m}$ of  primitives}%
{Initialize $U_{\xi_{{E}}}^{\nu}=\emptyset$, $\nu=1,\dots,S$, $S$ number of primitives in $\Gamma_{G_m}$}\\
\ForEach{Feature vector ${\mathcal F}_i$ in $C_w^{G_m}$ }{
     compute $g(\nu)$ and associate it with  the trajectory $\xi_{E}^{\nu}$;\\
     $U_{\xi_{E}}^{\nu} = \{{\mathcal F}_i\}\cup U_{\xi_{{E}}}^{\nu}$\;
     compute $g(\nu_{|\mathcal F|})$,  number of feature vectors the trajectory $\xi_{E}^{\nu}$ is decomposed into;\\
     }    
\If{$|U_{\xi_{E}}^{\nu}| \geq 0.8g(\nu_{|\mathcal F|})$} 
          {find the primitive $\gamma_{\nu}\in \Gamma_{G_m}$ designated by $\xi_{E}^{\nu}$\\
           assign the pair $(\gamma_{\nu},\Theta_w)$ to ${\mathcal L}_{w}^{G_m}$}%
   \Return Class  ${\mathcal L}_{w}^{G_m}$.
\end{algorithm}%

At this point  we have generated the classes ${\mathcal L}_{w}^{G_m}$, $w =1,\ldots,k$, $k\in\{7,10,16\}$ of primitive  for each group $G_m$. To label the classes we proceed as follows. 
Let $p(\gamma_{\nu}|\Theta_w) = 1/g(\nu_{|\mathcal F|}) \sum_i p({\mathcal F}_i | \Theta_w)\delta({\mathcal F}_i)$, where  $\delta({\mathcal F}_i)=1$ if ${\mathcal F}_i\in U_{\xi_{E}}^{\nu}$ and $0$ otherwise.
For each  class  ${\mathcal L}_{w}^{G_m}$   the class representative is the primitive maximizing $p(\gamma_{\nu}|\Theta_w)$. The representative primitive is  observed and labeled by inspection, according to the nomenclature given in biomechanics, see  \cite{hamill2006}. The same label is assigned to the  class  ${\mathcal L}_{w}^{G_m}$, without need to inspect all other primitives assigned to the class.

Average Hausdorff  distances between each primitive in a class and its class representative, for each class in group $G_2$, are given in  Table \ref{tab:hd}, where classes for $G_2$ are enumerated according to the labels illustrated in Fig. \ref{fig:primitives}. 
Note that in  Table \ref{tab:hd}  $R_{w}$  is the class representative, so $R_{w}\in{\mathcal L}_{w}^{G_m}$, $w=1,\ldots,7$;  $\forall\xi_{E\backslash R_{w}} $  abbreviates $\forall \xi_{E}\in{\mathcal L}_{w}^{G_2}, \xi_{E}\neq R_{w}$. Finally, ${\mathcal L}_{w}$ abbreviates 
${\mathcal L}_{w}^{G_2}$.  Note that distances with elements of other classes are obviously not considered, hence the dashes in  other classes columns.
\begin{table}[h!]
\centering
\caption{Average Hausdorff distance to each class representative in $G_2$ }\label{tab:hd}
\begin{tabular}{|l|lllllll|}
\hline
    & $R_1$ & $R_2$ & $R_3$ & $R_4$ & $R_5$ & $R_6$ & $R_7$\\
\hline
$\forall\xi_{E\backslash R_1}{\in} {\mathcal L}_1 $  &  0.121 &- &- &- &- &- &- \\
$\forall\xi_{E\backslash R_2}{\in} {\mathcal L}_2  $ & - & 0.173 &- & - & - &- &- \\
$\forall\xi_{E\backslash R_3}{\in} {\mathcal L}_3 $ & - & - & 0.144 &- & - & - & - \\
$\forall\xi_{E\backslash R_4}{\in} {\mathcal L}_4 $ &- & - & -  & 0.112 & -  & - &- \\
$\forall\xi_{E\backslash R_5}{\in} {\mathcal L}_5 $ & - & - & -  & - & 0.081 &- & - \\
$\forall\xi_{E\backslash R_6}{\in} {\mathcal L}_6 $ & - &- & - & - & - &0.142 & - \\
$\forall\xi_{E\backslash R_7}{\in} {\mathcal L}_7 $ & - & - & -  & - & - & -  & 0.114 \\
\hline
\end{tabular}
\end{table}

\subsection{Models  for recognition}\label{subsect:models}

The recognition problem is stated as follows. Given an unlabeled primitive $\gamma_u$, for group $G_m$ obtained by segmenting an activity (from any dataset) with the motion flux method, $\gamma_u$ is labeled by the label of class ${\mathcal L}^{G_m}_w$, if:
\begin{equation}\label{eq:a}
p(\gamma_u |\Theta_w)>p(\gamma_u|\Theta_i), \ \ \forall i, i\neq w
\end{equation}

We found experimentally that relying on the same parameters used for finding the classes of primitives, described in the previous sub-section, does not lead to optimal results. In fact, recomputing a DPM model for each class and introducing a loss function on the set of hypotheses, computed by thresholding the best classes,  leads to an improvement up to the $20\%$ in the recognition of an unknown primitive. 

To this end we compute a DPM for each class ${\mathcal L}^{G_m}_w$ using as observations the primitives collected in the class, by Algorithm \ref{algo:algo0}. Therefore the generated DPM model ${\mathcal M}_w$ for each class ${\mathcal L}^{G_m}_w$ is made by a number of components  with parameters $\Theta_w = \{\Theta_{w_1},\ldots, \Theta_{w_{\rho}}\}$, with $\rho$ varying according to the components generated for class ${\mathcal L}^{G_m}_w$. The number of components mirrors the idiosyncratic behavior of each class of primitives, therefore $\rho$ varies for each class ${\mathcal L}^{G_m}_w$.
To generate these DPM models we use all the three trajectories of the primitives $\gamma\in {\mathcal L}^{G_m}_w$, and for each of them we use the same decimation and feature vector as shown in Fig. \ref{fig:fv}. 

Given the refined classification, the recognition problem, at this point, is stated as follows. Let $\gamma_u=(\xi_{u_1},\xi_{u_2},\xi_{u_3})$ be an unknown primitive, of a specific group $G$, and let $\{{\mathcal F}_{u_1},\ldots,{\mathcal F}_{u_q}\}$ be the set of features the three trajectories are decomposed into.  Then $\gamma_u\in {\mathcal L}^{G_m}_w$, hence is labeled by the label of this class, if:
\begin{equation}\label{eq:choice0}
p({\mathcal F}_{u_1},{\ldots},{\mathcal F}_{u_q} |\Theta_w){=} \sum_{j=1}^{\rho}\pi_j\prod_{n=1}^q p({\mathcal F}_{u_n} |\Theta_{w_j}){>} p({\mathcal F}_{u_1},{\ldots},{\mathcal F}_{u_q} |\Theta_h) {=} \sum_{j=1}^{{\rho'}}\pi'_j\prod_{n=1}^q p({\mathcal F}_{u_n} |\Theta_{h_j})
\end{equation}
\noindent
for any parameter set $\Theta_h$ associated with a class ${\mathcal L}^{G_m}_h$ of the group $G_m$. Here $\pi_j$ and $\pi'_j$ are the mixture weights, with $\sum_j\pi_j=1$ and $\rho,\rho'$ indicate the number of components of the chosen models. For example, the model of class ${\mathcal L}^{G_2}_w$, with $w=1$, will have a set of parameters $\Theta_w=\{\Theta_{w_1},\ldots,\Theta_{w_{\rho}}\}$, while the model of 
class ${\mathcal L}^{G_2}_{w'}$, with $w'=3$, will have a set of parameters $\Theta_{w'}=\{\Theta_{w'_1},\ldots,\Theta_{w'_{\rho'}}\}$, with $w_{\rho}\neq w'_{\rho'}$.

This formulation is much more flexible than (\ref{eq:a}), also because it computes the class label by considering all the components and therefore it does not care whether the features are scattered amid components, and does not need to reconstruct the whole trajectories as was done for generating the classes of primitives. Furthermore, under this refined classification we can improve (\ref{eq:choice0}) considering a geometric measure to reinforce the statistics measure in the choice of the class label for $\gamma_u$. 

More precisely, let us form a set of hypotheses for an unknown primitive with feature set $\{{\mathcal F}_{u_1},\ldots,{\mathcal F}_{u_q}\}$ as follows (we are still assuming a specific group $G_m$):
\begin{equation}\label{eq:H}
{\mathbb H} = \{\langle C_{w_j},\Theta_{w_j} \rangle   | \prod_{n=1}^q p({\mathcal F}_{u_n} |\Theta_{w_j}){>}\eta, \langle C_{w_j},\Theta_{w_j}\rangle\in {\mathcal M}_w, w = 1,\ldots,k\}
\end{equation}
\noindent
Namely $C_{w_j}$ is a component of the DPM ${\mathcal M}_w$, with $w=1,\ldots,k$, $k$ the number of classes in group $G_m$, and $j =1,\ldots,\rho$,  such that the associated parameter $\Theta_{w_j}$ makes the joint probability of the features,  the primitive is decomposed into, greater than a threshold $\eta$. This means that we are collecting in  ${\mathbb H}$ those components coming from  all the models of group $G_m$, whose joint probability of the feature set of the unknown primitives $\gamma_u$ forms an hypotheses set, or a set from which we can select the correct label to assign to $\gamma_u$. 

\begin{figure}[t!]
\captionsetup[subfigure]{justification=centering}
  \centering
   \begin{subfigure}{0.5\textwidth}
   \includegraphics[width=0.99\textwidth]{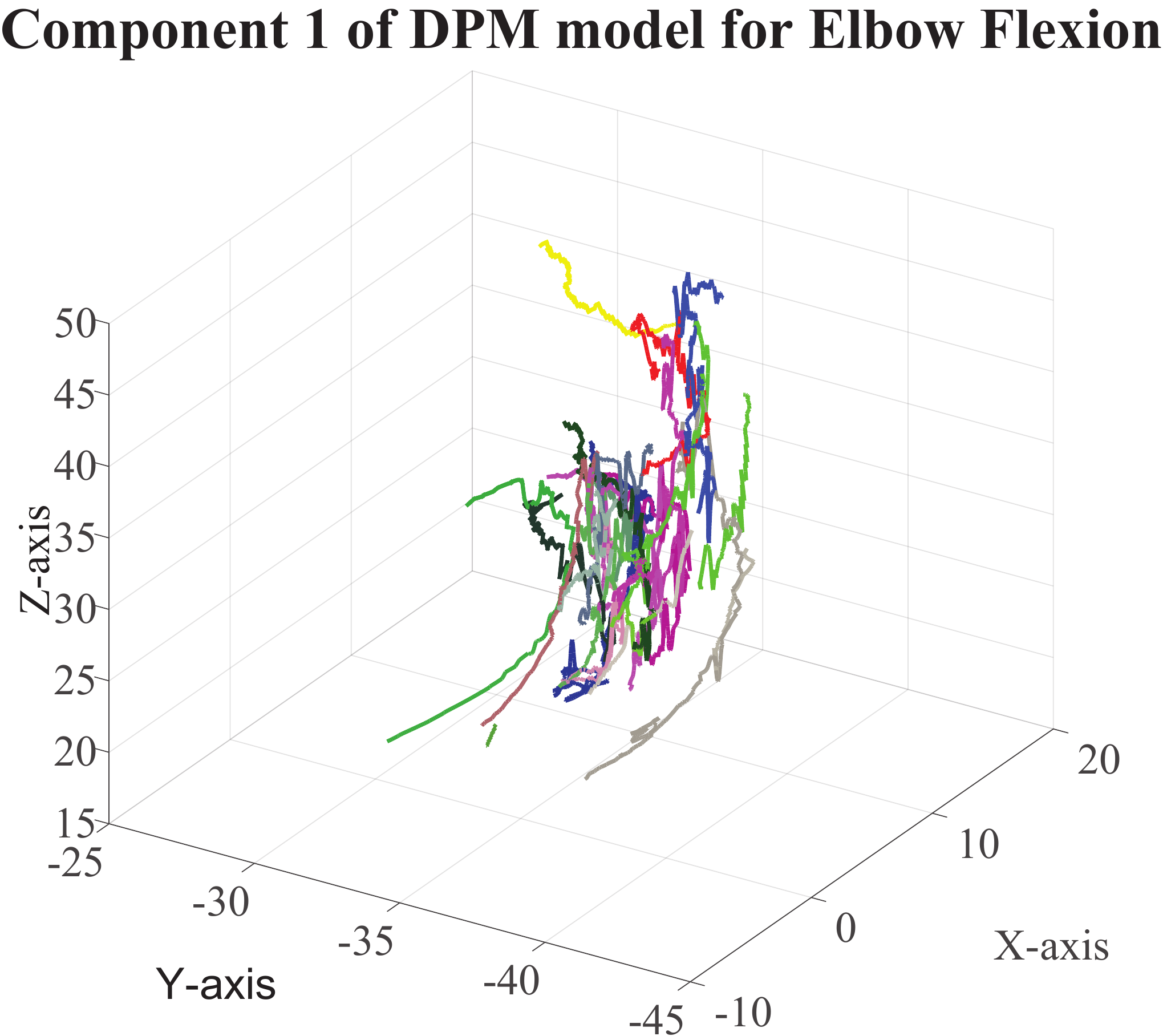}
   \caption{}
   \end{subfigure}%
   \begin{subfigure}{0.5\textwidth}
   \includegraphics[width=0.99\textwidth]{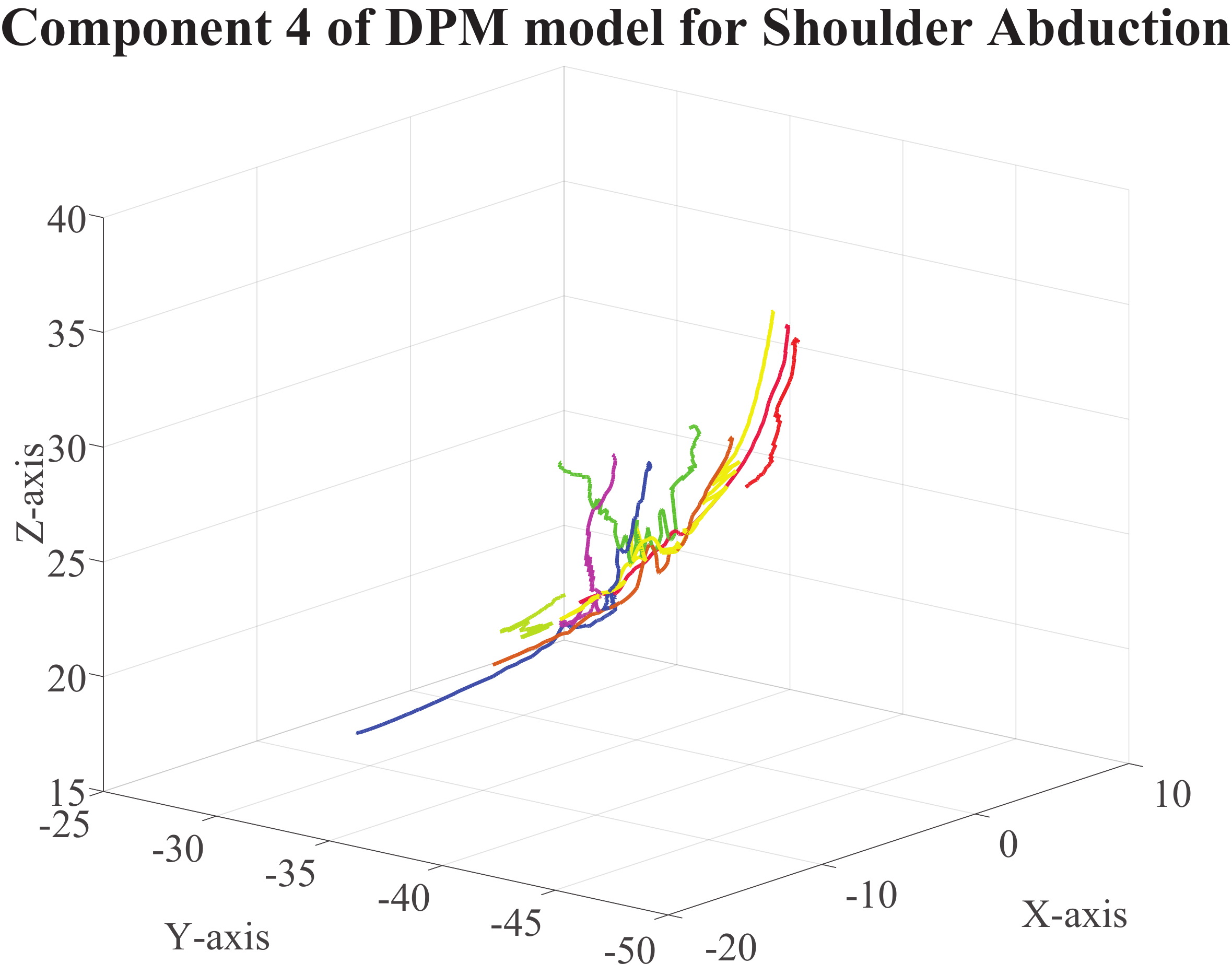}
   \caption{}
   \end{subfigure}
\caption{Manifold generated by a component of the DPM model for Elbow Flexion on the left and from a component of Shoulder Abduction on the right.}\label{fig:manifold}
\end{figure}

The advantage of the hypotheses set is that we delay the decision of choosing the labeled class for the unknown primitive to further evidence, which we collect by using geometric measures. The role of these geometric measures is essentially to evaluate the similarity between the  curve segments coming out from the features of $\gamma_u$ and those coming from the observations which are indexed in the components in ${\mathbb H}$. 
In the following we succinctly describe the  new geometric features, which are  computed as follows, both for the features of the unknown primitive $\gamma_u$ and for the features coming from the observations indexed in $C_{w_j}$. 
Let us consider any pair $\langle C_{w_j},\Theta_{w_j} \rangle\in {\mathbb H}$,
by definition (\ref{eq:H}), $C_{w_j}$ indexes features $\{{\mathcal F}_{\nu_1},\ldots,{\mathcal F}_{\nu_s}\}$, $s$ varying according to the specific component $C_{w_j}$. For each of these features we consider the points of the trajectory $\xi^{\nu}$, recovered from  the decimated trajectory $\hat{\xi}_{\nu}$,  between $(x_{i-1},y_{i-1},z_{i-1})$ and $(x_{i+1},y_{i+1},z_{i+1})$.  Let us consider these curve segments, which we combine whenever they occur in sequence  in $C_{w_j}$ and call any of these  curve segments ${\bf y}$. In particular,  the collection of these segments  in  $C_{w_j}$ is called the manifold of $C_{w_j}$, denoted $man(C_{w_j})$, and the collection of segments generated from the features of $\gamma_u$ is denoted $man(\gamma_u)$, examples are given in Fig. \ref{fig:manifold}.  
\noindent

We compute for each ${\bf y}$ both in  $man(C_{w_j})$ and in $man(\gamma_u)$  the tangent ${\bf t}$,  normal ${\bf n}$ and binormal ${\bf b}$ vectors. Based on these vectors, we compute the ruled surface ${\mathcal R} = \frac{{\bf n} \times {\bf n}'}{\|{\bf n} \times {\bf n'}\|}$, where ${\bf n}'$ is the derivative of ${\bf n}$. 
The ruled surface forms a ribbon of tangent planes to the curve segment ${\bf y}$. 
In particular, let us distinguish the curve segments in $man(\gamma_u)$ denoting them ${\bf y}_{u}$.
We compute the distances between any curve segment ${\bf y}\in man(C_{w_j})$  and ${\bf y}_{u}\in man(\gamma_u)$ as the distance between the projection ${\bf y}_{{\pi}}$ of ${\bf y}$ on the ruled surface tangent to ${\bf y}$,  and the {\em closest point} ${\bf q}$ of ${\bf y}_{u}$ to ${\bf y}_{{\pi}}$. 
We denote this distance $\delta({\bf y}_{u},{\bf y})$. 
We consider also the distance between the Frenet frames at closest points ${\bf q}$ of ${\bf y}_{u}$ and point ${\bf q}'$ of ${\bf y_{\pi}}$ denoted $F_R$ and computed as follows: $F_R({\bf q},{\bf q}') = \mbox{trace}(({\mathcal I}-R_{{\bf q},{\bf q}'})({\mathcal I}-R_{{\bf q},{\bf q}'})^{\top})$, with ${\mathcal I}$ the identity matrix and $R_{{\bf q},{\bf q}'}$ the rotation, in the direction from ${\bf q}$ to ${\bf q}'$. Then the cost of a component $C_{w_j}$ in ${\mathbb H}$, given an unknown primitive $\gamma_u$, with feature set $\{{\mathcal F}_{u_1},\ldots,{\mathcal F}_{u_q}\}$, is defined as:
\begin{equation}
Cost(C_{w_j} \in {\mathbb H} | \gamma_u ) = max\{\delta({\bf y}_{u},{\bf y}) + F_R({\bf q},{\bf q}') |  {\bf y_{u}}\in man(\gamma_u)\mbox{ and } {\bf y}\in man(C_{w_j}) \} 
\end{equation}
Note that both $\delta({\bf y}_{u},{\bf y})$ and  $F_R({\bf q},{\bf q}')$ were both computed looking at the minimum distance between a considered curve segment and the projection on the ruled surface of the other curve segment.
Hence the component minimizing the above cost and maximizing the probability in (\ref{eq:choice0}) will indicate the class label, since its related parameter indicates exactly a component of one of the classes ${\mathcal L}_w^{G_m}$. Note that if in (\ref{eq:choice0}) $\eta$ is taken to be equal to $max(\prod_{n=1}^q p({\mathcal F}_{u_n} |\Theta_{w_j}))$ then ${\mathbb H}$ would have only a single element $\langle C_{w_j},\Theta_{w_j} \rangle$. Hence to find the correct label for $\gamma_u$ we push $\eta$ as high as possible using the above cost. More precisely,  the  component of the class ${\mathcal L}_w^{G_m}$ which should label the unknown primitive $\gamma_u$ is computed as follows:
\begin{equation} \label{eq:best}
C^{\star} = \displaystyle{ \argmin_{C_{w_j}} \sup_{\eta} \{ Cost(C_{w_j} \in {\mathbb H} | \gamma_u ) |  \prod_{n=1}^q p({\mathcal F}_{u_n} |\Theta_{w_j}){>}\eta \}}
\end{equation}

To conclude this section we can note that the computation of the hierarchical model that first  generates the primitive classes and then uses these generated sets to estimate model parameters to be used in the recognition of an unknown primitive, has an exponential cost,  in the dimension of the features and in the size of the observations.  However using the computed models to recognize an unknown primitive is ${\mathcal O}(n^2 log\ n)$ where $n$ is the size of $\gamma_u$, since all the curve segments in the models can be precomputed together with the models. Results on both the primitive generation and on recognition are given in the next section. 

\section{Experiments}\label{sec:prim_results}

\begin{figure}[t!]
\captionsetup[subfigure]{justification=centering}
  \centering
   \begin{subfigure}{0.5\textwidth}
   \includegraphics[width=0.99\textwidth]{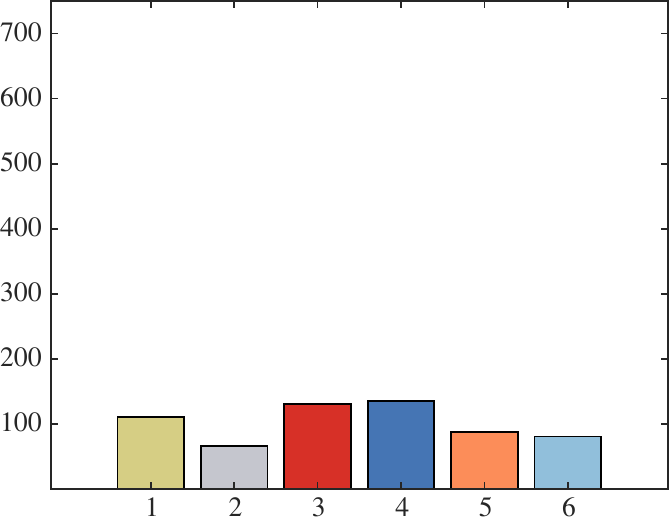}
   \caption{}
   \end{subfigure}%
   \begin{subfigure}{0.5\textwidth}
   \includegraphics[width=0.99\textwidth]{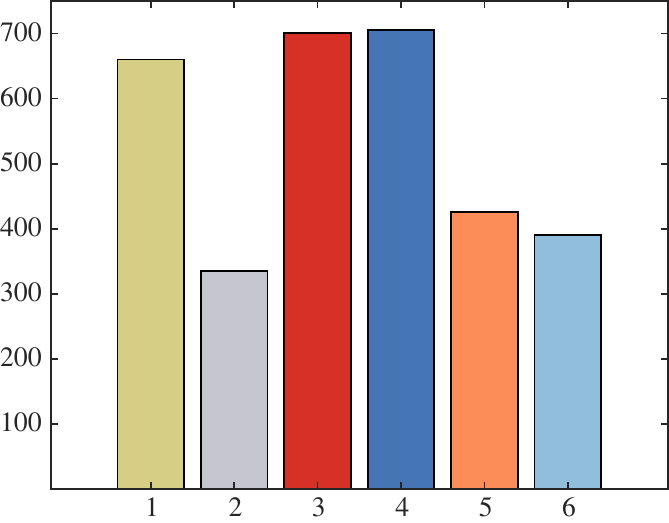}
   \caption{}
   \end{subfigure}
   \begin{subfigure}{0.5\textwidth}
   \includegraphics[width=0.99\textwidth]{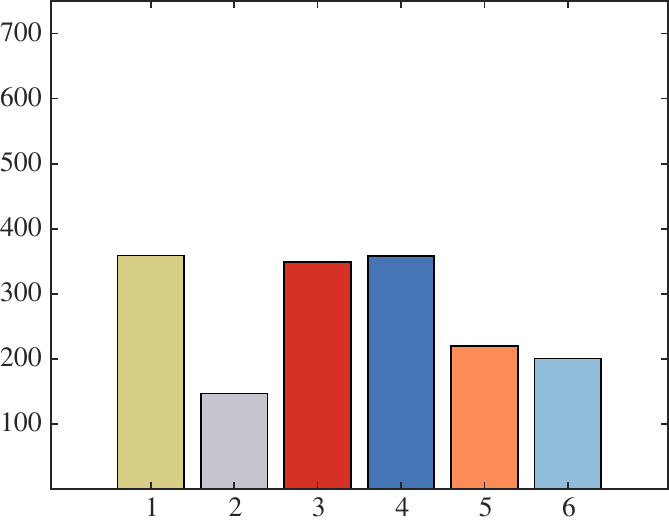}
   \caption{}
   \end{subfigure}%
   \begin{subfigure}{0.5\textwidth}
   \includegraphics[width=0.99\textwidth]{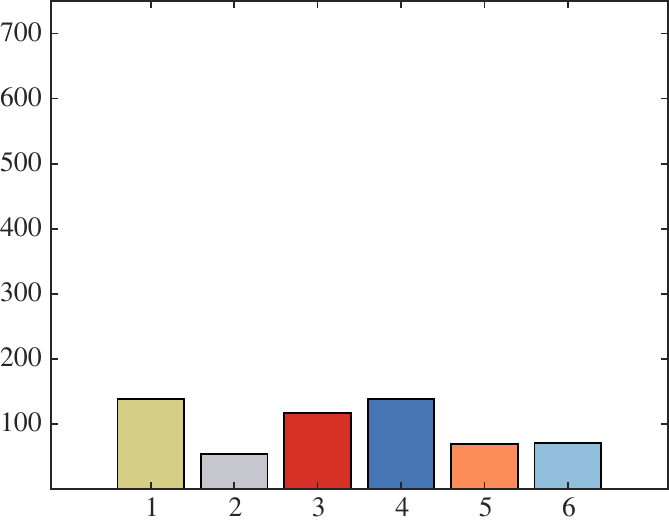}
   \caption{}
   \end{subfigure}
   \begin{subfigure}{0.5\textwidth}
   \includegraphics[width=0.99\textwidth]{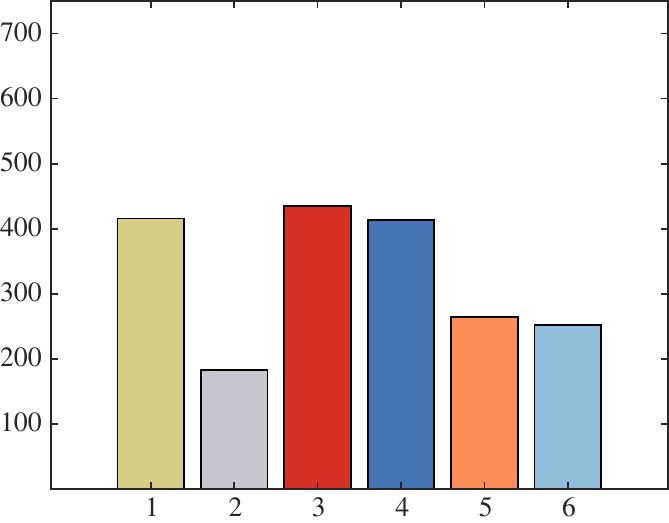}
   \caption{}
   \end{subfigure}
\caption{Total number of discovered primitives for each group for the five most general categories of the ActivityNet dataset. Clock-wise from top-left: \emph{Eating and drinking Activities}; \emph{Sports, Exercise, and Recreation}; \emph{Socializing, Relaxing, and Leisure}; \emph{Personal Care}; \emph{Household Activities}. Each color corresponds to a different group following the convention of Fig.~\ref{fig:primitives}. Note: Axes scale is shared among the plots.}
\label{fig:primdata}
\end{figure}

 In this section we evaluate the  proposed framework for discover and classification of human motion primitives. For all the evaluations we consider three reference MoCap public datasets \cite{mandery2015,ionescu2014,CMU}.

First we evaluate the accuracy of the motion primitives discovered using the motion flux, further we evaluate the accuracy of the classification and recognition.
 Additionally, we examine the distribution of recognized primitives with respect to the type of performed activity on the ActivityNet dataset \cite{ghanem2017}. 
 Finally, we address the dataset of human motion primitives we have created, which consists of the primitives discovered on the three reference MoCap datasets using the motion flux, and the DPM models established for each primitive category.
 
\subsection{ Reference Datasets}\label{sec:datasets}
 The datasets we consider for the evaluation of the motion flux are the Human3.6M dataset (H3.6M) \cite{ionescu2014}, the CMU Graphics Lab MoCap database (CMU) \cite{CMU} and the KIT Whole-Body Human Motion Database (KIT-WB) \cite{mandery2015}. The sampling rates used in these datasets are 50Hz for H3.6M, 60/120Hz for CMU and 100Hz for KIT-WB. In order to have the same sampling rate for all sequences we have transformed all of them to 50Hz.
 The pose of the joints specified in Fig.~\ref{fig:groups} are extracted for each frame of the sequences as described in the preliminaries, considering the ground-truth 3D poses. For KIT-WB the trajectories of the joints are computed from the marker positions taken from the C3D files.
 We considered $40$ activities from the three reference datasets.  
 Fig.~\ref{fig:primdata} shows the total number of motion primitives discovered for the five most general activities according to the ActivityNet taxonomy based on the motion flux for each group $G_m$.
 Table~\ref{tab:Tab1} shows the total number of motion primitives discovered from the three datasets.

\begin{table}[ht!]
\caption{Total number of unlabeled primitives discovered for each group using the motion flux on the reference datasets }\label{tab:Tab1}
\centering
\begin{tabular}{r|cccccc}
& \textbf{G1} & \textbf{G2} & \textbf{G3} & \textbf{G4}& \textbf{G5} & \textbf{G6}\\\hline
\textbf{Total}& 1665 & 759 & 1773 & 1703 & 1152 & 1015\\ 
\end{tabular}
\end{table}

\begin{figure}[t!]
  \centering
   \includegraphics[width=1\textwidth]{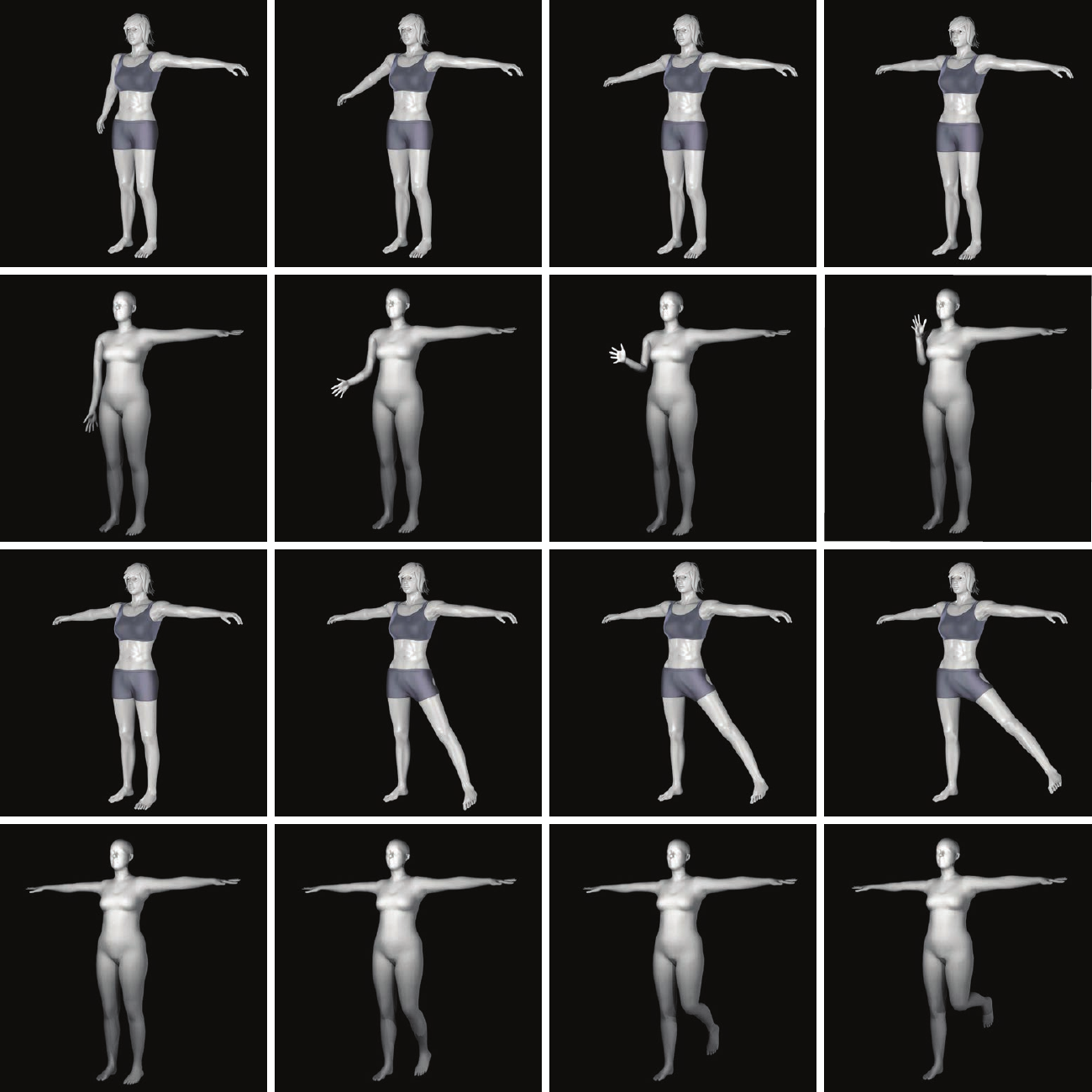}

\caption{Example of synthetic motion primitive, specifically right arm Shoulder Abduction (first row) and Elbow Flexion (second row), left leg Hip abduction (third row) and Knee Flexion (fourth row). For each synthetic motion primitive the four imaged poses match four representative poses extracted from the animation of the aforementioned primitive.}
\label{fig:character}
\end{figure}

\subsection{Motion Primitive Discovery}
To evaluate the accuracy of primitive discovery based on the motion flux, we created a baseline relying on a synthetic dataset of motion primitives. This was necessary to mitigate the difficulty  in measuring accuracy, due to the lack of a ground truth.

The synthetic dataset of motion primitives we created is formed by animations of 3D human models for each of the 69 primitive classes discovered in Sec. \ref{sec:features}. The human models were downloaded from the dataset provided by \cite{loper2015} or acquired from \cite{turbo,renderpeople}. To obtain further characters the shapes of the human models were randomly modified taking care of human height and limb length limits. 

Animations of the characters were produced moving the skeleton joints belonging to the 3D human models from a start pose to an end pose representing the primitives. Specifically, for each primitive of each skeleton group the animation was generated in Maya or Blender (depending on the 3D human model format) moving the group joints according to angles, gait speed and limbs proportions as described in \cite{de2014,gates2016,hamill2006,abernethy2013}.

The  dataset reference skeleton, see Fig. \ref{fig:groups}  is matched with the 3D human mesh models  by fitting the joint poses of the synthetic data to the reference skeleton, basing on MoSh \cite{loper2014,varol2017}.
Examples of synthetic motion primitives, namely the primitives Shoulder abduction and Elbow flexion for the right arm, and Hip abduction and Knee flexion for the left leg, are illustrated in Fig. \ref{fig:character}, where for each primitive four representative poses extracted from the animations are shown.

The baseline for evaluating accuracy  was created generating 4500 random length sequences of synthetic motion primitives placing them one after another in a random order. Between two consecutive primitives a transition phase from the end pose of the preceding one to the beginning pose of the subsequent one was added.  

With this procedure we know precisely the endpoints of each primitive.

Then we applied the `motion flux' method described in Sec. \ref{sec:flux} to the 3D joints trajectories extracted from the automatically generated sequences and collected the end points of the discovered primitives. 

We use the Mean Absolute Error (MAE) and Root Mean Square Error (RMSE) metrics to assess the accuracy of the collected endpoints with respect to the known end points in the generated sequences. Let $S$ be the total number of generated sequences. Let $\{\hat{e}_{i,s}\}_{i=1}^{N_G^{(s)}}$ be the $i$-th automatically discovered endpoint based on the motion flux for the generated sequence $s=\{1,\ldots,S\}$, with $N_G^{(s)}$ the number of primitives for Group $G$ and sequence $s$. Denoting $\{\bar{e}_{i,s}\}_{i=1}^{N_G^{(s)}}$ the $i$-th endpoint in the generated sequence $s$, the MAE and RMSE metrics are  defined as follows:
\begin{equation*}
\begin{array}{cc}
\displaystyle{MAE {=} \frac{1}{S}\sum_{s=1}^{S}\frac{\sum_{i=1}^{N_G^{(s)}} |\bar{e}_{i,s} - \hat{e}_{i,s}|}{N_G^{(s)}}}, & \displaystyle{RMSE {=} \sqrt{\frac{1}{S}\sum_{s=1}^{S}\frac{\sum_{i=1}^{N_G^{(s)}} (\bar{e}_{i,s} - \hat{e}_{i,s})^2}{N_G^{(s)}}}}.
\end{array}
\end{equation*}

Results shown in Table~\ref{tab:primexper} prove that the proposed method discovers motion primitives quite accurately, since the  endpoints are close to those of the automatically generated sequences.

\begin{figure}[t!]
\captionsetup[subfigure]{justification=centering}
  \centering
    \begin{subfigure}{0.5\textwidth}
   \includegraphics[width=.99\textwidth]{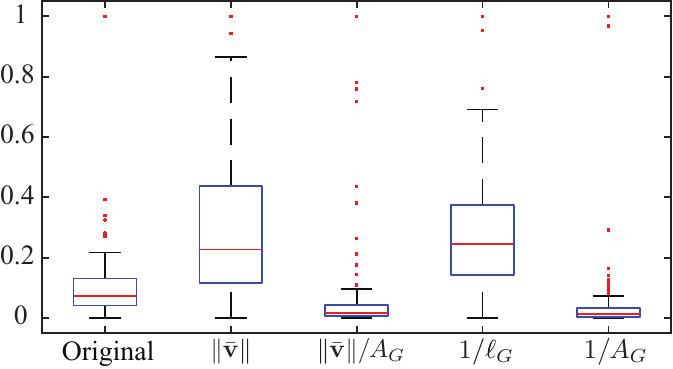}
   \caption{}
   \end{subfigure}%
       \begin{subfigure}{0.5\textwidth}
   \includegraphics[width=.99\textwidth]{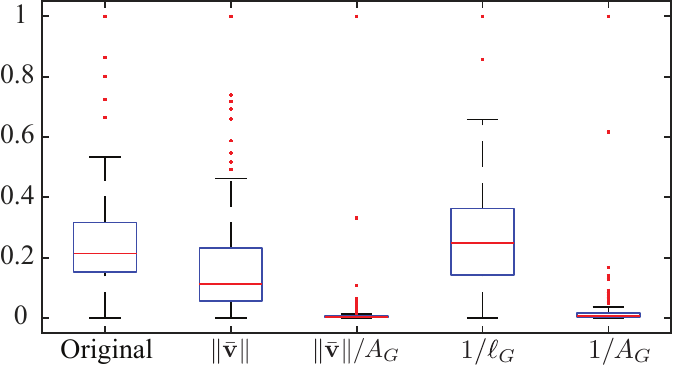}
   \caption{}
   \end{subfigure}
\caption{Arc length distribution of original and scaled primitives of a specific category for group $G_1$ (\textbf{left}) and $G_4$ (\textbf{right}). The first box in each box plot, corresponds to the original arc length distribution, the next four are the arc length distributions obtained scaling the primitives original data using the detailed scaling factors. 
  Each box indicates the inner 50th percentile of the trajectory data,  top and bottom of the box are the 25th and 75th percentiles, the whiskers extend to the most extreme data points  not considered outliers,  crosses are the outliers.
}
\label{fig:boxplot}
\end{figure}

\begin{table}[ht!]
\caption{Accuracy of discovered primitive endpoints (in number of frames)}\label{tab:primexper}
\centering
\begin{tabular}{r|ccccccc}
 & \textbf{G1} & \textbf{G2} & \textbf{G3} & \textbf{G4} & \textbf{G5} & \textbf{G6} & \textbf{Overall}\\\hline
\textbf{MAE}& 2.8 & 3.2 & 2.9 & 3.4 & 3.6 & 4.1 & 3.3\\
\textbf{RMSE} & 3.7 & 4.2 & 4.1 & 4.6 & 4.8 & 5.2 & 4.4\\
\end{tabular}
\end{table}

Furthermore, to evaluate the effects of the normalization in Fig.~\ref{fig:boxplot} we show the arc length distribution of motion primitives with and without normalization, as well as considering different normalization constants. 
 
For comparison we consider alternative normalization constants based on anatomical properties and execution style. Specifically, we consider normalization based on the average velocity along $\gamma\in \Gamma_G$, denoted as $\|\bar{\bf v}\|$, and based on the area $A_G$ covered by group $G$ during its motion. The first is related to the execution speed of the motion and the sampling rate of the data, while the latter is considering anatomical differences among the subjects.

In Fig.~\ref{fig:boxplot} the first box in each plot corresponds to the original distribution and the following boxes correspond to the distributions resulting by scaling the original one  with $\|\bar{\bf v}\|$, $\|\bar{\bf v}\|/A_G$, $1/\ell_G$, and $1/A_G$, respectively. We note that normalizing the primitives based on the inverse of the limb length, i.e. $\ell_G$, consistently results to an arc length distribution closer to the normal, minimizing the number of outliers indicated by red crosses in the figure. 
 This result is consistent across different activities and groups justifying the choice of $k_G=1/\ell_G$ for anatomical normalization.

\begin{figure*}\label{fig:primitivesEll}
  \centering
   \includegraphics[width=.99\textwidth]{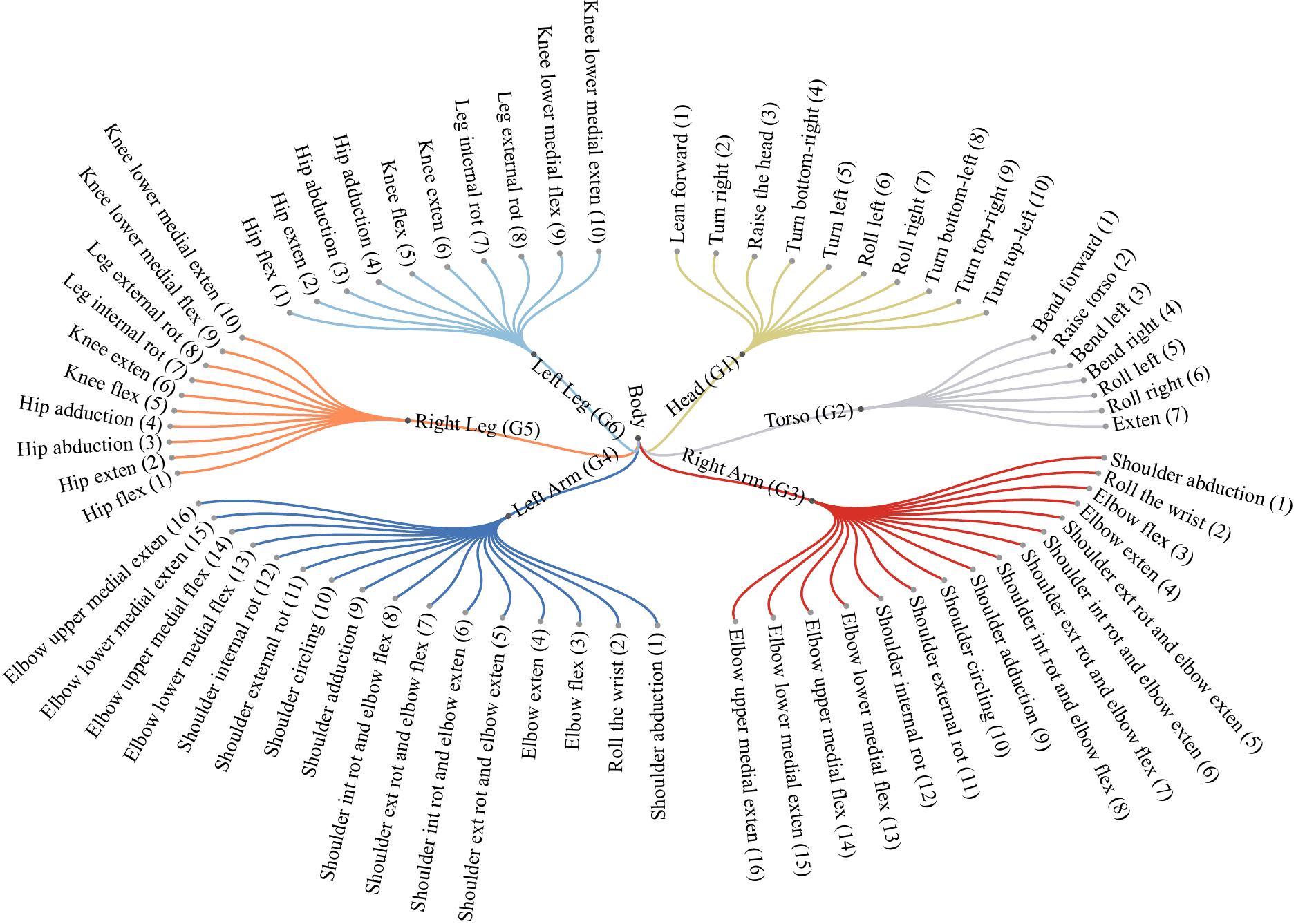}\hfill
\caption{Diagram showing the motion primitives of each group.  Abbreviation \emph{ext} stands for external, \emph{int} for internal, \emph{rot} for rotation, \emph{exten} for extension, and \emph{flex} for flexion.}\label{fig:primitives}
\end{figure*}

\subsection{ Motion Primitive Classification and Recognition}
 As discussed in Section~\ref{sec:features}, the set of primitive categories for each group is generated by  a DPM model given the collection of discovered primitives as observations. In this way a total of 69 types of primitives were identified, each described by the distribution parameters. By  inspecting  a representative primitives for each category, we  observed that they correspond to a subset of motion primitives defined in biomechanics. Therefore we generated new DPM models to obtain parameters and corresponding labels for each category. The labeled collection of motion primitives is depicted in Fig.~\ref{fig:primitives}. 

 To evaluate the coherence of the generated classes we performed 10 cycles of random sampling, with a rate of 10\% at each cycle,  of the primitives in each class and  verified  the  class consistency. Only $\sim 2\%$ of the primitives were not correctly classified, according to the label  assigned to the class. 
 
\begin{figure}[thp!]
  \centering
  \captionsetup[subfigure]{justification=centering}
   \begin{subfigure}{0.5\textwidth}
   \includegraphics[width=0.95\textwidth]{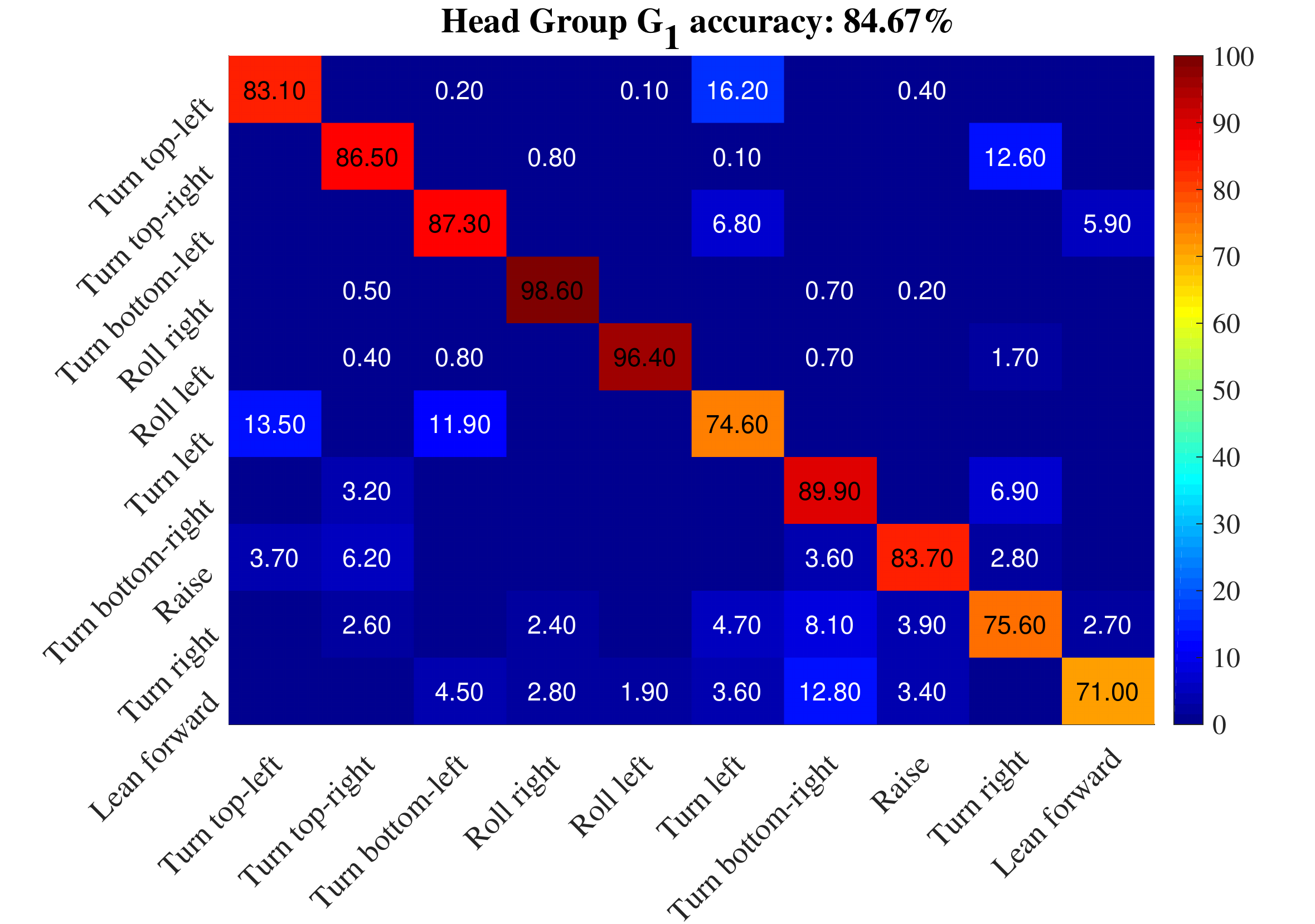}
   \caption{}
   \end{subfigure}%
   \begin{subfigure}{0.5\textwidth}
   \includegraphics[width=0.99\textwidth]{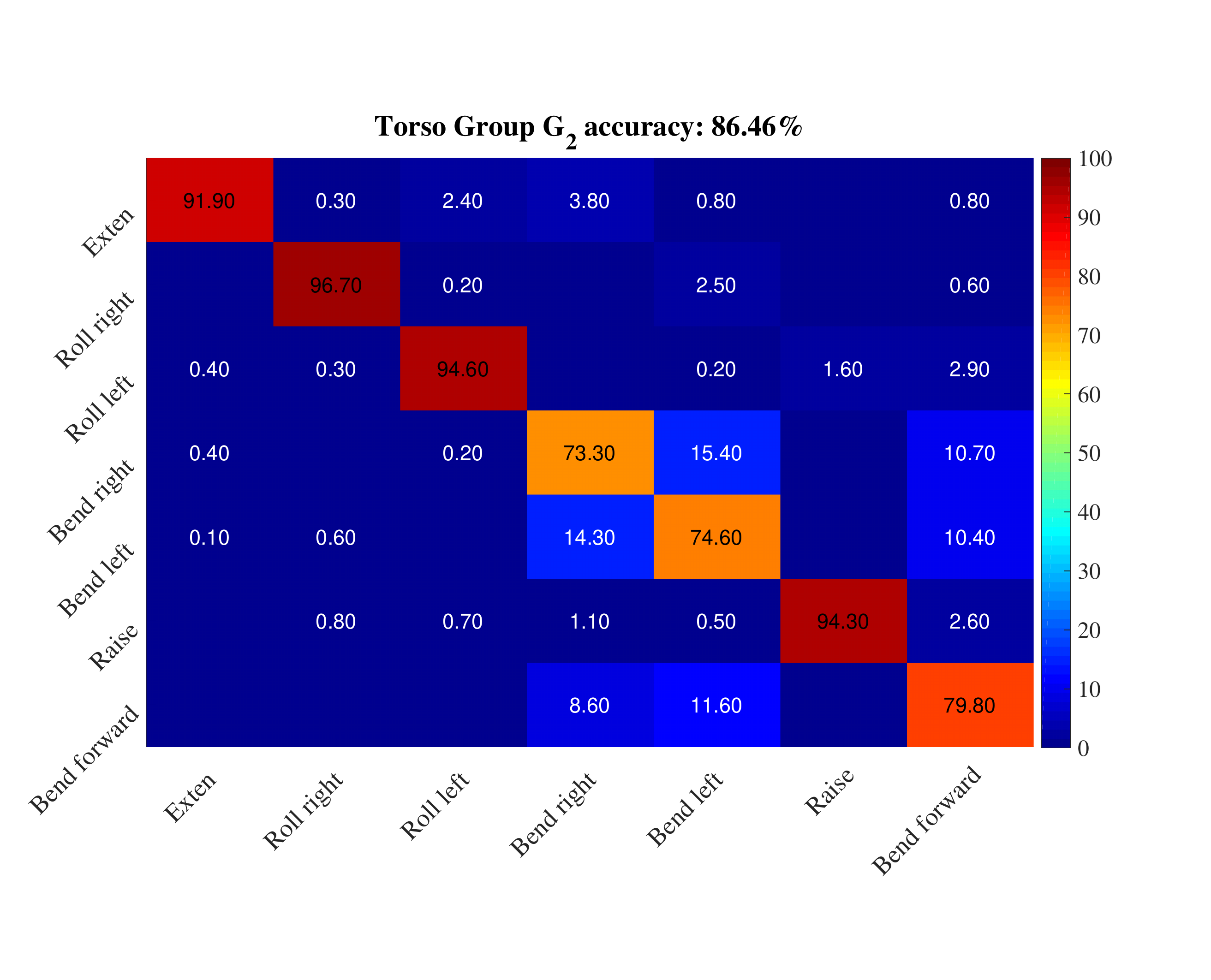}
   \caption{}
   \end{subfigure}
   \begin{subfigure}{0.5\textwidth}
   \includegraphics[width=0.99\textwidth]{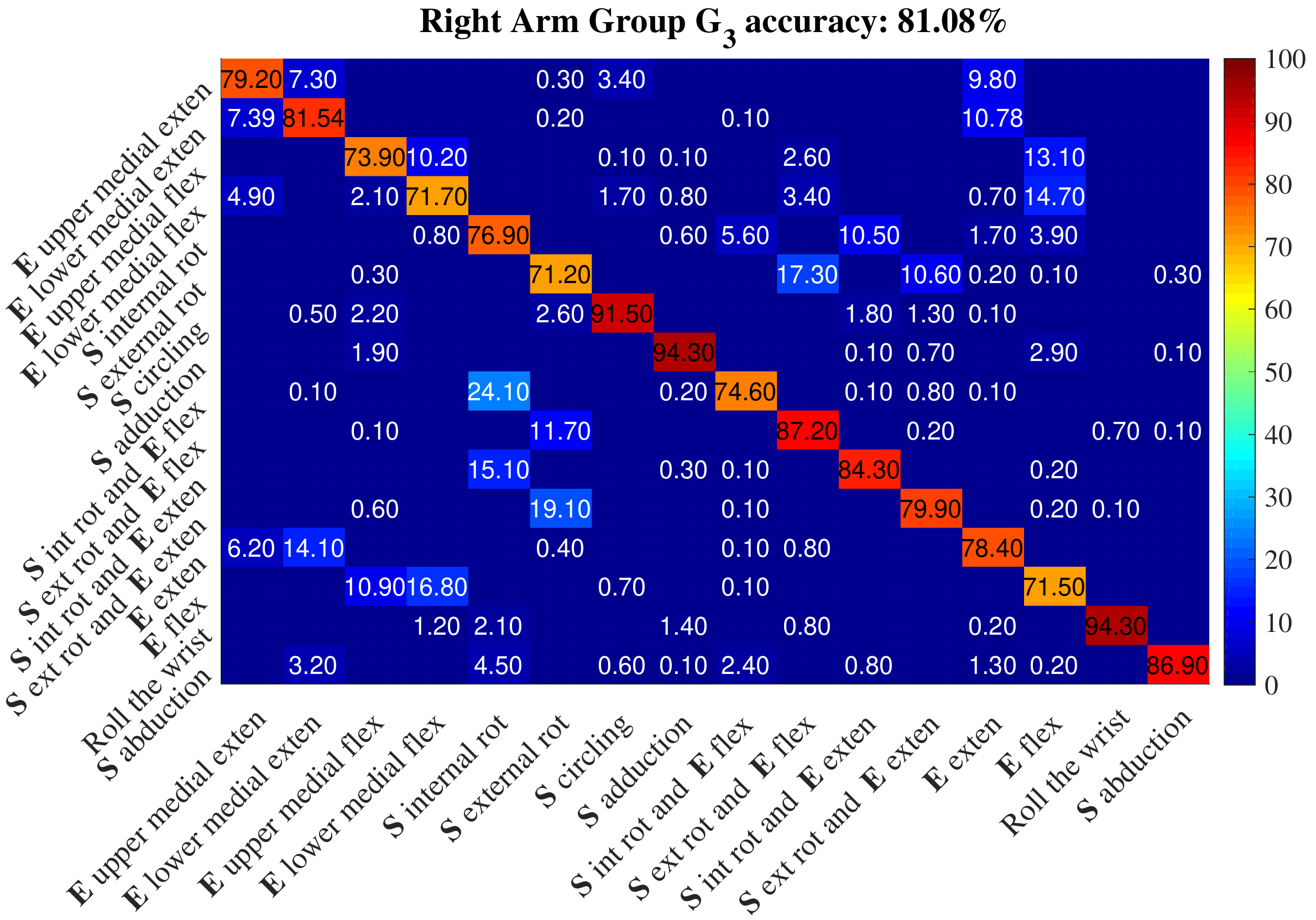}
   \caption{{\bf S} indicates the {\em shoulder}, {\bf E}  the {\em elbow}.}
   \end{subfigure}%
   \begin{subfigure}{0.5\textwidth}
   \includegraphics[width=0.99\textwidth]{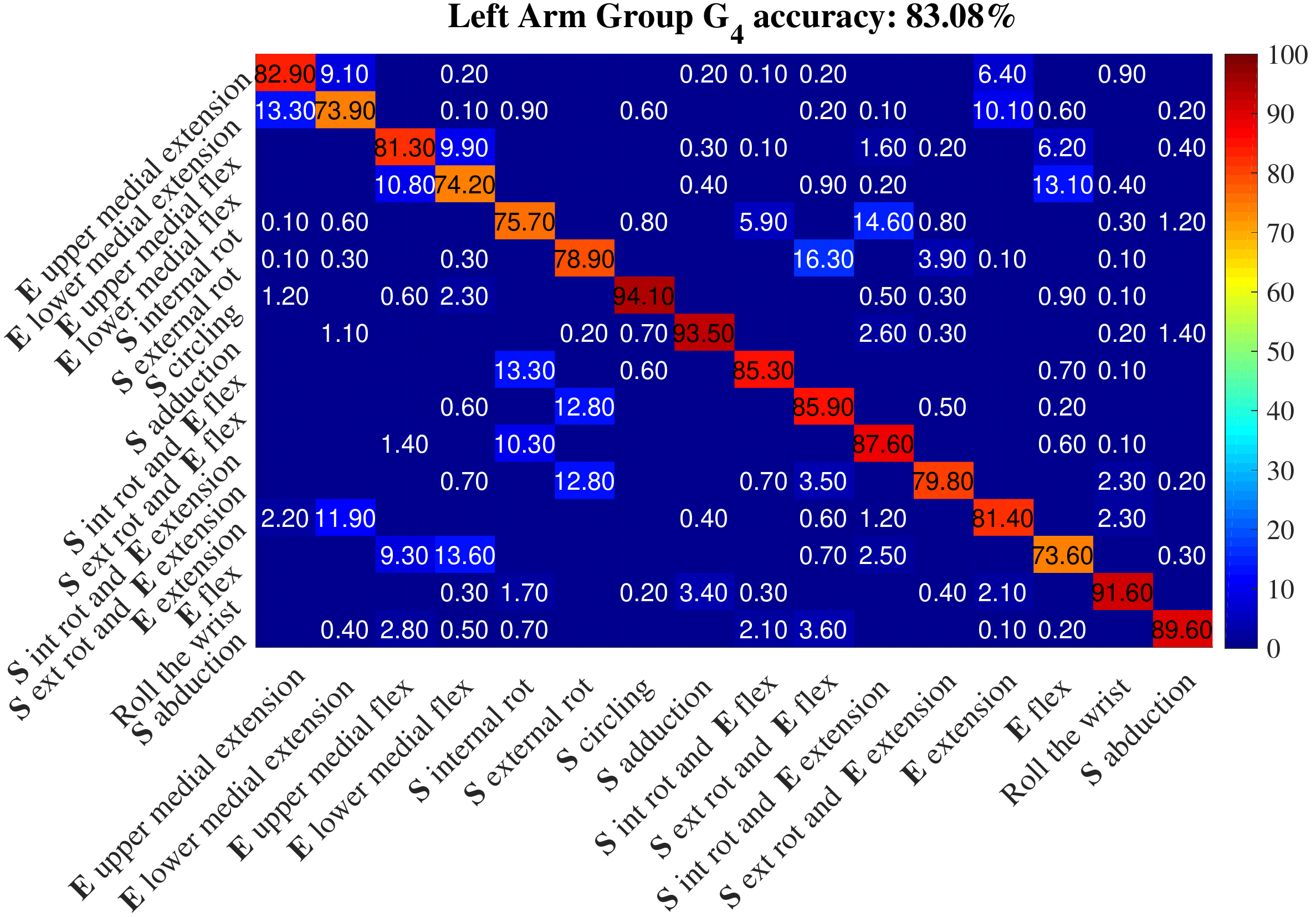}
   \caption{{\bf S} indicates the {\em shoulder}, {\bf E}  the {\em elbow}.}
   \end{subfigure}
   \begin{subfigure}{0.5\textwidth}
   \includegraphics[width=0.99\textwidth]{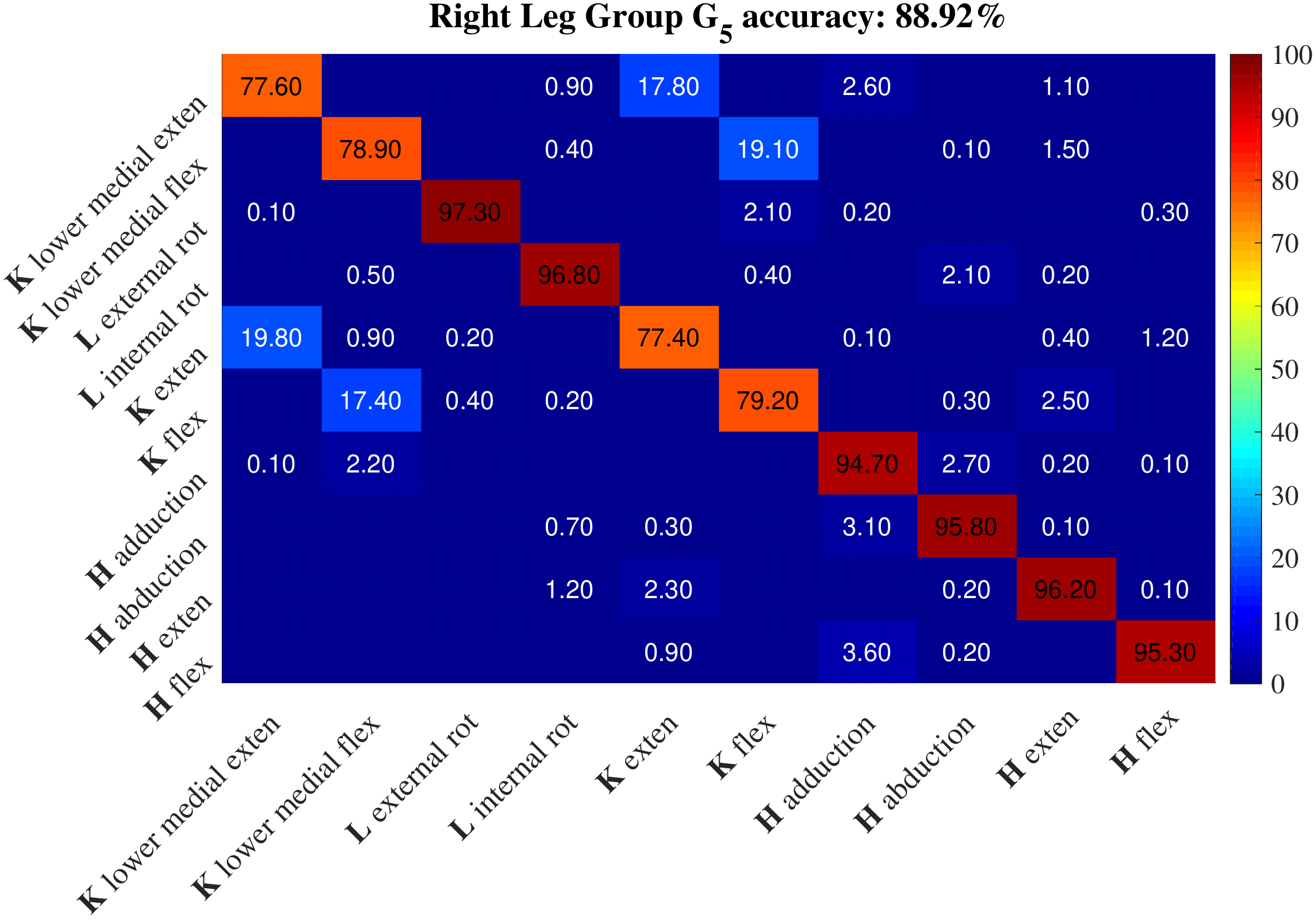}
   \caption{{\bf K} indicates the {\em knee}, {\bf L} the {\em leg} and {\bf H} the {\em hip}}
   \end{subfigure}%
   \begin{subfigure}{0.5\textwidth}
   \includegraphics[width=0.99\textwidth]{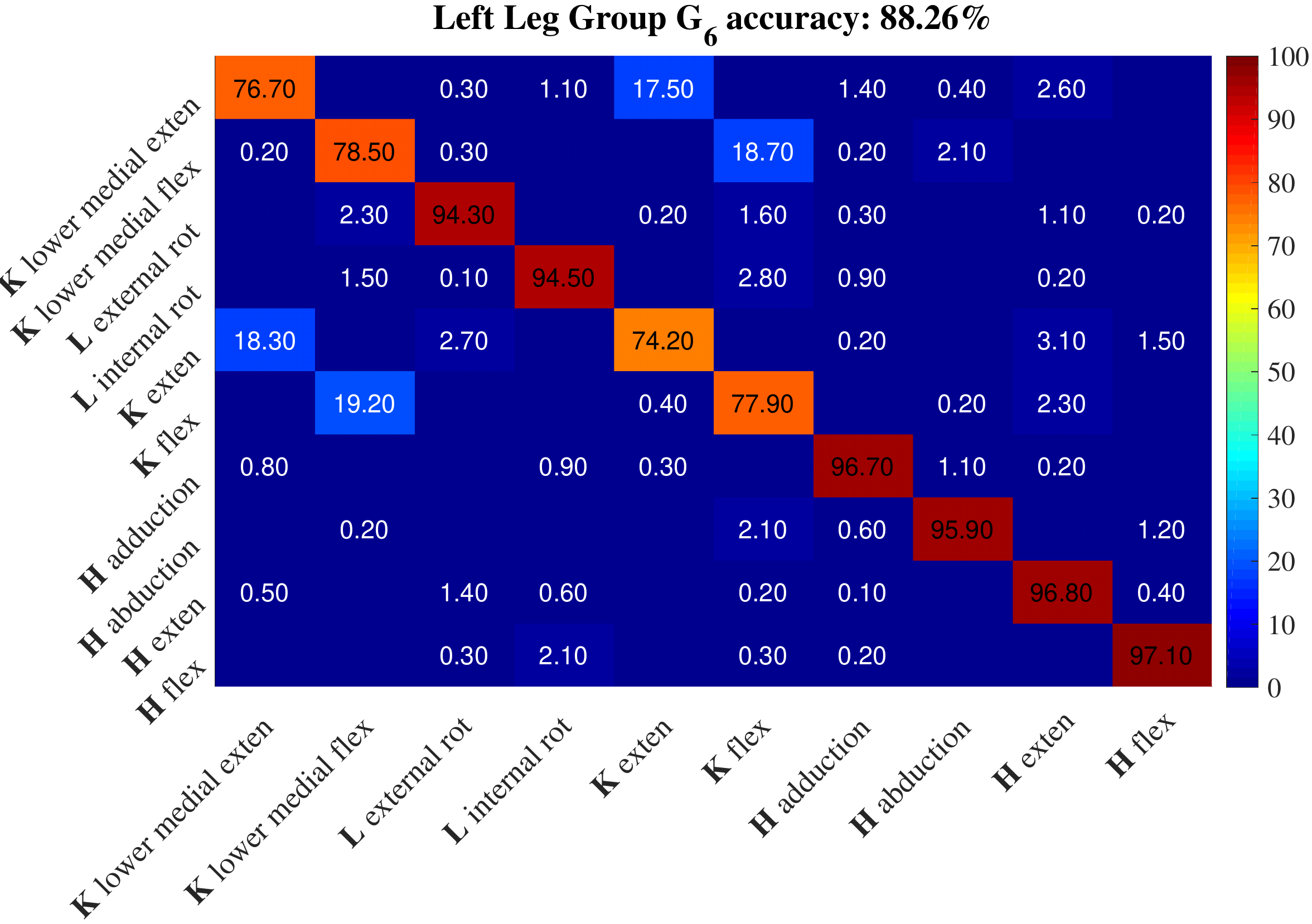}
   \caption{{\bf K} indicates the {\em knee}, {\bf L} the {\em leg} and {\bf H} the {\em hip}}
   \end{subfigure}

\caption{Confusion matrices for motion primitive recognition.  The matrices for G1 and G2 are shown at the top, G3 and G4 at the middle, while G5 and G6 are shown at the bottom. 
}
\label{fig:Confflux}
\end{figure}

For the recognition we adopted the protocol P2 used for pose estimation (see \cite{sanzari2016,tekin2016}) using one specific subject for testing.   Table~\ref{tab:ablation} presents the average accuracy of the recognition for each group, as well as an ablation study with respect to the components of the cost function used in eq. (\ref{eq:best}). Fig.~\ref{fig:Confflux} shows the corresponding confusion matrices. The results suggest that the DPM classification together with the proposed recognition method capture the main characteristics of each motion primitive category. 
 
 Finally, we evaluate the recognition accuracy by considering the same sequences though computing the subject's pose directly from the video frames using \cite{sanzari2016}. The corresponding results are shown in parentheses in the last column of Table~\ref{tab:ablation}. We note that the recognition accuracy decreases in average just by 4\% by using the estimated pose.
 
\typeout{********************* COnfusion matrices***************}
 
 \newcommand{\confdim}{5.6cm}

\begin{table}[ht!]
\caption{Primitive recognition accuracy and ablation study}\label{tab:ablation}
\centering
\resizebox{0.98\columnwidth}{!}{
\begin{tabular}{cccccc}
\textbf{Group} & \textbf{Projection on} & \textbf{Frenet frame} & \textbf{Torsion} & \textbf{Curvature} &\textbf{All} \\
{} & \textbf{tangent plane} & \textbf{rotation} & {} & {} & {} \\

\textbf{G1} & 0.82 & 0.80 & 0.70 & 0.72 & 0.84 (0.82) \\
\textbf{G2} & 0.85 & 0.82 & 0.75 & 0.75 & 0.86 (0.84) \\
\textbf{G3} & 0.80 & 0.80 & 0.73 & 0.74 & 0.82 (0.78) \\
\textbf{G4} & 0.80 & 0.79 & 0.75 & 0.77 & 0.83 (0.76) \\
\textbf{G5} & 0.87 & 0.86 & 0.72 & 0.72 & 0.88 (0.81) \\
\textbf{G6} & 0.86 & 0.86 & 0.71 & 0.73 & 0.88 (0.82) \\ \hline\hline
\textbf{Average} & 0.83 & 0.82 & 0.73 & 0.76 & 0.85 (0.81) \\
\end{tabular}}
\end{table}

\noindent

\subsection{Primitives in Activities}
\begin{figure}[thp!]
\captionsetup[subfigure]{justification=centering}
  \centering
   \begin{subfigure}{.5\textwidth}
   \centering
   \includegraphics[width=.99\textwidth]{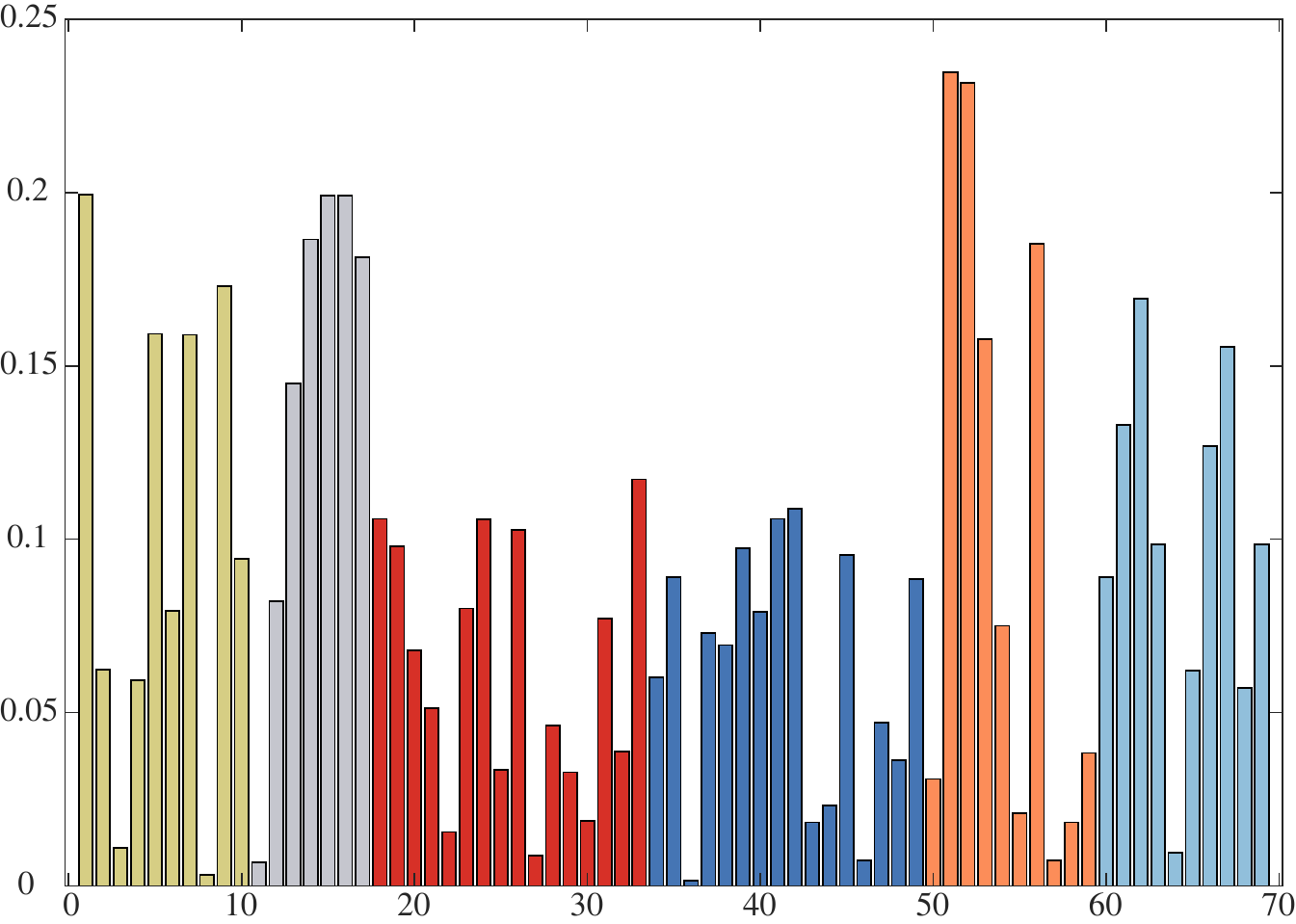}
   \caption{}
   \end{subfigure}%
   \begin{subfigure}{.5\textwidth}
   \centering
   \includegraphics[width=.99\textwidth]{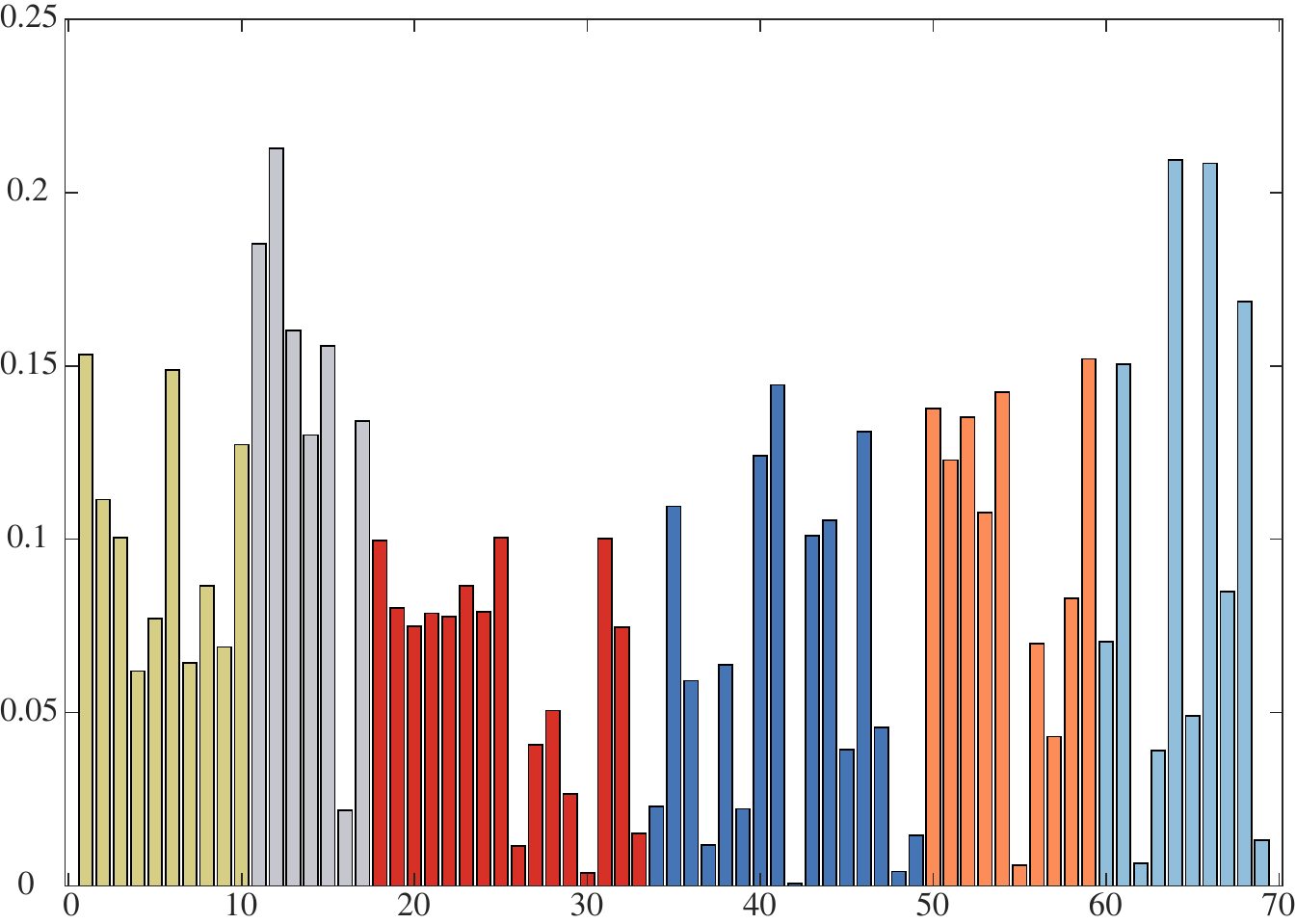}
   \caption{}
   \end{subfigure}
   
   \begin{subfigure}{.5\textwidth}
   \centering
   \includegraphics[width=.99\textwidth]{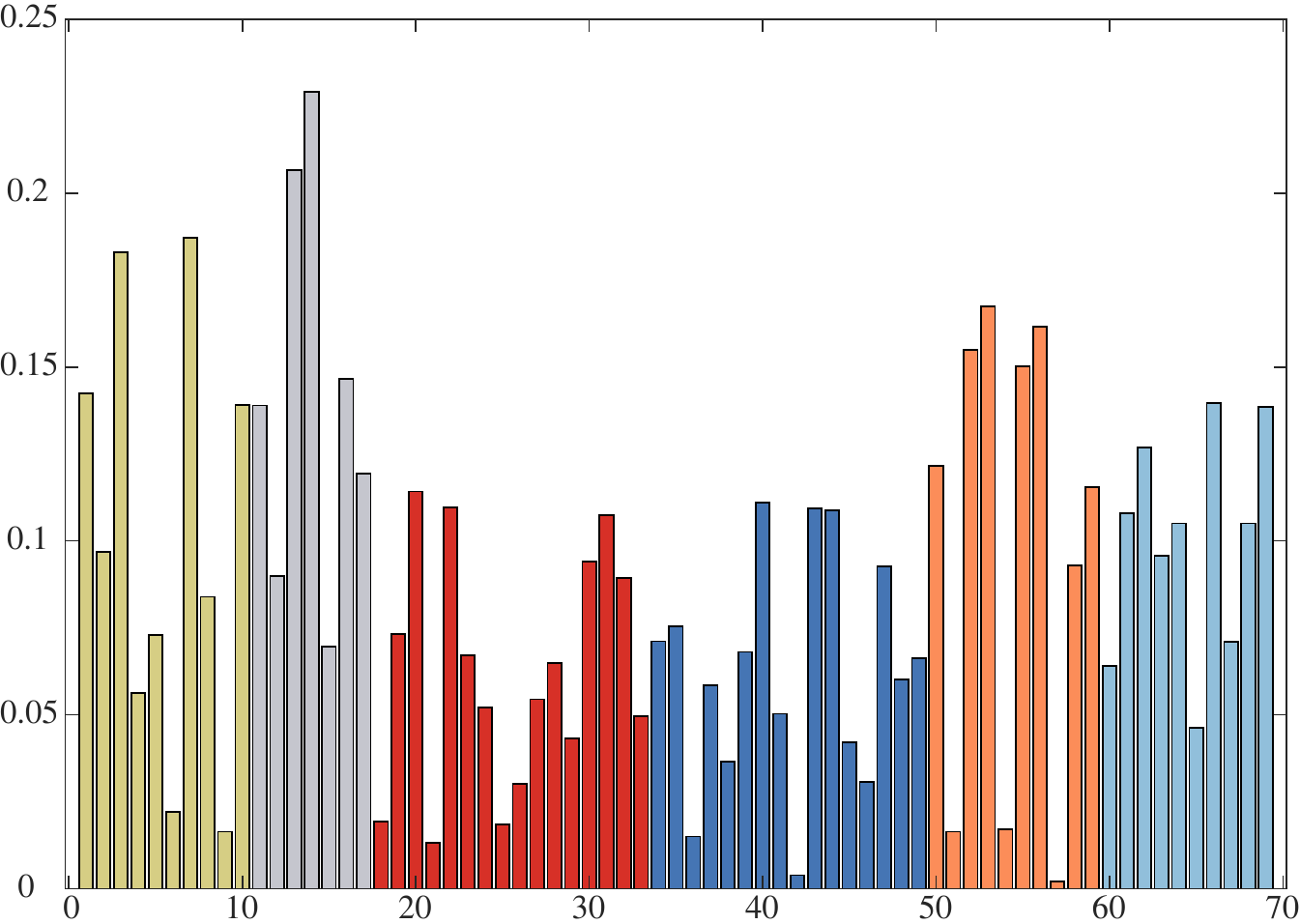}
   \caption{}
   \end{subfigure}%
   \begin{subfigure}{.5\textwidth}
   \centering
   \includegraphics[width=.99\textwidth]{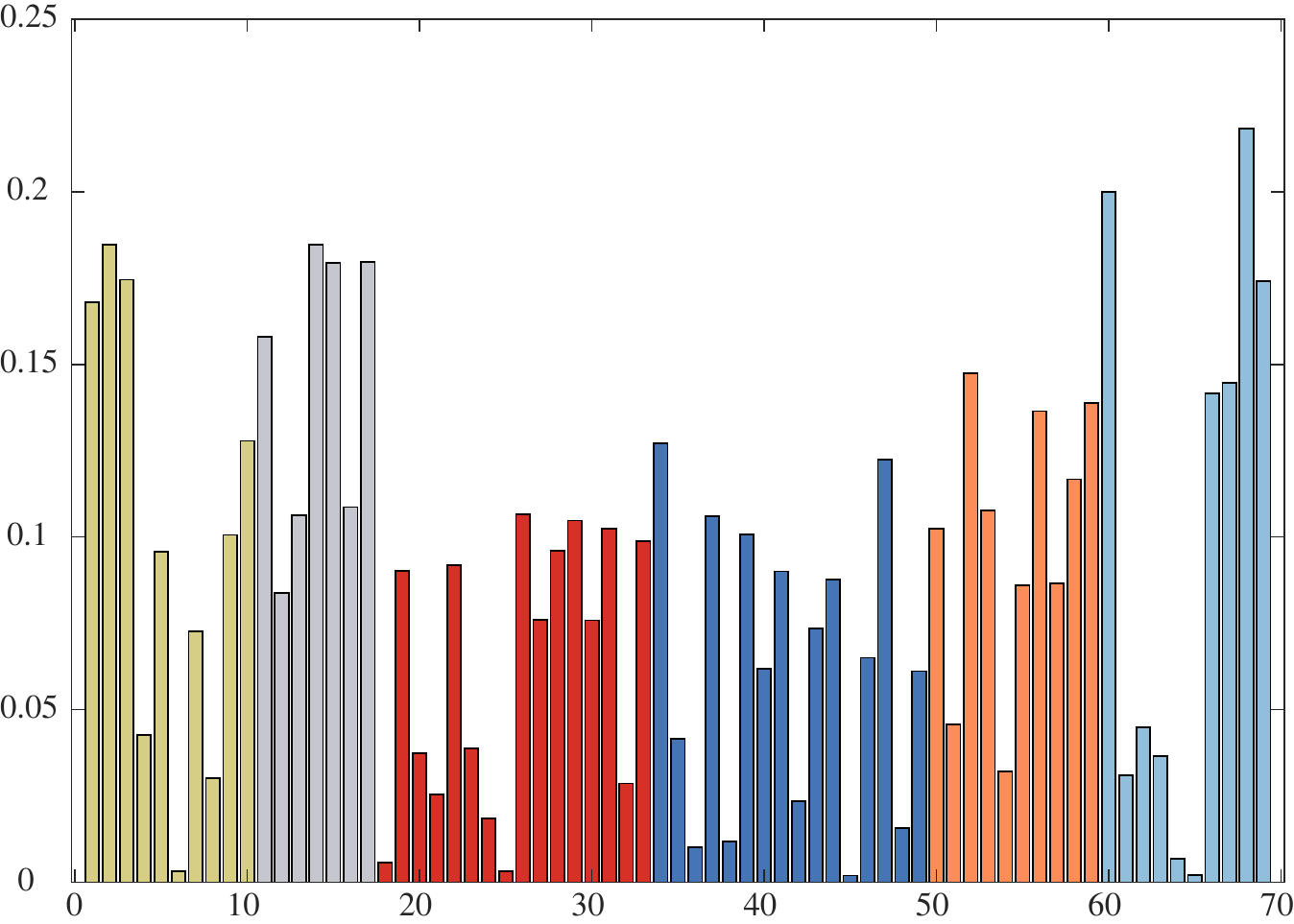}
   \caption{}
   \end{subfigure}
   \begin{subfigure}{.5\textwidth}
   \centering
   \includegraphics[width=.99\textwidth]{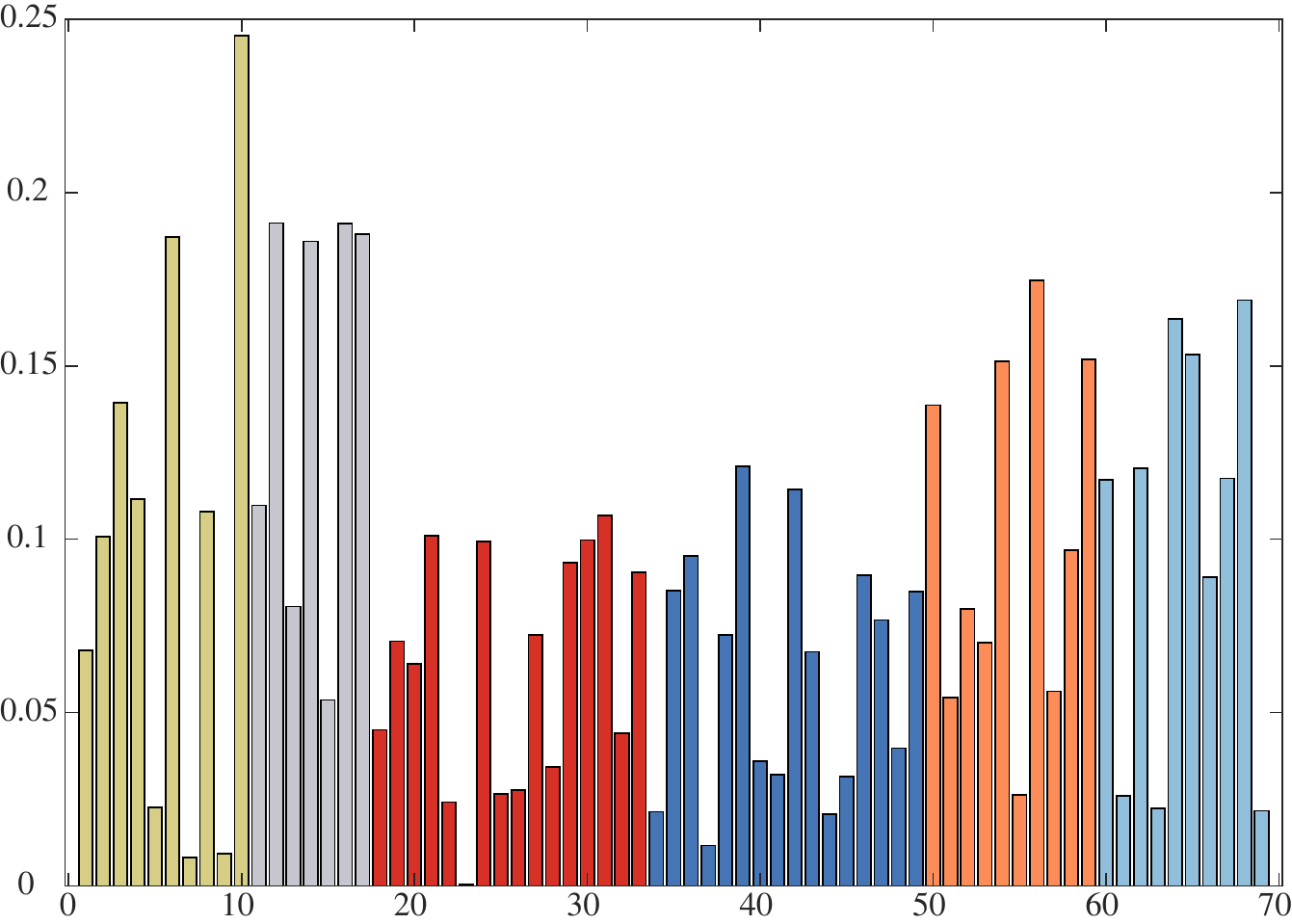}
   \caption{}
   \end{subfigure}
\caption{Distribution of the 69 primitives for the five most general categories of the ActivityNet dataset. Clock-wise from top-left: \emph{Eating and drinking Activities}; \emph{Sports, Exercise, and Recreation}; \emph{Socializing, Relaxing, and Leisure}; \emph{Personal Care}; \emph{Household Activities}. Each color corresponds to a different group following the convention of Fig.~\ref{fig:primitives}.}
\label{fig:ANet}
\end{figure}

 We examine the distribution of discovered motion primitives with respect to the activities been performed by the subjects. We perform our analysis on the sequences of the ActivityNet dataset. More specifically we use the 3D pose estimation algorithm of \cite{sanzari2016} on the video sequences of ActivityNet. We then extract motion primitives using the motion flux and perform recognition based on the extracted poses. We consider only the segments of the videos labeled with a corresponding activity. Additionally, we use only the segments were a single subject is detected and at least the upper body is visible. Fig.~\ref{fig:ANet} display the distribution of the motion primitives for the five most general activities according to the ActivityNet taxonomy.

\noindent
\subsection{ Motion Primitives Dataset}
 The dataset of annotated motion primitives extracted from the MoCap sequences of H3.6M \cite{ionescu2014}, CMU \cite{CMU} and KIT-WB \cite{mandery2015} has been made publicly available at \url{https://github.com/MotionPrimitives/MotionPrimitives}. 
 The dataset provides the start and end frames of each motion primitive together with the corresponding label as well as a reference to the MoCap sequence from which the motion primitive has been extracted.

{
\if\highlight1
\color{blue}
\fi
\typeout{----------------Changes--------------}
\subsection{Comparisons with state of the art on motion primitive recognition}

 We consider here the results of \cite{holte2010}, so far the only work providing quantitative results on human motion primitives, as far as we know. Here performance is evaluated for  $4$ actions  of the arms (gestures), namely \textit{Point right}, \textit{Raise arm}, \textit{Clap} and \textit{Wave}. The authors perform two tests, one without noise in the start and end frames of the primitives and one where the primitives are affected by noise. In the noise-free case their overall accuracy is 94.4\% while in the presence of noise the accuracy is 86.9\%.
 Our results are not immediately comparable with the ones of \cite{holte2010} since we use public datasets  (see above \S \ref{sec:datasets}, while they have built their own dataset, which is not publicly available. Furthermore, we have obtained by our classification process 16 primitives for each arm which are in accordance with biomechanics primitives. This notwithstanding,  we mapped their 22 primitives, denoted by the letters $A,\ldots,V$   to our defined primitives of the groups of {\em Left arm} and {\em Right arm} (see Table \ref{tab:comparisonHolte}).  To maintain the use of public datasets we have extracted videos from our reference datasets (see above \S \ref{sec:datasets}) to obtain the $4$ above mentioned gestures from 10 different subjects.  Hence, we have  computed the motion primitives  recognition accuracy on these video sets, to compare with \cite{holte2010}. The results are shown in Table \ref{tab:comparisonHolte}.
\typeout{*********************** TABLE}

\begin{table}[htbp]
  \centering
  \caption{Comparison with the 22 motion primitives of \cite{holte2010}}
  \resizebox{0.95\textwidth}{!}{
    \begin{tabular}{r|lllllll|c|c|}
\cline{3-10}    \multicolumn{1}{r}{} & \multicolumn{1}{c|}{} & \multicolumn{1}{p{2.945em}|}{Shoulder abd.} & \multicolumn{1}{p{2.89em}|}{Shoulder add.} & \multicolumn{1}{p{2.835em}|}{Elbow ext.} & \multicolumn{1}{p{2.22em}|}{Elbow flex.} & \multicolumn{1}{p{3.335em}|}{Shoulder Int. Rot. and elbow flex.} & \multicolumn{1}{p{3.055em}|}{Shoulder Ext. Rot. and elbow ext.} & \multicolumn{1}{p{2.78em}|}{Elbow Upper med. flex.} & \multicolumn{1}{p{2.22em}|}{Elbow Upper med ext.} \\
    \hline
    \multicolumn{1}{|l|}{A,B,C} & \multicolumn{1}{p{2.055em}|}{Point } & \multicolumn{1}{l|}{92.3} & \multicolumn{1}{l|}{{\bf 96.8}} & \multicolumn{1}{l|}{} & \multicolumn{1}{l|}{} & \multicolumn{1}{l|}{} &       &       &  \\
\cline{1-1}\cline{5-10}    \multicolumn{1}{|l|}{D,E,F} & \multicolumn{1}{p{2.055em}|}{right} & \multicolumn{1}{l|}{(89.6)} & \multicolumn{1}{l|}{(93.5)} & \multicolumn{1}{l|}{} & \multicolumn{1}{l|}{} & \multicolumn{1}{l|}{} &       &       &  \\
    \cline{1-2}\cline{4-10}
    \multicolumn{1}{|r}{} & \multicolumn{1}{r|}{} & \multicolumn{2}{c|}{\cellcolor[rgb]{ .906,  .902,  .902}82.5} & \multicolumn{1}{l|}{} & \multicolumn{1}{l|}{} & \multicolumn{1}{l|}{}        &  \multicolumn{1}{l|}{} & \multicolumn{1}{l|}{} &    \\
    \hline
    \multicolumn{1}{|l|}{G,H,I} & \multicolumn{1}{p{2.055em}|}{Raise } & \multicolumn{1}{l|}{} & \multicolumn{1}{l|}{} & \multicolumn{1}{l|}{84.5} & \multicolumn{1}{l|}{77.5} & \multicolumn{1}{l|}{} &       &       &  \\
\cline{1-1}\cline{3-4}\cline{7-10}    \multicolumn{1}{|l|}{J,K,L} & \multicolumn{1}{p{2.055em}|}{arm} & \multicolumn{1}{l|}{} & \multicolumn{1}{l|}{} & \multicolumn{1}{l|}{(81.4)} & \multicolumn{1}{l|}{(73.6)} & \multicolumn{1}{l|}{} &       &       &  \\
\cline{1-4}\cline{7-10}
    \multicolumn{4}{|r|}{}        & \multicolumn{2}{|c|}{\cellcolor[rgb]{ .906,  .902,  .902}{\bf 87.5}} & \multicolumn{1}{l|}{}  & \multicolumn{1}{l|}{} &          &  \\
    \hline
    \multicolumn{1}{|l|}{M,N,O} & 
\multicolumn{1}{p{2.055em}|}{Clap} & \multicolumn{1}{l|}{} & \multicolumn{1}{l|}{} & \multicolumn{1}{l|}{} & \multicolumn{1}{l|}{} & \multicolumn{1}{l|}{{\bf 91.7}} & 89.2  &       &  \\
\cline{1-1}\cline{3-10}    
\multicolumn{1}{|l|}{P,Q,R} & 
\multicolumn{1}{l|}{} & \multicolumn{1}{l|}{} & \multicolumn{1}{l|}{} & \multicolumn{1}{l|}{} & \multicolumn{1}{l|}{} & \multicolumn{1}{l|}{(87.6)} & (85.9) &       &  \\
    \hline
    \multicolumn{6}{|r|}{}& \multicolumn{2}{c|}{\cellcolor[rgb]{ .906,  .902,  .902}90.0} &   \multicolumn{1}{l|}{} &          \\
    \hline
    \multicolumn{1}{|l|}{S,T} & \multicolumn{1}{p{2.055em}|}{Wave} & \multicolumn{1}{l|}{} & \multicolumn{1}{l|}{} & \multicolumn{1}{l|}{} & \multicolumn{1}{l|}{} & \multicolumn{1}{l|}{} &       & \multicolumn{1}{l|}{85.4} & \multicolumn{1}{l|}{{\bf 87.7}} \\
\cline{1-1}\cline{3-10}  
\multicolumn{1}{|l|}{U,V} & \multicolumn{1}{l|}{} & \multicolumn{1}{l|}{} & \multicolumn{1}{l|}{} & \multicolumn{1}{l|}{} & \multicolumn{1}{l|}{} & \multicolumn{1}{l|}{} &   \multicolumn{1}{l|}{}    & \multicolumn{1}{l|}{(81.3)} & \multicolumn{1}{l|}{(82.9)} \\
\cline{1-8}
\multicolumn{8}{|r|}{}  & \multicolumn{2}{c|}{\cellcolor[rgb]{ .906,  .902,  .902}87.5} \\
\hline
\end{tabular}%
}
  \label{tab:comparisonHolte}%
\end{table}%



In Table \ref{tab:comparisonHolte} the capital letters in the first column  indicate the primitives in the language of \cite{holte2010}. In the second column are listed the actions formed by the primitives indicated in the first column. In the first row are indicated the primitive taken from our biomechanics language, which we mapped on the \cite{holte2010} primitives. Results are on the diagonal, in gray the results of \cite{holte2010}. We have indicated in parentheses the  values illustrated in the confusion matrices. While the values in the confusion matrices were  mean precision averages over all experiments for all actions in all the considered datasets, here the results are with respect to an amount of videos comparable to the experiments of \cite{holte2010}, hence they are significantly better for the indicated primitives. Despite the results are not quite comparable since we have measured our results on public databases, and in 3D, we can observe that our approach outperforms in all but one case the results in \cite{holte2010}. 

}

\subsection{Discussion}
  The results show that our framework discovers and recognizes motion primitives with high accuracy with respect to the manually defined baseline while providing competitive results with respect to \cite{holte2010}, the only work, to the best of our knowledge, providing quantitative results on similarly defined motion primitives. 
  
 Additionally, given the importance of studying human motion in a wide spectrum of research fields, ranging from robotics to bioscience, we believe that the human motion primitives dataset will be particularly useful in exploring new ideas and for enriching knowledge in these areas.

\section{An application of the motion primitives model to surveillance videos}\label{sec:application}
{
\if\highlight1
\color{blue}
\fi
In this section  we show how to set up an experiment by using motion primitives.
In particular, the application we have chosen is the detection in surveillance videos of dangerous human behaviors.
 To set up the experiment we consider videos of anomalous and dangerous  behaviors, and prove that idiosyncratic  primitives, among those identified in Figure \ref{fig:primitives},  appear to characterize these behaviors.
The application is quite interesting because  it highlights how the combination of  primitives allows to detect specific human behaviors. On the one side the motion primitives are used for detection and on the other side they can be used also   for characterizing classes of actions or classes of activities. 

\subsection{Related works and datasets on abnormal behaviors}
There is a significant amount of literature on {\em abnormality} detection in  surveillance videos. Only few of them, though, are concerned with dangerous behaviors. 
These methods can be further divided into those detecting dangerous crowd behaviors, in which the individual motion is superseded by large flows as in \cite{mohammadi2016,mohammadi2015,mousavi2015,hassner2012}, and those detecting closer dangerous human behaviors.

Among the latter there are methods focusing on fights \cite{gracia2015}, methods specialized on violence \cite{zhou2018,gao2016,deniz2014,xu2014}, on aggressive behaviors \cite{kooij2016}, and  on crime \cite{sultani2018}. 
 A review on  methods for detecting abnormal behaviors, taking into account some of the above mentioned ones, and  also discussing available datasets, is provided in \cite{mabrouk2018}. 

In the last years, also due to the above studies, a number of datasets have been created from real surveillance videos, or from movies repositories. The most used ones are  {\em UCSD Anomaly} \cite{mahadevan2010}, {\em Avenue Dataset} \cite{lu2013}, the {\em Behave} \cite{blunsden2010} dataset,  the {\em Violent Flows} dataset \cite{hassner2012}, the {\em Hockey Fight  Dataset}  \cite{nievas2011}, the {\em Movies Fight  Dataset} from \cite{nievas2011} too and, finally, the recent {\em UCF-crime} introduced by \cite{sultani2018}. To these datasets some authors, studying abnormal behaviors in surveillance videos,  have added specific activities from {\em UCF101}  \cite{soomro-ucf101}.

To detect dangerous behaviors we considered four of the above datasets most suitable for the task of analyzing human behaviors with small groups of subjects. The first dataset is the {\em Hockey Fight  Dataset} provided by \cite{nievas2011}, which is formed by 1000 clips of actions
from hockey games of the National Hockey League (NHL). A second dataset, also introduced by \cite{nievas2011} is the {\em Movies Fight  dataset}, which is composed of  200 video clips obtained from action movies, 100 of which show a fight. Videos in both these datasets are untrimmed but divided in those where there are fights and those where there are no fights. The third dataset is the {\em UCF-Crime dataset} introduced by \cite{sultani2018}. This dataset is formed by 1900 untrimmed surveillance videos of 13 realworld anomalies, including {\em abuse, arrest, arson, assault, road accident, burglary, explosion, fighting, robbery, shooting, stealing, shoplifting, and vandalism}, and normal videos. These videos have varying length from 30 sec. up to several minutes. In a  number of these videos, like explosion and road accident,  no  human behavior is observable. Among the others there are a number of videos not including  human behaviors. 
Therefore we have chosen a subset of all the UCF-crime dataset for both training and testing. In particular, we have chosen abuse, arrest,  assault, burglary,  fighting, robbery, shooting, stealing,  and vandalism.  Finally we have taken videos from {\em UCF101} dataset, which includes 101 human activities.

Given the above selected datasets we aim at showing that once the primitives are computed an off-the-shelf classifier can be used to detect specific behaviors, in this case the dangerous ones. 

The method we propose requires to compute the primitives on a selected training set, separating the untrimmed videos  with dangerous behaviors from the normal ones, as described below, and then training a non-linear kernel SVM on the two datasets, as illustrated in \S \ref{par:svm}. The trained classifier is then tested on the test sets and results are reported in \S \ref{par:test}, comparing with state of the art approaches. 

The main idea we want to convey here is that once primitives are computed all the relevant features for distinguishing a behavior are embedded in the primitive category of the specific group (see \S \ref{par:test}) and therefore the classifier has to deal just with them and not with other features such as poses, images, time and tracking, in so alleviating the classifier burden and allowing to deal with state of the art classifiers.
Furthermore, the primitive  parameters, used to estimate the primitive classes, are no more needed for the further classification of behaviors. This is  the main advantage of human motion primitives modeling, namely their effectiveness in characterizing specific behaviors. 

\subsection{Primitives computation}
For primitives computation we  collected all the videos from hokey and fight-movie datasets,  we collected from the UCF-crime dataset  the videos from  {\em abuse, arrest,  assault, burglary,  fighting, robbery, shooting, stealing},  and {\em vandalism}. Finally, from UCF101 we collected    276 videos from the datasets {\em Punch} and {\em SumoWrestling} and further 276 videos from other sports, randomly chosen as in \cite{gracia2015}. The total number of videos collected is 3050 for primitive computation, as illustrated in the following table:

\begin{table}[h]
  \centering
  \caption{Datasets for primitive computation in dangerous behaviors detection}
  \resizebox{0.95\textwidth}{!}{
    \begin{tabular}{|l|r|r|r|r|r|r|r|r|}
\cline{2-9}    \multicolumn{1}{l|}{ } & \multicolumn{2}{c|}{Hockey} & \multicolumn{2}{c|}{Fight-Movies} & \multicolumn{2}{c|}{UCF-crime} & \multicolumn{2}{c|}{UCF101} \\
\cline{2-9}    \multicolumn{1}{r|}{} & \multicolumn{1}{l|}{Danger.} & \multicolumn{1}{l|}{Normal} & \multicolumn{1}{l|}{Danger.} & \multicolumn{1}{l|}{Normal} & \multicolumn{1}{l|}{Danger.} & \multicolumn{1}{l|}{Normal} & \multicolumn{1}{l|}{Danger.} & \multicolumn{1}{l|}{Normal} \\
    \hline
    Video sets & 500   & 500   & 100   & 100   & 650   & 650   & 276   & 276 \\
    \hline
    Training & 70\%  & 70\%  & 70\%  & 70\%  & 70\%  & 70\%  & 100\% & 70\% \\
    \hline
    Test  & 30\%  & 30\%  & 30\%  & 30\%  & 30\%  & 30\%  & 0\%   & 30\% \\
    \hline
    \end{tabular}%
    }
  \label{tab:datasets}%
\end{table}%

To compute the primitives for each subject from a small group of people appearing in a frame of a video, we have fitted 3D poses basing on the SMPL model \cite{loper2015} of  {\em human mesh recovery} (HMR) \cite{kanazawa2018end}. HMR recovers together with joints and pose also a full 3D mesh from a single image (see Figures~\ref{fig:graphs} and \ref{fig:graphs2}), and it is accurate enough to estimate multiple subject poses in a single frame. 

Having more than a subject requires to  track each subject pose across frames, in order to compute the motion primitives for each of them. To this end we used the joints given by SMPL model in world frame, for the following body joints (see the preliminary Section \ref{sec:prim_prelim}): left and right {\em hip}, left and right {\em clavicle } (called shoulder in HMR), and the {\em head}. These joints are well suited for tracking since they have slower motion with respect to other body parts.  Tracking amounts to find the rotations and translations amid all the bodies appearing in two consecutive frames, and identifying the rotation and translation pertaining to each subject across the two frames. Consider two consecutive frames indexed by $t$ and $t{+}1$, and  let ${\mathcal J}^{(t)}=\{j_1^{(t)},\ldots,j_5^{(t)}\}$ and ${\mathcal J'}^{(t+1)}=\{{j'}_1^{(t+1)},\ldots,{j'}_5^{(t+1)}\}$ be the  joints in world frame of the above mentioned body components,  where  joint subscripts indicate in the order left and right hip, left and right clavicle and head.   We first find the translation ${\bf d}$ and rotation $R$ between any two set of joints appearing in the frames $t$ and $t{+}1$ 
(see also Section \ref{sec:prim_prelim}):
\begin{equation}\label{eq:transf}
\displaystyle{ (R,{\bf d}) = \argmin_{R\in SO(3),{\bf d}\in{\mathbb R}^3}\sum_{i=1}^5 w_i \|(R\ j^{(t)}_i+{\bf d})-  {j'}_i^{(t+1)}\|^2
}
\end{equation}
\noindent
With $w_i>0$ weights for each pair of joints in $(t)$ and $(t+1)$. Let $\hat{{\mathcal J}} =(\sum_{i=1}^5w_ij_i)/\sum_{i=1}^5w_i $ be the weighted centroids of the set of joints ${\mathcal J}$.  The  minimization in (\ref{eq:transf}) is solved by computing the singular value decomposition $U\Sigma V^{\top}$ of the covariance matrix $\bar{\mathcal J}^{(t)}W(\bar{{\mathcal J}'}^{(t+1)})^{\top}$ of the normalized joints $\bar{\mathcal J}^{(t)},\bar{{\mathcal J}'}^{(t+1)}$, obtained by subtracting the weighted centroid to each joints set. Here $W$ is the diagonal matrix of the weights $w_i$. Let $H$ be the diagonal matrix $diag({\bf 1},det(VU^{\top}))$, then the rotations and translations between sets of joints are found as:
\begin{equation}\label{eq:finalRt}
R = VHU^{\top}\ \ \mbox{ and } {\bf d}=\hat{{\mathcal J}'}^{(t+1)}-R\hat{{\mathcal J}}^{(t)}
\end{equation}

 Finally, once we have obtained the rotation matrices and the translation vectors between the sets of considered joints of all the fitted skeletons, from frame $t$ to frame $t+1$, we can track each individual skeleton $S_k$. A skeleton $S_k^{(t+1)}$ belongs to the same subject fitted  by skeleton $S_k^{(t)}$, at frame $t$,  if  the rotation $R_k$ and translation ${\bf d}_k$, obtained according to eq. (\ref{eq:finalRt}) between the chosen joints ${\mathcal J}^{(t)}$ of $S_k^{(t)}$  and ${\mathcal J'}^{(t+1)}$ of $S_k^{(t+1)}$, satisfy
\begin{equation}
(R_k, {\bf d}_k) = \argmin_{{R_k}\in SO(3),{\bf d}_k\in {\mathbb R}^3,k=1:s} \|{{\mathcal J}'}^{(t+1)}- (({\mathcal J}^{(t)} R_k)^{\top} + {\bf d}_k)^{\top}\|_{F}
\end{equation}
\noindent
With $\|\cdot\|_F$ the Frobenious norm and $s = N_S!{/}((N_S{-}2)!2!)$, with $N_S$ the common number of fitted skeletons $S$ in both frame $t$ and $t+1$.

Once the skeletons are tracked we can compute the unknown primitives from the flux (see Section \ref{sec:flux}) as paths $\gamma_{G_m}^T:I\subset {\mathbb R}\mapsto {\mathbb R}^9$, for each group $G_m$, with $I$ the time interval, specified by the frame sequence, and scale it as described in Section \ref{sec:flux}. We can then use the parameters $\Theta$ learned with the recognition model, detailed in \S\ref{subsect:models}, to assign a label ${\mathcal L}_w^{G_m}$ to each primitive segmented by the motion flux as precised  in eq. (\ref{eq:best}). Namely, we find the model identified by the parameter $\Theta_w$, which maximizes the probability of the primitive under consideration. We recall that for each group $G_m$, $m=1,\ldots,6$ there are $q$ models with $q\in\{7,10,16\}$ (see the primitives representation in Figure \ref{fig:primitives}). 

Our model of motion primitives relies significantly on the accuracy of the 3D pose estimation. We have chosen the model HMR \cite{kanazawa2018end}  based on SMPL \cite{loper2015}, in place of \cite{natola2015,tome2017}, since it is most recent and highly accurate. Still not all the videos chosen obtain a reasonable fitting, therefore after skeleton fitting and tracking   a number of videos from  UCF-crime have been removed from the considered set.

\subsection{Training a non-linear binary classifier}\label{par:svm}
All the computed primitives are labeled by their name (e.g. {\em Elbow flex}), according to the recognition model, as specified above.   
A set of primitives for a given video is formed as follows. Primitive names are embedded into real numbers $r\sim Unif(0,1)$,  such that for each primitive name there is a precise real number. 
Given frame $t$ for each skeleton appearing in the frame we form a vector of dimension $6\times 1$, where the  6 elements are the corresponding embedded primitive names occurring at frame $t$. Let $\gamma_{G_m}^{(t)}$ denote the primitive of the body group $G_m$, and $u$ the mapping of the primitive name to the real number:
\begin{equation}
{\bf x}_j^{(t)} = (u(\gamma_{G_1}^{(t)}), u(\gamma_{G_2}^{(t)}),\ldots,u(\gamma_{G_6}^{(t)}))^{\top}
\end{equation}
Where $j$ indicates the $j$-th skeleton appearing in frame $t$. Note that $t$ and $j$ are actually indicated just for forming the training set, to select from all the gathered vectors ${\bf x}$ those that have changing primitives. Namely, for training, from the set of all vectors in each frame, we have retained only those vectors in which at least one primitive changes, for each recorded skeleton.

For training we have selected  videos for both dangerous behaviors and normal behaviors, thus labeling them with $1$ for dangerous and ${-}1$ for normal behaviors, as follows. We selected $70\%$ of fighting and $70\%$ of not fighting from both hockey and fight movies; from UCF101 we have selected all videos in {\em Punch} and {\em SumoWrestling}, getting $276$ videos and further  $276$ videos randomly from sport activities. For UCF-crime we proceeded as follows. We have selected the  videos from all the crime activities specified above with time length less than $60 sec.$ and cropped the first and last $10 sec.$, in order to do a weak supervised training, namely,  as in \cite{sultani2018} we have not trimmed the video. Thus we obtained $173$ videos for abnormal activities and  we selected $173$ videos from the normal activities. The total number of videos for training is $1634$ videos. All the remaining video with computed primitives have been used for testing. 

 The resulting data structure is:
\begin{equation}
\{({\bf x}_1,y_1),\ldots, ({\bf x}_n,y_n)\} \mbox{\   with  } {\bf x}\in {\mathbb R}^6, y\in \{{-}1,1 | \mbox{ ${-}1$ if normal, $1$ if dangerous}\}
\end{equation}

The SVM \cite{vapnik2013} is a popular classification method computing, for  two non-separable classes, the classifier:
\begin{equation}\label{eq:dp}
\begin{array}{l}
 f({\bf x})= \left(\sum_{i=1}^n y_i\alpha_i K({\bf x}_i,{\bf x})+b \right)\\
\hat{y} = sgn(f({\bf x}))
\end{array}
\end{equation}
where $K$ is the kernel function $\varphi({\bf x}_i)^{\top}\varphi({\bf x}_j)$ with $\varphi$ the feature map, here we considered the RBF kernel 
$\exp\left(-\eta \|{\bf x}_i-{\bf x}_j\|_{\ell_2}^2\right)$, with $\eta$ a tunable parameter. 
Classification is obtained by solving the 
constrained optimization problem: 
\begin{equation}
\max_{\alpha} \frac{1}{2} {\boldsymbol \alpha}^{\top} \Omega {\boldsymbol \alpha }- {\bf e}^{\top}{\boldsymbol \alpha}
\mbox{\ \ \  subject to \ \ } {\bf y}^{\top}\alpha =0, \ 0{\leq}\alpha_i{\leq}\lambda
\end{equation}
\noindent
Here $\Omega$ is a square  $n\times n$ positive semidefinite matrix, with $\omega_{i,j} = y_iy_jK({\bf x}_i,{\bf x}_j)$,  ${\bf e}$ is a vector of ones, the non zero $\alpha_i$ define the support vectors, and $\lambda$ is the regularization parameter of the primal optimization problem $min_{{\bf w},b,\xi}\frac{1}{2}{\bf w}{\bf w}^{\top}+\lambda\sum_{i=1}^n\xi_i$ \cite{scholkopf2001}. 
To obtain posterior probabilities we applied the Platt scaling \cite{platt1999}, proposing a sigmoid model to fit a posterior on the SVM output:
\begin{equation}\label{eq:prob}
P(y=1|f({\bf x})) = \frac{1}{1+\exp(Af({\bf x})+B)}
\end{equation}
\noindent Here the parameters $A$ and $B$ are fitted by solving the maximum likelihood problem:
\begin{equation}
min_{z=(A,B)} F(z)=-\sum_{i=1}^n\left(t_i \log(p_i)+(1-t_i)\log(1-p_i)\right)
\end{equation}
\noindent
Using  as prior the number of positive $N_+$ and negative $N_-$ examples in the training data, with $p_i=P(y=1| f({\bf x}_i))$, $t_i=(N_++1)/(N_++2)$ if $y_i=1$ and $1/(N_-+2)$ if $y_i={-}1$. See also \cite{lin2007note} for an improved algorithm with respect to \cite{platt1999}. 

To obtain the probability that at a given frame $t$ a dangerous event occurs we compute the average response to the primitives of each subject which has been detected. 
More precisely, let $s$ be the number of subjects in frame $t$ for which the primitives are computed, then the observation ${\bf x}^{(t)} = ({\bf x}_1^{(t)}, \ldots, {\bf x}_s^{(t)})$. Given ${\bf x}^{(t)}$, and assuming that the SVM scores for each ${\bf x}_i^{(t)}$ are independent, we can define the probability that a dangerous event $Y$ is occurring at $t$,  in a surveillance video,
as the expectation:
\begin{equation}\label{eq:probcrime}
P(Y | {\bf x}^{(t)}) = \sum_{i=1}^s p(\hat{y}_i^{(t)}|{\bf x}_i^{(t)}) P(y_i=1| f({\bf x}_i^{(t)}))
\end{equation}
Here $p(\hat{y}^{(t)}|{\bf x}^{(t)})$ is computed by remapping the scores to $[0,1]$ such that $\sum_{i=1}^s p(\hat{y}_i^{(t)}|{\bf x}_i^{(t)})=1$.
Testing has been done on the videos on which the primitives have been precomputed, and the results are shown together with  comparisons with the state of the art  in \S \ref{par:test}. Note that the method is not yet suitable for online detection of dangerous behaviors, still it can be advanced to online detection, by lifting the computation of the flux with  motion  anticipation.

\subsection{Results and comparisons with the state of the art}\label{par:test}
 We discuss now the results achieved by our method for abnormal behavior detection based on human motion primitives. 
 Figure~\ref{fig:graphs} shows some qualitative results of dangerous behaviors detection  in four videos. Three videos correspond to crime activities, namely {\em Abuse}, {\em Fighting} and {\em Shooting}, while the last displays a normal activity. The curve plotted in the graphs provides for each frame the probability that a dangerous event is occurring, according to eq. (\ref{eq:probcrime}). The highlighted region corresponds to the interval where a crime activity occurs. From this graphs it is evident that the crime activity detection follows closely the ground truth. For each example we also show two representative frames overlaid with the human meshes identified by HMR. Similarly, Figure~\ref{fig:fights} shows some representative examples of fitted human meshes for videos taken from Hockey and Movie Fights datasets.
 
\begin{figure}[tp!]
\centering
\captionsetup[subfigure]{justification=centering}
   \begin{subfigure}{1\textwidth}
   \centering
   \includegraphics[width=.95\textwidth]{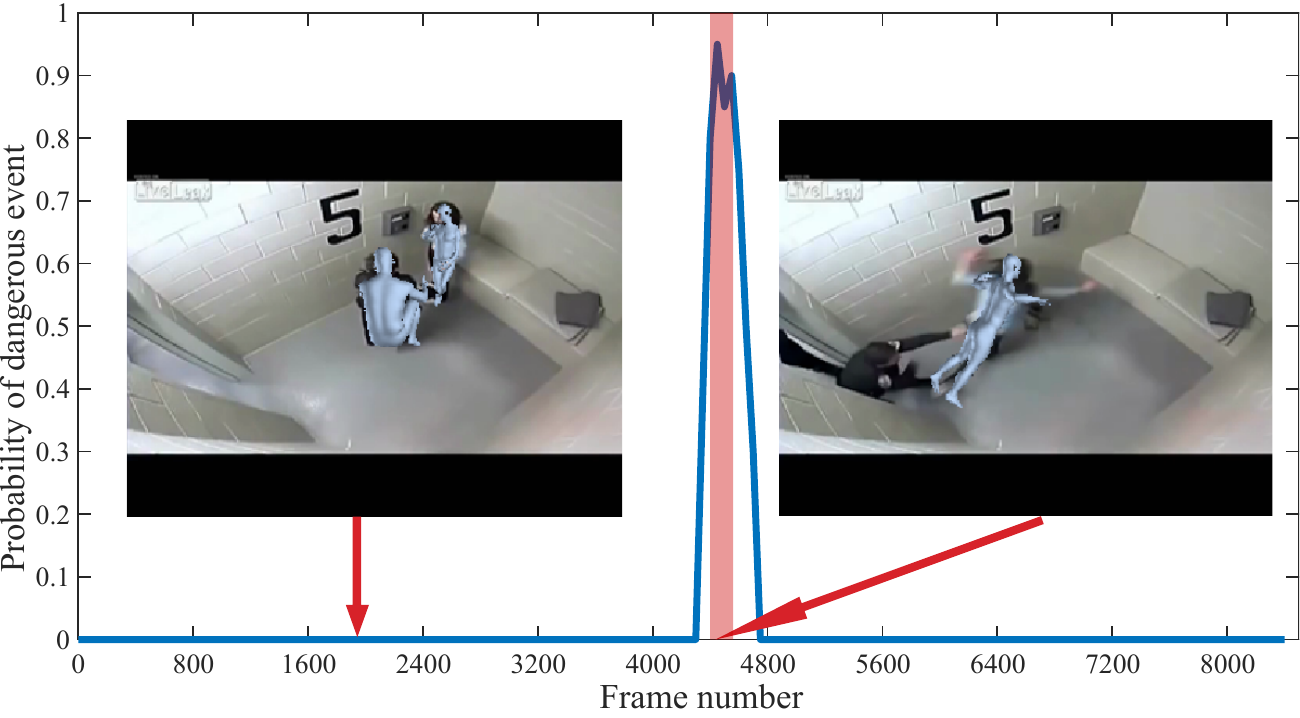}
   \caption{}
   \end{subfigure}
   \begin{subfigure}{1\textwidth}
   \centering
   \includegraphics[width=.95\textwidth]{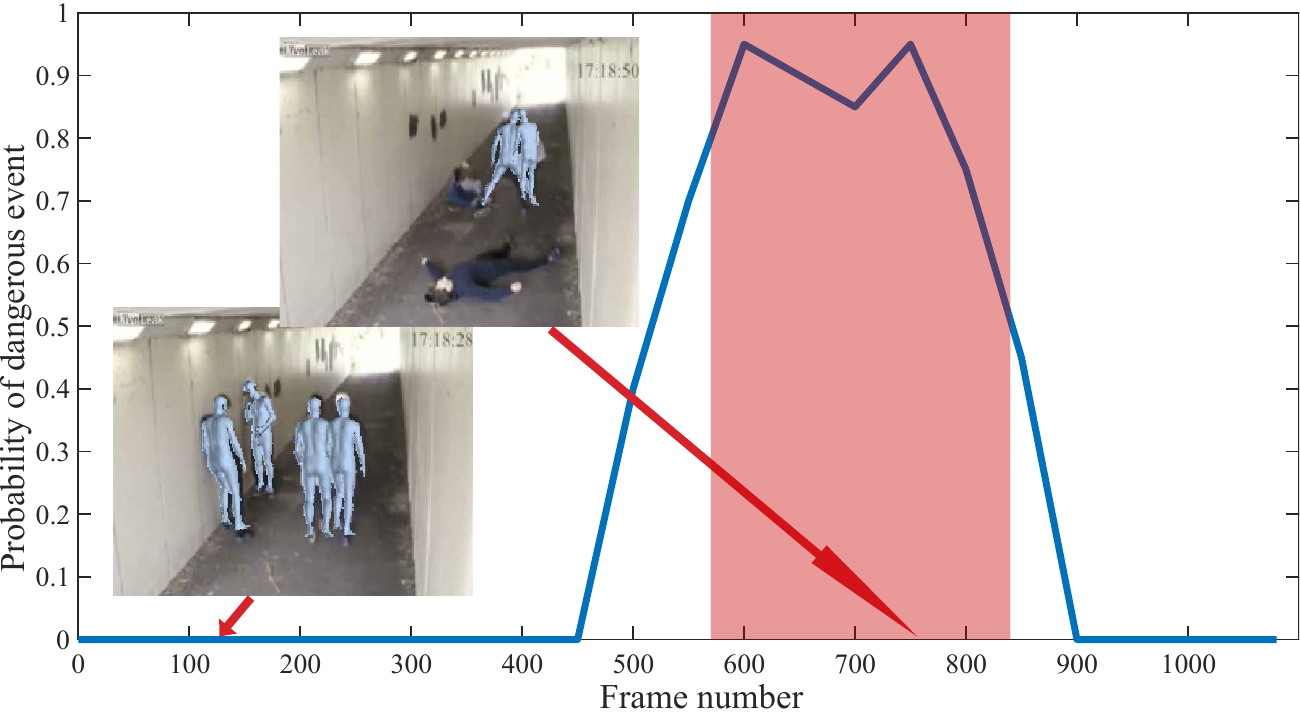}
   \caption{}
   \end{subfigure}
   \caption{Results of the proposed method on videos from UCF-Crime dataset. From top: {\em Abuse}, {\em Fighting}.  Colored window shows ground truth anomalous region.}\label{fig:graphs}
\end{figure}

\begin{figure}[tp!]
\centering
\captionsetup[subfigure]{justification=centering}
      \begin{subfigure}{1\textwidth}
   \centering
   \includegraphics[width=.95\textwidth]{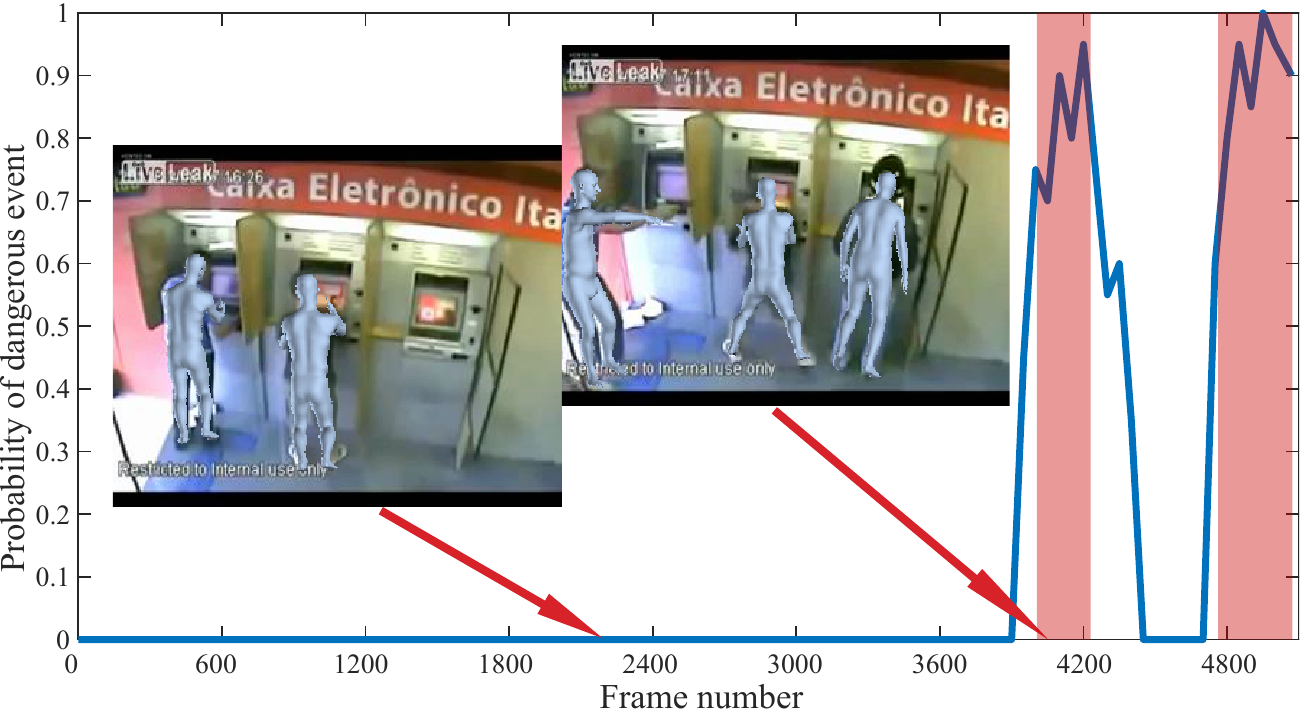}
   \caption{}
   \end{subfigure}
      \begin{subfigure}{1\textwidth}
   \centering
   \includegraphics[width=.95\textwidth]{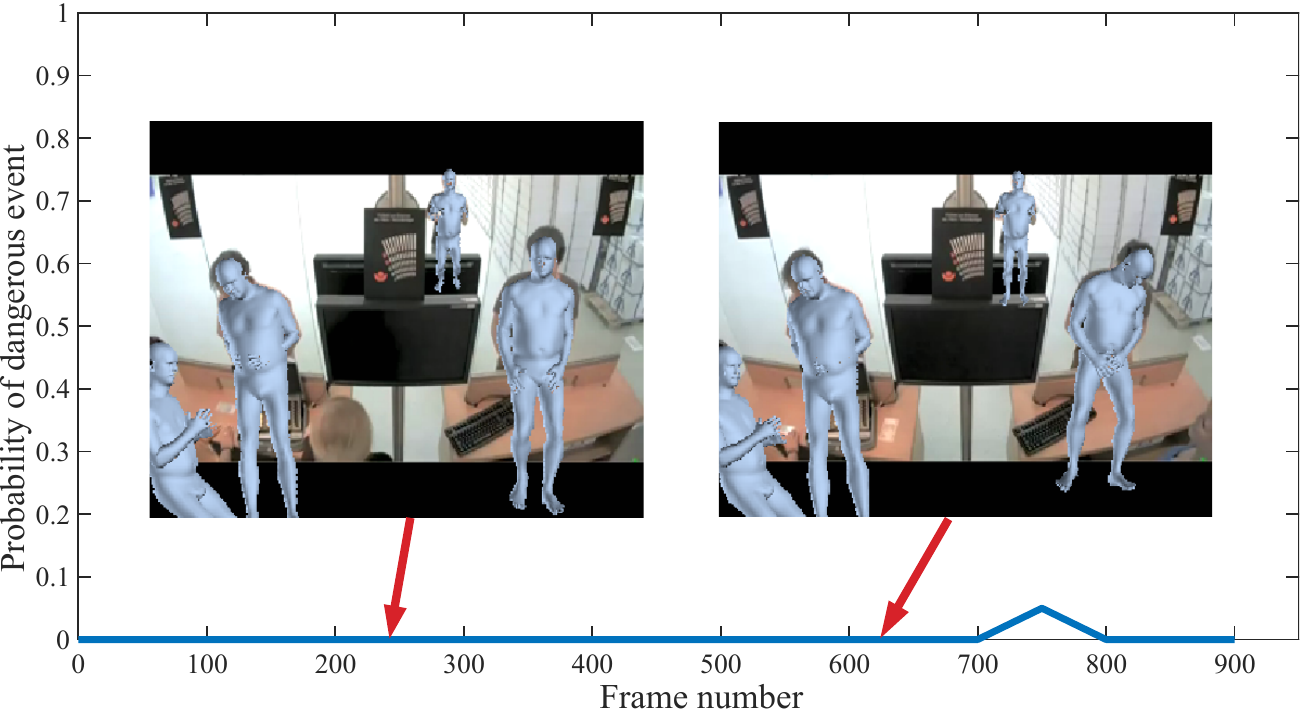}
   \caption{}
   \end{subfigure}
\caption{Results of the proposed method on videos from UCF-Crime dataset. From top: {\em Shooting}, {\em Normal}.  Colored window shows ground truth anomalous region.}\label{fig:graphs2}
\end{figure}

\begin{figure}[t!]
\centering
\captionsetup[subfigure]{justification=centering}
 \begin{subfigure}{.5\textwidth}
   \centering
   \includegraphics[width=.99\textwidth]{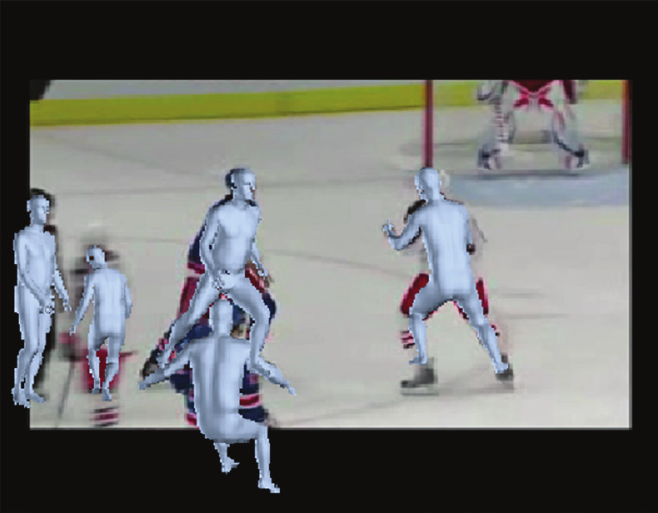}
   \caption{}
   \end{subfigure}%
   \begin{subfigure}{.5\textwidth}
   \centering
   \includegraphics[width=.99\textwidth]{crime/hfight2}
   \caption{}
   \end{subfigure}
   \begin{subfigure}{.5\textwidth}
   \centering
   \includegraphics[width=.99\textwidth]{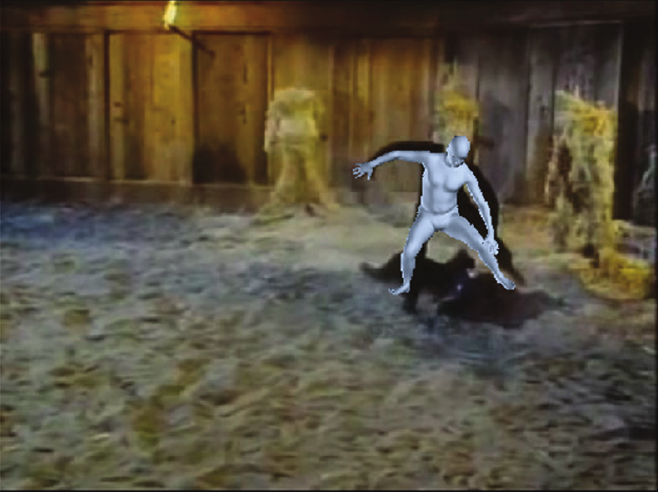}
   \caption{}
   \end{subfigure}%
   \begin{subfigure}{.5\textwidth}
   \centering
   \includegraphics[width=.99\textwidth]{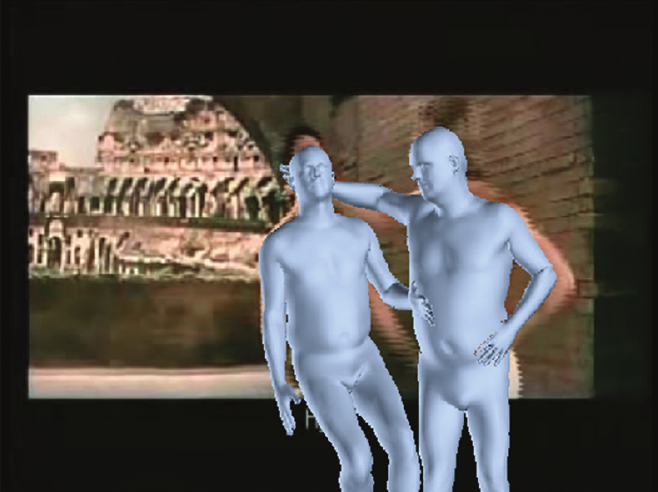}
   \caption{}
   \end{subfigure}
\caption{Instances of videos with human meshes fitted using HMR from Hockey and Movies datasets \cite{nievas2011}.}\label{fig:fights}
\end{figure}

\begin{figure}[t!]
\centering
\captionsetup[subfigure]{justification=centering}
 \begin{subfigure}{.5\textwidth}
   \centering
   \includegraphics[width=.99\textwidth]{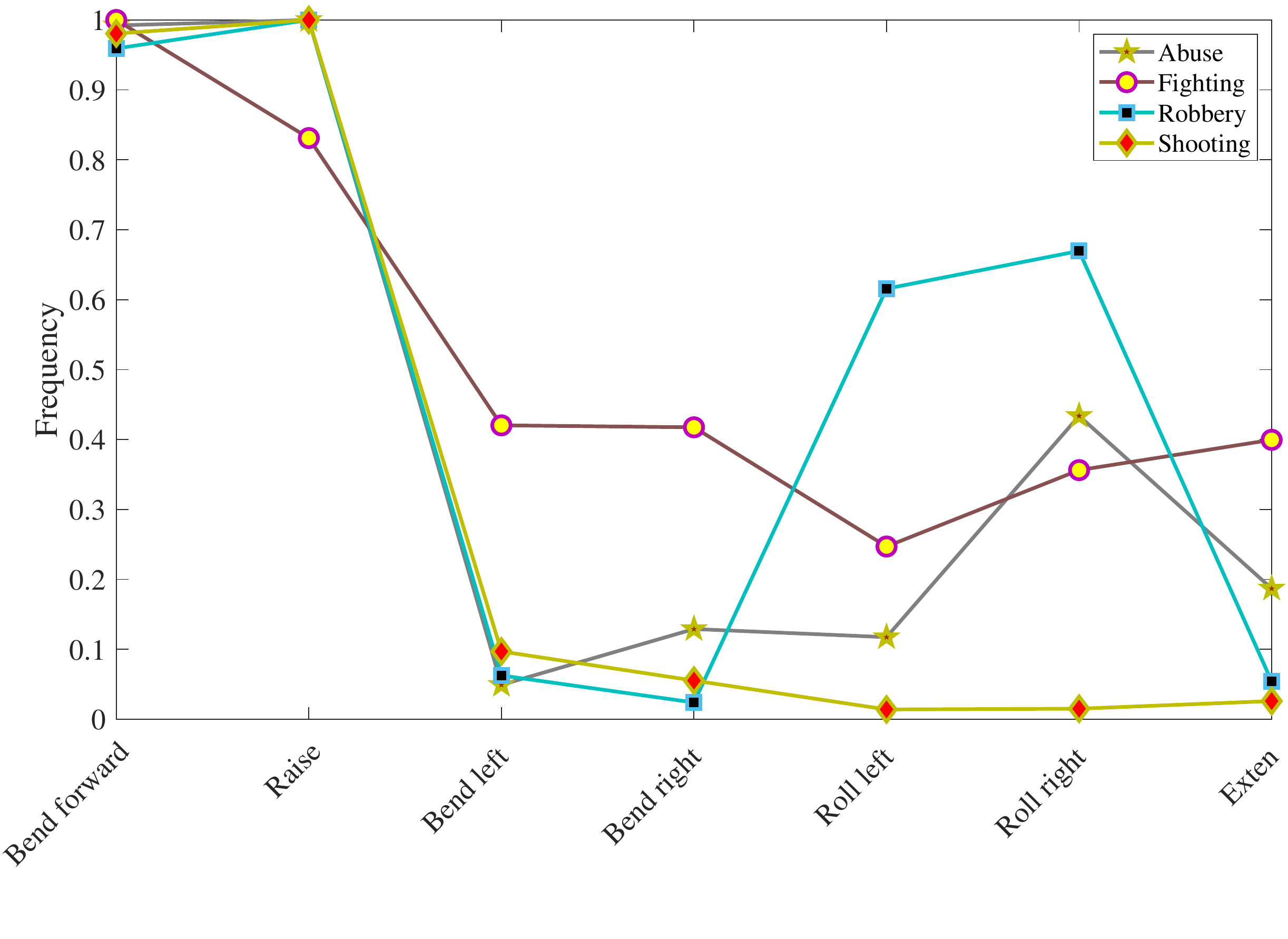}
   \caption{Primitives of {\em torso}, group $G_2$.}
   \end{subfigure}%
    \begin{subfigure}{.5\textwidth}
   \centering
   \includegraphics[width=.99\textwidth]{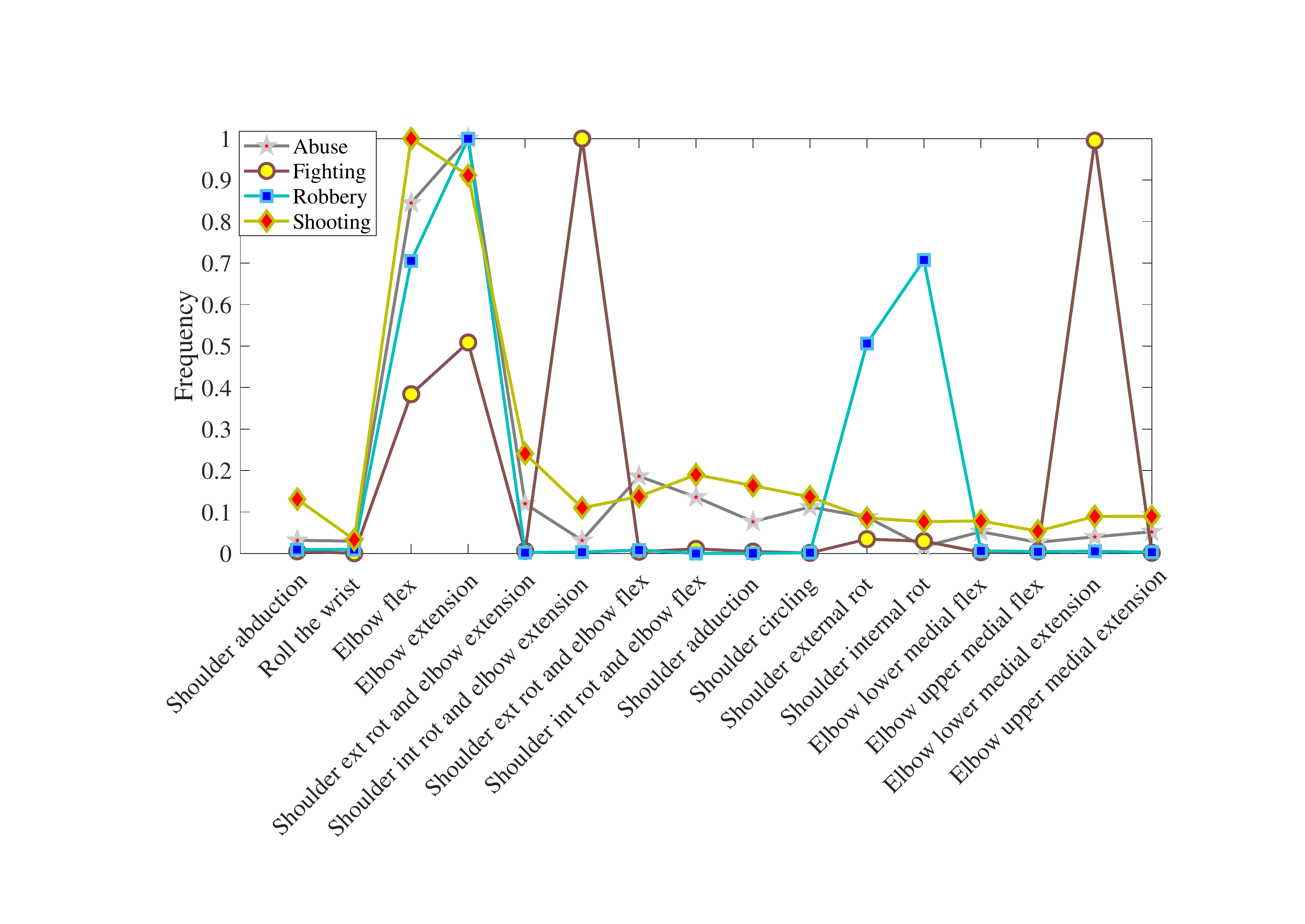}
   \caption{Primitives of {\em right arm}, group $G_3$.}
   \end{subfigure}
\caption{Frequency graphs of the occurrences of primitives for groups $G_2$ (torso) and $G_3$ (right arm) in the videos of {\em Abuse}, {\em Fighting}, {\em Robbery}, and  {\em Shooting}of the dataset UCF-crime.}\label{fig:histos}
\end{figure}

 Fig.~\ref{fig:rocs} presents the ROC curves of the proposed method for the four datasets considered, namely UCF-Crime, UCF101, Hockey Fights and Movie Fights. The corresponding values of the area under curve (AUC) are 76.15\%, 91.92\%, 98.44\% and 98.77\%, respectively. 
  Table~\ref{tab:compfight} presents the mean accuracy, its standard deviation and the area under the receiver-operating-characteristic (ROC) curve of our method in comparison with other state-of-the-art methods. The results of the other methods are taken from \cite{gracia2015}. We observe that our method achieves better performance on the Hockey Fights and Movies Fights datasets while it has very similar performance with the best performing method on the UCF101 dataset.

\begin{table}[t!]
\caption{AUC comparison with state-of-the-art methods on the UCF-Crime dataset.}\label{tab:compcrime}
\resizebox{0.95\textwidth}{!}{
\begin{tabular}{c|cccccc}
\textbf{Method} & Binary classifier & Hasan et al. \cite{Hasan_2016_CVPR} & Lu et al. \cite{Lu_2013_ICCV} & \cite{sultani2018} & \cite{sultani2018} w. constraints & Ours \\ \hline
\textbf{AUC} & 50.0  & 50.6  & 65.51 & 74.44  & 75.41 & \textbf{76.15}
\end{tabular}}
\end{table}

\begin{figure}[t!]
\includegraphics[width=0.75\textwidth]{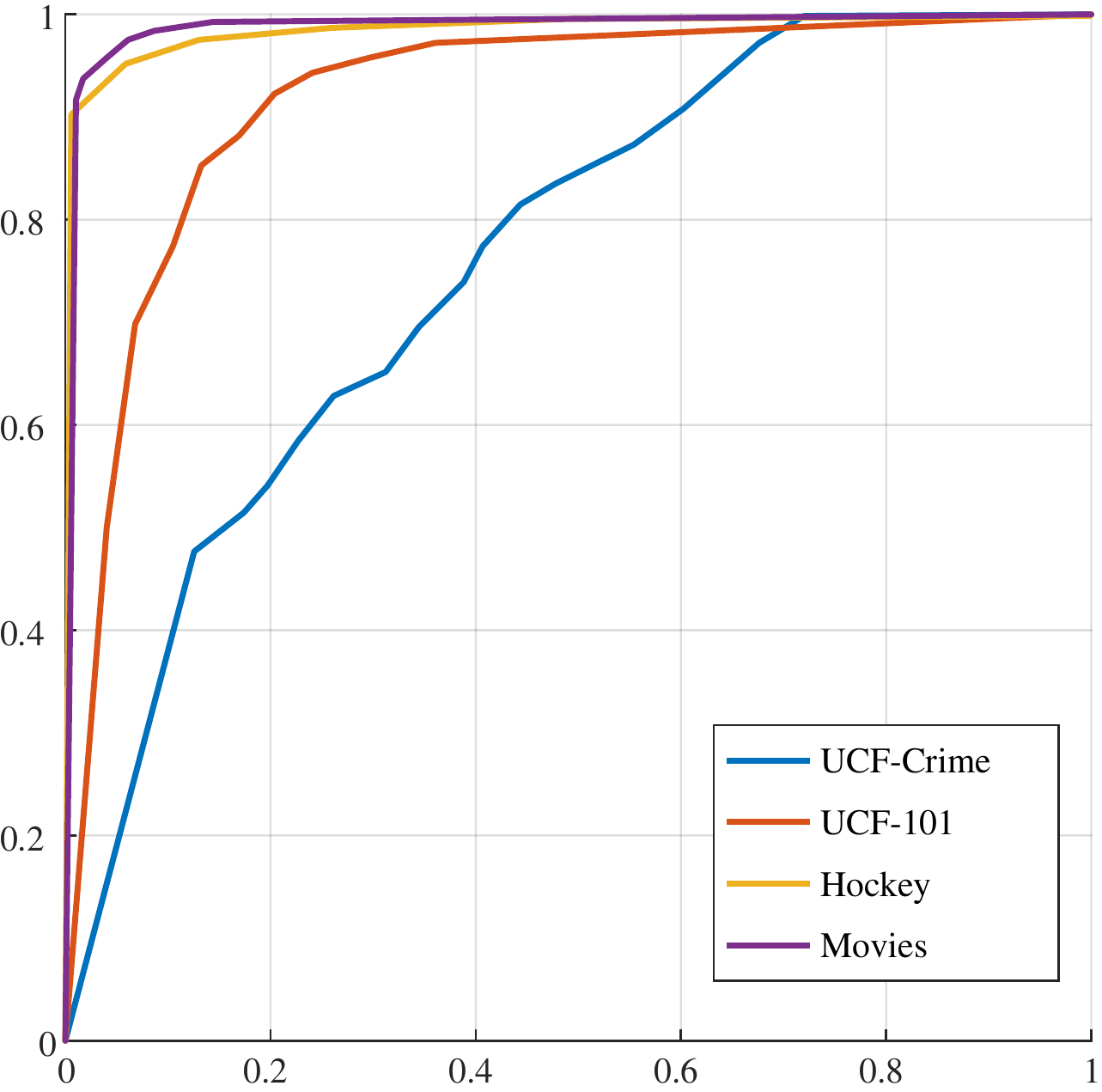}
\caption{ROC curves of the proposed method for UFC-Crime, UFC101, Hockey and Movies datasets.}\label{fig:rocs}
\end{figure}

\begin{table}[t!]
\caption{Comparison with state-of-the-art methods on the datasets Movies, UCF101 and Hockey.}\label{tab:compfight}
\begin{tabular}{lllll}
\textbf{Method} & \textbf{Classifier} & \multicolumn{3}{c}{\textbf{Datasets}} \\
 & & \textbf{Movies} & \textbf{Hockey} & \textbf{UCF101} \\ \hline \\
BoW (STIP) 	& SVM 		& 82.3$\pm$0.9/0.88  & 88.5±0.2/0.95 & 72.5±1.5/0.74 \\
			& AdaBoost 	& 75.3±0.83/0.83 & 87.1$\pm$0.2/0.93 & 63.1$\pm$1.9/0.68 \\
			& RF 		& 97.7$\pm$0.5/0.99  & 96.5$\pm$0.2/0.99 & 87.3$\pm$0.8/0.94 \\
BoW (MoSIFT)& SVM 		& 63.4$\pm$1.6/0.72  & 83.9$\pm$0.6/0.93 & 81.3$\pm$ 1/0.86  \\
			& AdaBoost 	& 65.3$\pm$2.1/0.72  & 86.9$\pm$1.6/0.96 & 52.8$\pm$3.6/0.62 \\
			& RF		& 75.1$\pm$1.6/0.81  & 96.7$\pm$0.7/0.99 & 86.3$\pm$0.8/0.93 \\
ViF 		& SVM 		& 96.7$\pm$0.3/0.98  & 82.3$\pm$0.2/0.91 & 77.7$\pm$2.16/0.87\\
			& AdaBoost  & 92.8$\pm$0.4/0.97  & 82.2$\pm$0.4/0.91 & 78.4$\pm$1.7/0.86 \\
			& RF 		& 88.9$\pm$1.2/0.97  & 82.4$\pm$0.6/0.9  & 77$\pm$1.2/0.85   \\
LMP 		& SVM 		& 84.4$\pm$0.8/0.92  & 75.9$\pm$0.3/0.84 & 65.9$\pm$1.5/0.74 \\
			& AdaBoost  & 81.5$\pm$2.1/0.86  & 76.5$\pm$0.9/0.82 & 67.1$\pm$1/0.71   \\
			& RF		& 92$\pm$1/0.96 	 & 77.7$\pm$0.6/0.85 & 71.4$\pm$1.6/0.78 \\
\cite{deniz2014}& SVM 		& 85.4$\pm$9.3/0.74  & 90.1$\pm$0/0.95 	 & \textbf{93.4$\pm$6.1/0.94} \\
			& AdaBoost  & 98.9$\pm$0.22/0.99 & 90.1$\pm$0/0.90 	 & 92.8$\pm$6.2/0.94 \\
			& RF 		& 90.4$\pm$3.1/0.99  & 61.5$\pm$6.8/0.96 & 64.8$\pm$15.9/0.93\\
\cite{gracia2015} v1 & SVM 	& 87.9$\pm$1/0.97 	 & 70.8$\pm$0.4/0.75 & 72.1$\pm$0.9/0.78 \\
			& AdaBoost  & 81.8$\pm$0.5/0.82  & 70.7$\pm$0.2/0.7  & 71.7$\pm$0.9/0.72 \\
			& RF 		& 97.7$\pm$0.4/0.98  & 79.3$\pm$0.5/0.88 & 74.8$\pm$1.5/0.83 \\
\cite{gracia2015} v2 & SVM	& 87.2$\pm$0.7/0.97  & 72.5$\pm$0.5/0.76 & 71.2$\pm$0.7/0.78 \\
			& AdaBoost  & 81.7$\pm$0.2/0.82  & 71.7$\pm$0.3/0.72 & 71$\pm$0.8/0.72   \\
			& RF 		& 97.8$\pm$0.4/0.97  & 82.4$\pm$0.6/0.9  & 79.5$\pm$0.9/0.85 \\
Ours		& SVM		& \textbf{99.1$\pm$0.3/0.99} & \textbf{97.2$\pm$0.8/0.98}  & 93.3$\pm$2.1/0.92 \\
\end{tabular}
\end{table}

  Additionally, in Figure~\ref{fig:histos} we present the frequency graphs of primitive occurrences for groups G2 and G3, for the crime activities {\em Abuse}, {\em Fighting},  {\em Robbery}, and {\em Shooting}.  The graphs show that each type of activity manifests itself by a different combination of idiosyncratic motions of the limbs. This fact can be used to achieve finer grained categorization of the crime activities, however, we do not examine further this possibility in this work.
  
 Figure~\ref{fig:rocs} presents the ROC curves of the proposed method for the four datasets considered, namely UCF-Crime, UCF101, Hockey Fights and Movie Fights. The corresponding values of the area under curve (AUC) are 76.15\%, 91.92\%, 98.44\% and 98.77\%, respectively. 
  Table~\ref{tab:compfight} presents the mean accuracy, its standard deviation and the area under the receiver-operating-characteristic (ROC) curve of our method in comparison with other state-of-the-art methods. The results of the other methods are taken from \cite{gracia2015}. We observe that our method achieves better performance on the Hockey Fights and Movies Fights datasets while it has very similar performance with the best performing method on the UCF101 dataset.
  
  Finally, Table~\ref{tab:compcrime} gives a comparison of the results achieved by our method on the UCF-Crime dataset in comparison with results from other state-of-the-art methods as reported in \cite{sultani2018}. In this case we have to highlight that our results are not directly comparable with the ones reported in \cite{sultani2018} as we restrict our analysis on videos where human subjects are visible. Nevertheless, the results indicate that also on this database the proposed method is able to achieve state-of-the-art performance on crime activity detection. 
}

\section{Conclusions}\label{sec:prim_conclusion}
We presented a framework for automatically discovering and recognizing human motion primitives from video sequences based on the motion of groups of joints of a subject.  To this end the motion flux is introduced which captures the variation of the velocity of the joints within a specific interval. Motion primitives are discovered by identifying intervals between rest instances that maximize the motion flux. The unlabeled discovered primitives have been separated into different categories using a non-parametric Bayesian mixture model. 

 We experimentally show that each primitive category naturally corresponds to movements described using biomechanical terms. Models of each primitive category are built which are then used for primitive recognition in new sequences. 
 The results show that the proposed method is able to robustly discover and recognize motion primitives from videos, by using state-of-the-art methods for estimating the 3D pose of the subject of interest. Additionally, the results suggest that the motion primitives categories are highly discriminative for characterizing the activity been performed by the subject.
 
 Finally, a dataset of motion primitives is made publicly available to further encourage result reproducibility and benchmarking of methods dealing with the discovery and recognition of human motion primitives.

\section{Acknowledgments}
This research is supported by European Union's Horizon 2020 Research and Innovation programme under grant agreement No 643950, project SecondHand \url{https://secondhands.eu/}.

\end{document}